\documentclass{article}

\usepackage{amsmath}

\usepackage{microtype}
\usepackage{graphicx}
\usepackage{subfigure}
\usepackage{booktabs} 

\usepackage{hyperref}

\usepackage{accents}

\usepackage{amsfonts}
\usepackage{amssymb}
\usepackage{amsthm}
\usepackage{bbm}
\usepackage{colortbl}
\usepackage{enumitem}
\usepackage{graphics}
\usepackage[utf8]{inputenc}
\usepackage{mathtools}
\usepackage{multirow}
\usepackage{thmtools}
\usepackage{thm-restate}
\usepackage{tikz}
\usetikzlibrary{calc}
\usetikzlibrary{positioning}
\usetikzlibrary{shapes.geometric}
\usepackage{xcolor}

\newcommand{\psuedosection}[1]{\textbf{#1. }}

\newtheorem{lemma}{Lemma}
\newtheorem{definition}{Definition}
\newtheorem{theorem}{Theorem}

\newtheorem{proof-sketch}{Proof-Sketch}
\newtheorem{claim}{Claim}

\newtheorem{corollary}{Corollary}

\newenvironment{proofsketch}{\proof}{\endproof}

\newcommand{\E}{\mathbb{E}}

\newcommand{\R}{\mathbb{R}}

\newcommand{\Otilde}{\tilde{O}}
\DeclareMathOperator{\trace}{trace}

\newcommand{\utildeS}{\underaccent{\tilde}{S}}

\newcommand{\inner}[2]{\left\langle #1, #2 \right\rangle}
\newcommand{\inneri}[3]{\sum_{i=1}^{#1} \left[\left(#2\right)_i \left(#3\right)_i \right]}
\newcommand{\abs}[1]{\left\lvert #1 \right\rvert}
\newcommand{\norm}[2]{\left\lvert \left\lvert #1 \right\rvert \right\rvert_{#2}}

\newcommand{\rina}[1]{\textcolor{red}{[R: #1]}}
\newcommand{\nuke}[1]{}

\newcommand{\dimension}[1]{d_{#1}}
\newcommand{\dsketch}{\dimension{\text{sketch}}}

\newcommand{\Enc}{\mathrm{Enc}}
\newcommand{\Dec}{\mathrm{Dec}}
\newcommand{\rsketch}{\R^{\dsketch}}
\newcommand{\rsketchn}{\R^{\dsketch \times N}}

\newcommand{\transparentx}[1]{\frac{I + #1}{2}}
\newcommand{\ptransparentx}[1]{\left(\transparentx{#1}\right)}
\newcommand{\ttransparentx}[1]{\left(\tfrac{I + #1}{2}\right)}

\newcommand{\thetastar}{{\theta^\star}}
\newcommand{\thetabar}{{\bar{\theta}}}
\newcommand{\Mstar}{{M^\star}}
\newcommand{\wstar}{{w^\star}}

\newcommand{\distribution}{\mathcal{D}}

\newcommand{\transparentdist}[1][\distribution]{\textsc{Transparent}\left(#1\right)}
\newcommand{\productdist}[1][\distribution_1, \distribution_2]{\textsc{Product}\left(#1\right)}

\newcommand{\weight}[1][\theta]{w_{#1}}
\newcommand{\depth}[1][\theta]{h(#1)}

\newcommand{\sketchx}[1]{s_{\text{#1}}}
\newcommand{\tuplesketchx}{\sketchx{tuple}}
\newcommand{\attributesketchx}{\sketchx{attr}}
\newcommand{\objectsketchx}{\sketchx{object}}
\newcommand{\inputsketchx}{\sketchx{input}}

\newcommand{\overallsketchx}{\sketchx{overall}}

\newcommand{\matrixprod}[2]{\overset{\curvearrowright}{\prod_{#1}^{#2}}}
\newcommand{\revmatrixprod}[2]{\overset{\curvearrowleft}{\prod_{#1}^{#2}}}

\newcommand{\finalsketchequations}{
  \begin{align*}
    \tuplesketchx(s_1, ..., s_k)
      &\coloneqq \sum_{i=1}^k w_i \ptransparentx{R_i} s_i \\
    \objectsketchx(\theta)
      &\coloneqq \ttransparentx{R_{M(\theta), 0}}
                 \tuplesketchx\left(
                   \attributesketchx(\theta), \inputsketchx(\theta), \tfrac12, \tfrac12
                 \right) \\
    \attributesketchx(\theta)
      &\coloneqq \tfrac12 R_{M(\theta), 1} x_\theta
                 + \tfrac12 R_{M(\theta), 2} e_1 \\
    \inputsketchx(\theta)
      &\coloneqq \tuplesketchx\left(
                   \objectsketchx(\theta_1), ..., \objectsketchx(\theta_k), w_1, ..., w_k
                 \right) \\
    \overallsketchx
      &\coloneqq \inputsketchx(\text{output pseudo-object})
  \end{align*}
}

\newcommand{\poly}{\text{poly}}

\newcommand{\property}[1]{\textbf{#1}}
\newcommand{\propattr}{\property{Attribute Recovery}}
\newcommand{\propsig}{\property{Object Signature Recovery}}
\newcommand{\propsim}{\property{Sketch-to-Sketch Similarity}}
\newcommand{\propstat}{\property{Summary Statistics}}
\newcommand{\properase}{\property{Graceful Erasure}}

\newcommand{\iproperty}[1]{\emph{#1}}
\newcommand{\ipropattr}{\iproperty{attribute recovery}}
\newcommand{\iproprecur}{\iproperty{recursive sketch}}
\newcommand{\ipropsim}{\iproperty{sketch-to-sketch similarity}}
\newcommand{\ipropstat}{\iproperty{summary statistics}}
\newcommand{\iproperase}{\iproperty{graceful erasure}}

\newcommand\FULL{}

\newcommand{\splitter}[2]{\ifdefined\FULL#1\else#2\fi}

\splitter{ 
  \newcommand{\supmat}[1]{Appendix{#1}}
  \newcommand{\Supmat}[1]{Appendix{#1}}
  
  \newcommand{\supmatopt}[1]{#1}
}
{ 
  \newcommand{\supmat}[1]{the supplementary material}
  \newcommand{\Supmat}[1]{The supplementary material}
  \newcommand{\supmatopt}[1]{}
}

\usepackage[accepted]{icml2019}

\icmltitlerunning{Recursive Sketches for Modular Deep Learning}

\begin{document}
\splitter{\onecolumn}{}

\icmltitle{Recursive Sketches for Modular Deep Learning}

\icmlsetsymbol{equal}{*}

\begin{icmlauthorlist}
\icmlauthor{Badih Ghazi}{equal,google}
\icmlauthor{Rina Panigrahy}{equal,google}
\icmlauthor{Joshua R. Wang}{equal,google}
\end{icmlauthorlist}
\icmlaffiliation{google}{Google Research, Mountain View, CA, USA.}

\icmlcorrespondingauthor{Rina Panigrahy}{rinap@google.com}

\icmlkeywords{Sketching, Modular Network}

\vskip 0.3in

\printAffiliationsAndNotice{\icmlEqualContribution} 

\splitter{}{
\setlength{\abovedisplayskip}{0pt}
\setlength{\belowdisplayskip}{0pt}
}

\begin{abstract}
We present a mechanism to compute a sketch (succinct summary) of how a complex modular deep network processes its inputs. The sketch summarizes essential information about the inputs and outputs of the network and can be used to quickly identify key components and summary statistics of the inputs. Furthermore, the sketch is recursive and can be unrolled to identify sub-components of these components and so forth, capturing a potentially complicated DAG structure. These sketches erase gracefully; even if we erase a fraction of the sketch at random, the remainder still retains the ``high-weight'' information present in the original sketch. The sketches can also be organized in a repository to implicitly form a ``knowledge graph''; it is possible to quickly retrieve sketches in the repository that are related to a sketch of interest; arranged in this fashion, the sketches can also be used to learn emerging concepts by looking for new clusters in sketch space. Finally, in the scenario where we want to learn a ground truth deep network, we show that augmenting input/output pairs with these sketches can theoretically make it easier to do so.
\end{abstract}
\section{Introduction}

Machine learning has leveraged our understanding of how the brain functions to devise better algorithms. Much of classical machine learning focuses on how to \emph{correctly} compute a function; we utilize the available data to make more accurate predictions. More recently, lines of work have considered other important objectives as well: we might like our algorithms to be small, efficient, and robust. This work aims to further explore one such sub-question: can we design a system on top of neural nets that efficiently stores information?

Our motivating example is the following everyday situation. Imagine stepping into a room and briefly viewing the objects within. Modern machine learning is excellent at answering immediate questions about this scene: ``Is there a cat? How big is said cat?'' Now, suppose we view this room every day over the course of a year. Humans can reminisce about the times they saw the room: ``How often did the room contain a cat? Was it usually morning or night when we saw the room?''; can we design systems that are also capable of efficiently answering such memory-based questions?

Our proposed solution works by leveraging an existing (already trained) machine learning model to understand individual inputs. For the sake of clarity of presentation, this base machine learning model will be a modular deep network.\footnote{Of course, it is possible to cast many models as deep modular networks by appropriately separating them into modules.} We then augment this model with sketches of its computation. We show how these sketches can be used to efficiently answer memory-based questions, despite the fact that they take up much less memory than storing the entire original computation.

\begin{figure}[ht!]
\centering
\splitter{
\includegraphics[width=120mm]{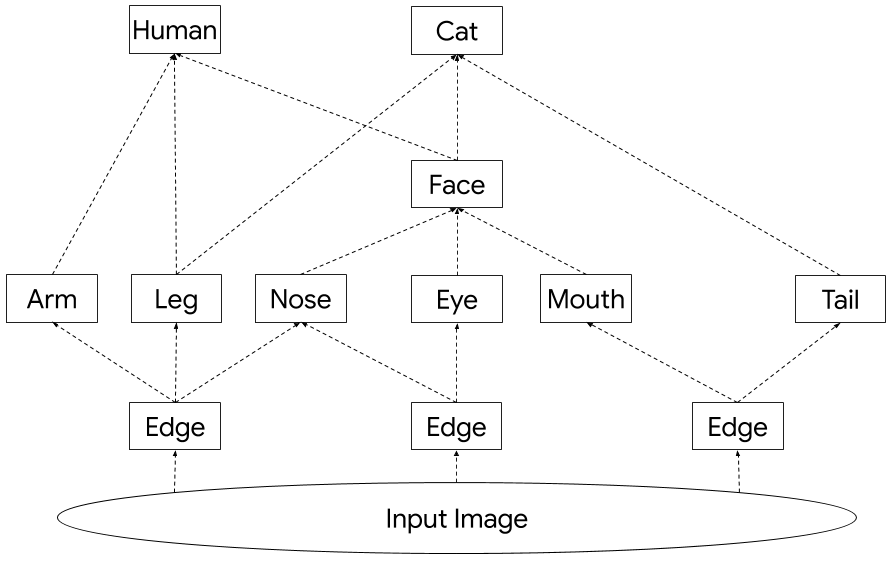}
}{
\includegraphics[width=55mm]{images/Modular_Network_Prime_Paper_Edition.png}
}
\caption{Cartoon depiction of a modular network processing an image containing a human and a cat. Modules are drawn as rectangles. The arrows run from module to module. There may be additional layers between the modules -- this is indicated by the dashed arrows.}
\label{fig:modular-network-prime-no-attributes}
\end{figure}

\begin{figure}[ht!]
\centering
\splitter{
\includegraphics[width=100mm]{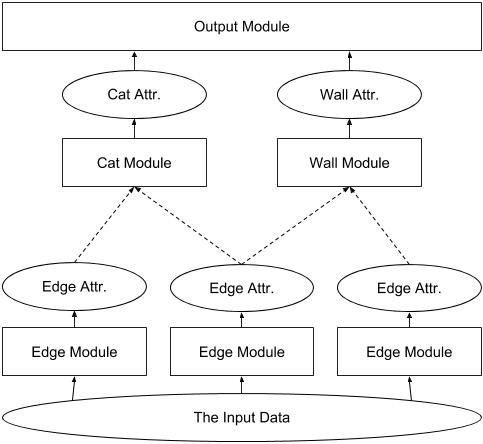}
}{
\includegraphics[width=55mm]{images/modular-network-prime.jpg}
}
\caption{Cartoon depiction of a modular network processing an image of a room. Modules are drawn as rectangles and their inputs/outputs (which we refer to as object attributes) are drawn as ovals. The arrows run from input vector to module and module to output vector. There may be additional layers between low level and high level modules, indicated by the dashed arrows. The output module here is a dummy module which groups together top-level objects.}
\label{fig:modular-network-prime}
\end{figure}

A modular deep network consists of several independent neural networks (modules) which only communicate via one's output serving as another's input. Figures~\ref{fig:modular-network-prime-no-attributes} and~\ref{fig:modular-network-prime} present cartoon depictions of modular networks. Modular networks have both a biological basis~\cite{azam2000biologically} and evolutionary justification~\cite{clune2013evolutionary}. They have inspired several practical architectures such as neural module networks~\cite{andreas2016neural, hu2017learning}, capsule neural networks~\cite{hinton2000learning,hinton2011transforming,sabour2017dynamic}, and PathNet~\cite{fernando2017pathnet} and have connections to suggested biological models of intelligence such as hierarchical temporal memory~\cite{hawkins2007intelligence}. We choose them as a useful abstraction to avoid discussing specific network architectures.

What do these modules represent in the context of our room task? Since we are searching for objects in the room, we think of each module as attempting to identify a particular type of object, from the low level edge to the high level cat. For the reader familiar with convolutional neural networks, it may help to think of each module as a filter or kernel. We denote the event where a module produces output as an object, the produced output vector as an attribute vector, and all objects produced by a module as a class. In fact, our sketching mechanism will detail how to sketch each object, and these sketches will be \emph{recursive}, taking into account the sketches corresponding to the inputs of the module.

Armed with this view of modular networks for image processing, we now entertain some possible sketch properties. One basic use case is that from the output sketch, we should be able to say something about the attribute vectors of the objects that went into it. This encompasses a question like ``How big was the cat?'' Note that we need to assume that our base network is capable of answering such questions, and the primary difficulty stems from our desire to maintain this capability despite only keeping a small sketch. Obviously, we should not be able to answer such questions as precisely as we could with the original input. Hence a reasonable \ipropattr{} goal is to be able to approximately recover an object's attribute vector from its sketch. Concretely, we should be able to recover cat attributes from our cat sketch.

Since we plan to make our sketches recursive, we would actually like to recover the cat attributes from the final output sketch. Can we recover the cat's sketch from final sketch? Again, we have the same tug-of-war with space and should expect some loss from incorporating information from many sketches into one. Our \iproprecur{} property is that we can recover (a functional approximation of) a sketch from a later sketch which incorporated it.

Zooming out to the entire scene, one fundamental question is ``Have I been in this room before?'' In other words, comparing sketches should give some information about how similar the modules and inputs that generated them. In the language of sketches, our desired \ipropsim{} property is that two completely unrelated sketches will be dissimilar while two sketches that share common modules and similar objects will be similar.

Humans also have the impressive ability to recall approximate counting information pertaining to previous encounters \cite{brannon2003nonverbal}. The \ipropstat{} property states that from a sketch, we can approximately recover the number of objects produced by a particular module, as well as their mean attributes.

Finally, we would like the nuance of being able to forget information gradually, remembering more important facts for longer. The \iproperase{} property expresses this idea: if (a known) portion of our sketch is erased, then the previous properties continue to hold, but with additional noise depending on the amount erased.

We present a sketching mechanism which simultaneously satisfies all these desiderata: \ipropattr{}, \iproprecur{}, \ipropsim{}, \ipropstat{}, and \iproperase{}.

\subsection{Potential Applications}

We justify our proposed properties by describing several potential applications for a sketching mechanism satisfying them.

\psuedosection{Sketch Repository} The most obvious use case suggested by our motivating example is to maintain a repository of sketches produced by the (augmented) network. By the \iproprecur{} and \ipropattr{} properties, we can query our repository for sketches with objects similar to a particular object. From the \ipropsim{} property, we can think of the sketches as forming a knowledge graph; we could cluster them to look for patterns in the data or use locality-sensitive hashing to quickly fish out related sketches. We can combine these overall sketches once more and use the \ipropstat{} property to get rough statistics of how common certain types of objects are, or what a typical object of a class looks like. Our repository may follow a generalization of a ``least recently used'' policy, taken advantage of the \iproperase{} property to only partially forget old sketches.

Since our sketches are recursively defined, such a repository is not limited to only keeping overall sketches of inputs; we can also store lower-level object sketches. When we perform object attribute recovery using the \ipropattr{} property, we can treat the result as a fingerprint to search for a more accurate sketch of that object in our repository.

\psuedosection{Learning New Modules} This is an extension of the sketch repository application above. Rather than assuming we start with a pre-trained modular network, we start with a bare-bones modular network and want to grow it. We feed it inputs and examine the resulting sketches for emerging clusters. When we find such a cluster, we use its points to train a corresponding (reversible) module. In fact, this clustering may be done in a hierarchical fashion to get finer classes. For example, we may develop a module corresponding to ``cat'' based on a coarse cluster of sketches which involve cats. We may get sub-clusters based on cat species or even finer sub-clusters based on individual cats repeatedly appearing.

The modules produced by this procedure can capture important primitives. For example, the \ipropstat{} property implies that a single-layer module could count the number of objects produced by another module. If the modules available are complex enough to capture curves, then this process could potentially allow us to identify different types of motion.

\psuedosection{Interpretability} Since our sketches store information regarding how the network processed an input, they can help explain how that network came to its final decision. The \ipropsim{} property tells us when the network thinks two rooms are similar, but what exactly did the network find similar? Using the \iproprecur{} property, we can drill down into two top level sketches, looking for the pairs of objects that the network found similar.

\psuedosection{Model Distillation} Model distillation \cite{bucilua2006model, hinton2015distilling} is a popular technique for improving machine learning models. The use case of model distillation is as follows. Imagine that we train a model on the original data set, but it is unsatisfactory in some regard (e.g. too large or ensemble). In model distillation, we take this first model and use its features to train a second model. These stepping stones allows the second model to be smaller and less complex. The textbook example of feature choice is to use class probabilities (e.g. the object is $90\%$ A and $10\%$ B) versus the exact class found in the original input (e.g. the object is of class A). We can think of this as an intermediate option between the two extremes of (i) providing just the original data set again and (ii) providing all activations in the first model. Our sketches are an alternative option. To showcase the utility of our sketches in the context of learning from a teacher network, we prove that (under certain assumptions) augmenting the teacher network with our sketches allows us to learn the network. Note that in general, it is not known how the student can learn solely from many input/output pairs.

\subsection{Techniques}

The building block of our sketching mechanism is applying a random matrix to a vector. We apply such transformations to object attribute vectors to produce sketches, and then recursively to merge sketches. Consequently, most of our analysis revolves around proving and utilizing properties of random matrices. In the case where we know the matrices used in the sketch, then we can use an approximate version of isometry to argue about how noise propagates through our sketches. On the other hand, if we do not know the matrices involved, then we use sparse recovery/dictionary learning techniques to recover them. Our recovery procedure is at partially inspired by several works on sparse coding and dictionary learning \cite{spielman2012exact,arora2014new}, \cite{arora2014provable} and \cite{arora2015simple} as well as their known theoretical connections to neural networks \cite{pmlr-v32-arora14, papyan2018theoretical}. We have more requirements on our procedure than previous literature, so we custom-design a distribution of block-random matrices reminiscent of the sparse Johnson-Lindenstrauss transform. Proving procedure correctness is done via probabilistic inequalities and analysis tools, including the Khintchine inequality and the Hanson-Wright inequality.

One could consider an alternate approach based on the serialization of structured data (for example, protocol buffers). Compared to this approach, ours provides more fine-grained control over the accuracy versus space trade-off and aligns more closely with machine learning models (it operates on real vectors).

\subsection{Related Work}

There have been several previous attempts at adding a notion of memory to neural networks. There is a line of work which considers augmenting recurrent neural networks with external memories \cite{graves2014neural, sukhbaatar2015end, joulin2015inferring, grefenstette2015learning, zaremba2015reinforcement, danihelka2016associative, graves2016hybrid}. In this line of work, the onus is on the neural network to learn how to write information to and read information from the memory. In contrast to this line of work, our approach does not attempt to use machine learning to manage memory, but rather demonstrates that proper use of random matrices suffices. Our results can hence be taken as theoretical evidence that this learning task is relatively easy.

Vector Symbolic Architectures (VSAs)~\cite{smolensky1990tensor, gayler2004vector, levy2008vector} are a class of memory models which use vectors to represent both objects and the relationships between them. There is some conceptual overlap between our sketching mechanism and VSAs. For example, in Gayler's MAP (Multiply, Add, Permute) scheme~\cite{gayler2004vector}, vector addition is used to represent superposition and permutation is used to encode quotations. This is comparable to how we recursively encode objects; we use addition to combine multiple input objects after applying random matrices to them. One key difference in our model is that the object attribute vectors we want to recover are provided by some pre-trained model and we make no assumptions on their distribution. In particular, their possible correlation makes it necessary to first apply a random transformation before combining them. In problems where data is generated by simple programs, several papers \cite{nsd, nsvqa, ellis2018learning, andreas2016neural, oh2017zero} attempt to infer programs from the generated training data possibly in a modular fashion.

Our result on the learnability of a ground truth network is related to the recent line of work on learning small-depth neural networks  \cite{du2017gradient, du2017convolutional, li2017convergence, zhong2017learning, zhang2017electron, zhong2017recovery}. Recent research on neural networks has considered evolving neural network architectures for image classification (e.g., \cite{real2018regularized}), which is conceptually related to our suggested Learning New Models application.

\subsection{Organization}

We review some basic definitions and assumptions in Section~\ref{sec:prelim}. An overview of our final sketching mechanism is given in Section~\ref{sec:final-sketch} (Theorems~\ref{thm:simple-attr}, \ref{thm:simple-sim}, \ref{thm:simple-stat}). \splitter{Section~\ref{sec:prototype-sketch} gives a sequence of sketch prototypes leading up to our final sketching mechanism.}{} In Section~\ref{sec:dict_learning}, we explain how to deduce the random parameters of our sketching mechanism from multiple sketches via dictionary learning. Further discussion and detailed proofs are provided in \splitter{the appendices}{the supplementary material}.
\section{Preliminaries}
\label{sec:prelim}

In this section, we cover some preliminaries that we use throughout the remainder of the paper. Throughout the paper, we denote by $[n]$ the set $\{1,\dots,n\}$. \splitter{We condense large matrix products with the following notation.
\begin{align*}
  \matrixprod{i=1}{n} R_i &\coloneqq R_1 R_2 \cdots R_n \\
  \revmatrixprod{i=1}{n} R_i &\coloneqq R_n \cdots R_2 R_1
\end{align*}}{}

\subsection{Modular Deep Networks for Object Detection}

For this paper, a modular deep network consists of a collection of modules $\{M\}$. Each module is responsible for detecting a particular class of object, e.g. a cat module may detect cats. These modules use the outputs of other modules, e.g. the cat module may look at outputs of the edge detection module rather than raw input pixels. When the network processes a single input, we can depict which objects feed into which other objects as a (weighted) communication graph. This graph may be data-dependent. For example, in the original Neural Module Networks of Andreas et al. \cite{andreas2016neural}, the communication graph is derived by taking questions, feeding them through a parser, and then converting parse trees into structured queries. Note that this pipeline is disjoint from the modular network itself, which operates on images. In this paper, we will not make assumptions about or try to learn the process producing these communication graphs. When a module does appear in the communication graph, we say that it detected an object $\theta$ and refer to its output as $x_\theta$, the attribute vector of object $\theta$. We assume that these attribute vectors $x_\theta$ are nonnegative and normalized. As a consequence of this view of modular networks, a communication graph may have far more objects than modules. We let our input size parameter $N$ denote twice the maximum between the number of modules and the number of objects in any communication graph.

We assume we have access to this communication graph, and we use $\theta_1, ..., \theta_k$ to denote the $k$ input objects to object $\theta$. We assume we have a notion of how important each input object was; for each input $\theta_i$, there is a nonnegative weight $w_i$ such that $\sum_{i=1}^k w_i = 1$. It is possible naively choose $w_i = 1 / k$, with the understanding that weights play a role in the guarantees of our sketching mechanism (it does a better job of storing information about high-weight inputs).

We handle the network-level input and output as follows. The input data can be used by modules, but it itself is not produced by a module and has no associated sketch. We assume the network has a special output module, which appears exactly once in the communication graph as its sink. It produces the output (pseudo-)object, which has associated sketches like other objects but does not count as an object when discussing sketching properties (e.g. two sketches are not similar just because they must have output pseudo-objects). This output module and pseudo-object only matter for our overall sketch insofar as they group together high-level objects.

\subsection{Additional Notation for Sketching}

We will (recursively) define a sketch for every object, but we consider our sketch for the \emph{output object} of the network to be the overall sketch. This is the sketch that we want to capture how the network processed a particular input. One consequence is that if an object does not have a path to the output module in the communication graph, it will be omitted from the overall sketch.

Our theoretical results will primarily focus on a single path from an object $\theta$ to the output object. We define the ``effective weight'' of an object, denoted $\weight$, to be the product of weights along this path. For example, if an edge was 10\% responsible for a cat and the cat was 50\% of the output, then this edge has an effective weight of 5\% (for this path through the cat object). We define the ``depth'' of an object, denoted $\depth$, to be the number of objects along this path. In the preceding example, the output object has a depth of one, the cat object has a depth of two, and the edge object has a depth of three. We make the technical assumption that all objects produced by a module must be at the same depth. As a consequence, modules $M$ also have a depth, denoted $\depth[M]$. To continue the example, the output \emph{module} has depth one, the cat \emph{module} has depth two, and the edge \emph{module} has depth three. Furthermore, this implies that each object and module can be assigned a unique depth (i.e. all paths from any object produced by a module to the output object have the same number of objects).

Our recursive object sketches are all $\dsketch$ dimensional vectors. We assume that $\dsketch$ is at least the size of any object attribute vector (i.e. the output of any module). In general, all of our vectors are $\dsketch$-dimensional; we assume that shorter object attribute vectors are zero-padded (and normalized). \splitter{For example, imagine that the cat object $\theta$ is output with the three-dimensional attribute vector $(0.6, 0.0, 0.8)$, but $\dsketch$ is five. Then we consider its attribute vector to be $x_\theta = (0.6, 0.0, 0.8, 0.0, 0.0)$.}{}

\splitter{
\begin{table}
\begin{center}
\begin{tabular}{@{}c c@{}}\toprule
  \textbf{Symbol} & \textbf{Meaning} \\ \midrule
  $b$ & a bit in $\{\pm 1\}$ \\
  $d$ & dimension size \\
  $\distribution$ & distribution over random matrices \\
  $f$ & a coin flip in $\{\pm 1\}$ \\
  $h$ & depth (one-indexed) \\
  $H$ & $3 \cdot \text{maximum depth}$ \\
  $I$ & identity matrix \\
  $k$ & \#inputs \\
  $\ell$ & $\ell$-th moment of a random variable \\
  $M$ & module \\
  $N$ & $3 \cdot \max(\text{\#modules}, \text{\#objects})$ \\
  $q$ & block non-zero probability \\
  $R$ & random matrix \\
  $s$ & sketch \\
  $S$ & \#samples \\
  $w$ & importance weight \\
  $x$ & attribute vector / unknown dictionary learning vector \\
  $y$ & dictionary learning sample vector \\
  $z$ & dictionary learning noise \\
  $\delta$ & error bound, isometry and desynchronization \\
  $\epsilon$ & error bound, dictionary learning \\
  $\theta$ & object \\
  \bottomrule
\end{tabular}
\end{center}
\caption{Symbols used in this paper.}
\label{tab:symbols}
\end{table}
}{}
\section{Overview of Sketching Mechanism}\label{sec:final-sketch}

In this section, we present our sketching mechanism, which attempts to summarize how a modular deep network understood an input. The mechanism is presented at a high level; see Section~\ref{sec:prototype-sketch} for how we arrived at this mechanism and \supmat{~\ref{apx:final-sketch}} for proofs of the results stated in this section.

\subsection{Desiderata}

As discussed in the introduction, we want our sketching mechanism to satisfy several properties, listed below.

\propattr{}. Object attribute vectors can be approximately recovered from the overall sketch, with additive error that decreases with the effective weight of the object.

\propsim{}. With high probability, two completely unrelated sketches (involving disjoint sets of modules) will have a small inner product. With high probability, two sketches that share common modules and similar objects will have a large inner product (increasing with the effective weights of the objects).

\propstat{}. If a single module detects several objects, then we can approximately recover summary statistics about them, such as the number of such objects or their mean attribute vector, with additive error that decreases with the effective weight of the objects.

\properase{}. Erasing everything but $\dsketch'$-prefix of the overall sketch yields an overall sketch with the same properties (albeit with larger error).

\subsection{Random Matrices}

The workhorse of our recursive sketching mechanism is random (square) matrices. We apply these to transform input sketches before incorporating them into the sketch of the current object. We will defer the exact specification of how to generate our random matrices to Section~\ref{sec:dict_learning}, but while reading this section it may be convenient for the reader to think of our random matrices as uniform random \emph{orthonormal matrices}. Our actual distribution choice is similar with respect to the properties needed for the results in this section. One notable parameter for our family of random matrices is $\delta \ge 0$, which expresses how well they allow us to estimate particular quantities with high probability. For both our random matrices and uniform random orthonormal matrices, $\delta$ is $\Otilde(1 / \sqrt{\dsketch})$. We prove various results about the properties of our random matrices in \supmat{~\ref{apx:block-random}}.

We will use $R$ with a subscript to denote such a random matrix. Each subscript indicates a fresh draw of a matrix from our distribution, so every $R_1$ is always equal to every other $R_1$ but $R_1$ and $R_{\text{cat}, 0}$ are independent. Throughout this section, we will assume that we have access to these matrices when we are operating on a sketch to retrieve information. In Section~\ref{sec:dict_learning}, we show how to recover these matrices given enough samples.

\subsection{The Sketching Mechanism}

\splitter{
\begin{figure}[ht!]
\centering
\includegraphics[width=100mm]{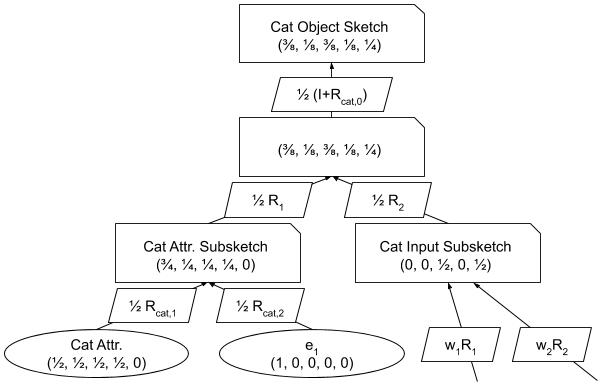}
\caption{Illustration of how to compute the sketch for a cat object from Figure~\ref{fig:modular-network-prime}. Attribute vectors (and $e_1$) are drawn as ovals; note how they have been normalized and padded to $\dsketch = 5$ dimensions. Sketches are drawn as rectangles with a cut corner. The arrows run from vectors used to compute other vectors, and are labeled with the matrix that is applied in the process. For the purposes of this diagram, all random matrices happen to be the identity matrix.}
\label{fig:sketch-prime}
\end{figure}
}{}

The basic building block of our sketching mechanism is the tuple sketch, which we denote $\tuplesketchx$. As the name might suggest, its purpose is to combine $k$ sketches $s_1, ..., s_k$ with respective weights $w_1, ..., w_k \ge 0$ into a single sketch (for proofs, we will assume that these weights sum to at most one). In the case where the $k$ sketches are input sketches, these weights will be the importance weights from our network. Otherwise, they will all be $1/k$. Naturally, sketches with higher weight can be recovered more precisely. The tuple sketch is formally defined as follows. If their values are obvious from context, we will omit $w_1, ..., w_k$ from the arguments to the tuple sketch. The tuple sketch is computed as follows.\footnote{Rather than taking a convex combination of $\ttransparentx{R_i} s_i$, one might instead sample a few of them. Doing so would have repercussions on the results in this section, swapping many bounds from high probability to conditioning on their objects getting through. However, it also makes it easier to do the dictionary learning in Section~\ref{sec:dict_learning}.}
\[ 
  \tuplesketchx(s_1, ..., s_k, w_1, ..., w_k) \coloneqq \sum_{i=1}^k w_i \ttransparentx{R_i} s_i
\]
Note that $I$ is the identity matrix, and we will define a tuple sketch of zero things to be the zero vector.

Each object $\theta$ is represented by a sketch, which naturally we will refer to as an object sketch. We denote such a sketch as $\objectsketchx$. We want the sketch of object $\theta$ to incorporate information about the attributes of $\theta$ itself as well as information about the inputs that produced it. Hence we also define two subsketches for object $\theta$. The attribute subsketch, denoted $\attributesketchx$, is a sketch representation of object $\theta$'s attributes. The input subsketch, denoted $\inputsketchx$, is a sketch representation of the inputs that produced object $\theta$. These three sketches are computed as follows.
\begin{align*}
  \objectsketchx(\theta)
    &\coloneqq \ttransparentx{R_{M(\theta), 0}}
               \tuplesketchx\left(
                 \attributesketchx(\theta), \inputsketchx(\theta), \tfrac12, \tfrac12
               \right) \\
  \attributesketchx(\theta)
    &\coloneqq \tfrac12 R_{M(\theta), 1} x_\theta
               + \tfrac12 R_{M(\theta), 2} e_1 \\
  \inputsketchx(\theta)
    &\coloneqq \tuplesketchx\left(
                 \objectsketchx(\theta_1), ..., \objectsketchx(\theta_k), w_1, ..., w_k
               \right)
\end{align*}
Note that $e_1$ in the attribute subsketch is the first standard basis vector; we will use it for frequency estimation as part of our \propstat{} property. Additionally, $\theta_1, ..., \theta_k$ in the input sketch are the input objects for $\theta$ and $w_1, ..., w_k$ are their importance weights from the network.

The overall sketch is just the output psuedo-object's \emph{input subsketch}. It is worth noting that we do not choose to use its object sketch.
\[ 
  \overallsketchx \coloneqq \inputsketchx(\text{output pseudo-object})
\]
We want to use this to (noisily) retrieve the information that originally went into these sketches. We are now ready to present our results for this sketching mechanism. Note that we provide the technical machinery for proofs of the following results in \supmat{~\ref{apx:tm}} and we provide the proofs themselves in \supmat{~\ref{apx:final-sketch}}. Our techniques involve using an approximate version of isometry and reasoning about the repeated application of matrices which satisfy our approximate isometry.

Our first result concerns \propattr{} (example use case: roughly describe the cat that was in this room).

\begin{theorem}\supmatopt{[Simplification of Theorem~\ref{thm:final-attr}]}\label{thm:simple-attr}
  Our sketch has the simplified \propattr{} property, which is as follows. Consider an object $\thetastar$ at constant depth $h(\thetastar)$ with effective weight $w_\thetastar$.
  \begin{enumerate}[nosep]
    \item[(i)] Suppose no other objects in the overall sketch are also produced by $M(\thetastar)$. We can produce a vector estimate of the attribute vector $x_\thetastar$, which with high probability has at most $O(\delta / w_\thetastar)$ $\ell_\infty$-error.
    \item[(ii)] Suppose we know the sequence of input indices to get to $\thetastar$ in the sketch. Then even if other objects in the overall sketch are produced by $M(\thetastar)$, we can still produce a vector estimate of the attribute vector $x_\thetastar$, which with high probability has at most $O(\delta / w_\thetastar)$ $\ell_\infty$-error.
  \end{enumerate}
\end{theorem}
As a reminder, $\ell_\infty$ error is just the maximum error over all coordinates. Also, note that if we are trying to recover a quantized attribute vector, then we may be able to recover our attribute vector \emph{exactly} when the quantization is larger than our additive error. This next result concerns \propsim{} (example use case: judge how similar two rooms are). We would like to stress that the output psuedo-object does not count as an object for the purposes of (i) or (ii) in the next result.
\begin{theorem}\supmatopt{[Simplification of Theorem~\ref{thm:final-sim}]}\label{thm:simple-sim}
  Our sketch has the \propsim{} property, which is as follows. Suppose we have two overall sketches $s$ and $\bar{s}$.
  \begin{enumerate}[nosep]
    \item[(i)] If the two sketches share no modules, then with high probability they have at most $O(\delta)$ dot-product.
    \item[(ii)] If $s$ has an object $\thetastar$ of weight $w_\thetastar$ and $\bar{s}$ has an object $\bar{\theta}^\star$ of weight $\bar{w}_\thetastar$, both objects are produced by the same module, and both objects are at constant depth $h(\thetastar)$ then with high probability they have at least $\Omega(w_\thetastar \bar{w}_{\bar{\theta}^\star}) - O(\delta)$ dot-product.
    \item[(iii)] If the two sketches are identical except that the attributes of any object $\theta$ differ by at most $\epsilon$ in $\ell_2$ distance, then with probability one the two sketches are at most $\epsilon / 2$ from each other in $\ell_2$ distance.
  \end{enumerate}
\end{theorem}
Although our \propsim{} result is framed in terms of two overall sketches, the recursive nature of our sketches makes it not too hard to see how it also applies to any pair of sketches computed along the way. \splitter{This is depicted in Figure~\ref{fig:prop-sim}.

\begin{figure}[ht!]
\centering
\includegraphics[width=100mm]{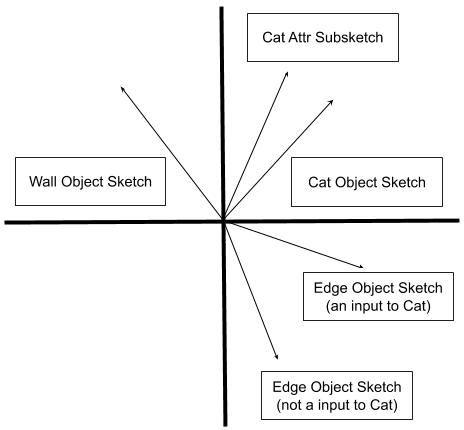}
\caption{A two-dimensional cartoon depiction of the \propsim{} property with respect to the various sketches we produce from a single input. Each vector represents a sketch, and related sketches are more likely to cluster together.}
\label{fig:prop-sim}
\end{figure}
}{}

This next result concerns \propstat{} (example use case: how many walls are in this room).
\begin{theorem}\supmatopt{[Simplification of Theorem~\ref{thm:final-stat}]}\label{thm:simple-stat}
  Our sketch has the simplified \propstat{} property, which is as follows. Consider a module $\Mstar$ which lies at constant depth $h(\Mstar)$, and suppose all the objects produced by $\Mstar$ has the same effective weight $\wstar$.
  \begin{enumerate}[nosep]
    \item[(i)] \property{Frequency Recovery}. We can produce an estimate of the number of objects produced by $\Mstar$. With high probability, our estimate has an additive error of at most $O(\delta / \wstar)$.
    \item[(ii)] \property{Summed Attribute Recovery}. We can produce a vector estimate of the summed attribute vectors of the objects produced by $\Mstar$. With high probability, our estimate has at most $O(\delta / \wstar)$ $\ell_\infty$-error.
  \end{enumerate}
\end{theorem}

Again, note that quantization may allow for exact recovery. In this case, the number of objects is an integer and hence can be recovered exactly when the additive error is less than $1/2$. If we can recover the exact number of objects, we can also estimate the mean attribute vector.

\splitter{Appendix Subsection~\ref{subsec:sig-sketch}}{The supplementary material} explains how to generalize our sketch to incorporate signatures for the \propsig{} property and \splitter{Appendix Subsection~\ref{subsec:erase-sketch}}{also} explains how to generalize these results to hold when part of the sketch is erased for the \properase{} property.
\splitter{
\section{The Road to a Sketch}\label{sec:prototype-sketch}

In this section, we present a series of sketch prototypes which gradually grow more complex, capturing more of our desiderata while also considering more complex network possibilities. The final prototype is our actual sketch.

We begin by making the following simplifying assumptions: (i) each object is produced by a unique module, (ii) our network is of depth one (i.e. all objects are direct inputs to the output pseudo-object), and (iii) we can generate random orthonormal matrices $R$ to use in our sketches. In this section, we eventually relax all three assumptions to arrive at our final sketch, which was presented in Section~\ref{sec:final-sketch}. Orthonormal matrices are convenient for us due to the following two properties.

\property{Isometry}: If $x$ is a unit vector and $R$ is an orthonormal matrix, then $\abs{\norm{Rx}{2}^2 - \norm{x}{2}^2} = 0$. One nice consequence of this fact is that $R^T$ is the inverse of $R$.

\property{Desynchronization}: If $x$ and $y$ are unit vectors and then $R$ is drawn to be a random orthonormal matrix, then $\abs{\inner{Rx}{y}} \le O(\sqrt{\log N} / \sqrt{\dsketch})$ with probability at least $1 - 1 / N^{\Omega(1)}$ (i.e. with high probability).

When we later use block sparse random matrices, these properties are noisier and we need to carefully control the magnitude of the resulting noise. Roughly speaking, the absolute values in both definitions will be bounded by a parameter $\delta$ (the same $\delta$ as in Section~\ref{sec:final-sketch}). In \splitter{Appendix~\ref{apx:tm}}{the supplementary material}, we establish the technical machinery to control this noise.

\psuedosection{Proof Ideas} \Supmat{~\ref{apx:prototype-sketch}} contains proof sketches of the claims in this section. At an even higher level, the key idea is as follows. Our sketch properties revolve around taking products of our sketches and random matrices. These products are made up of terms which in turn contain matrix products of the form $R_2^T R_1$. If these two random matrices are the same, then this is the identity matrix by Isometry. If they are independent, then this is a uniform random orthonormal matrix and by desynchronization everything involved in the product will turn into bounded noise. The sharp distinction between these two cases means that $Rx$ encodes the information of $x$ in a way that only applying $R^T$ will retrieve.

\psuedosection{Prototype A: A Basic Sketch} Here is our first attempt at a sketch, which we will refer to as prototype A. We use A superscripts to distinguish the definitions for this prototype. The fundamental building block of all our sketches is the tuple sketch, which combines $k$ (weighted) sketches into a single sketch. Its input is $k$ sketches $s_1, ..., s_k$ along with their nonnegative weights $w_1, ..., w_k \ge 0$ which sum to one. For now, we just take the convex combination of the sketches:
\[
  \tuplesketchx^A(s_1, ..., s_k, w_1, ..., w_k) \coloneqq \sum_{i=1}^k w_i s_i
\]

Next, we describe how to produce the sketch of an object $\theta$. Due to assumption (ii), we don't need to incorporate the sketches of input objects. We only need to take into account the object attribute vector. We will keep a random matrix $R_M$ for every module $M$ and apply it when computing object sketches:
\[
  \objectsketchx^A(\theta) \coloneqq R_{M(\theta)} x_\theta
\]

Finally, we will specially handle the sketch of our top level object so that it can incorporate its direct inputs (we will not bother with its attributes for now):
\[
  \overallsketchx^A \coloneqq \tuplesketchx^A( \objectsketchx^A( \theta_1 ), ..., \objectsketchx^A( \theta_k ) )
\]
where $\theta_1, ..., \theta_k$ are the direct inputs to the output pseudo-object and $w_1, ..., w_k$ are their importance weights from the network.

Our first result is that if we have the overall sketch $\overallsketchx^A$, we can (approximately) recover the attribute vector for object $\theta$ by computing $R_{M(\theta)}^T \overallsketchx^A$.
\begin{restatable}{claim}{aattr}\label{clm:a-attr}
  Prototype A has the \propattr{} property: for any object $\thetastar$, with high probability (over the random matrices), its attribute vector $x_\thetastar$ can be recovered up to at most $O(\log N / (w_\thetastar \sqrt{\dsketch}))$ $\ell_\infty$-error.
\end{restatable}

Our next result is that overall sketches which have similar direct input objects will be similar:
\begin{restatable}{claim}{asim}\label{clm:a-sim}
  Prototype A has the \propsim{} property: (i) if two sketches share no modules, then with high probability they have at most $O(\sqrt{\log N} / \sqrt{\dsketch})$ dot-product, (ii) if two sketches share an object $\thetastar$, which has $w_\thetastar$ in the one sketch and $\bar{w}_\thetastar$ in the other, then with high probability they have at least $w_\thetastar \bar{w}_\thetastar - O(\sqrt{\log N} / \sqrt{\dsketch})$ dot-product, (iii) if two sketches have objects produced by identical modules and each pair of objects produced by the same module differ by at most $\epsilon$ in $\ell_2$ distance and are given equal weights, then they are guaranteed to be at most $\epsilon$ from each other in $\ell_2$ distance.
\end{restatable}

\psuedosection{Prototype B: Handling Module Multiplicity} We will now relax assumption (i); a single module may appear multiple times in the communication graph, producing multiple objects. This assumption previously prevented the \propstat{} property from making sense, but it is now up for grabs. Observe that with this assumption removed, computing $R_M^T \overallsketchx^A$ now sums the attribute vectors of \emph{all objects produced by module} $M$. As a consequence, we will now need a more nuanced version of our \propattr{} property, but we can leverage this fact for the \propstat{} property. In order to estimate object frequencies, we will (in spirit) augment our object attribute vectors with a entry which is always one. In reality, we draw another random matrix for the module and use its first column.
\[
  \objectsketchx^B(\theta) \coloneqq \tfrac12 R_{M(\theta), 1} x_\theta + \tfrac12 R_{M(\theta), 2} e_1
\]
where $e_1$ is the first standard basis vector in $\dsketch$ dimensions.

We still need to fix attribute recovery; how can we recover the attributes of an object that shares a module with another object? We could apply another random matrix to every input in a tuple sketch, but then we would not be able to get summed statistics. The solution is to both apply another random matrix and apply the identity mapping. We will informally refer to $\ptransparentx{R}$ as the transparent version of matrix $R$. The point is that when applied to a vector $x$, it both lets through information about $x$ and sets things up so that $R^T$ will be able to retrieve the same information about $x$. Our new tuple sketch is:
\[
  \tuplesketchx^B(s_1, \ldots, s_k, w_1, \ldots, w_k) \coloneqq \sum_{i=1}^k w_i \ttransparentx{R_i} s_i
\]

Our special overall sketch is computed as before.
\[
  \overallsketchx^B \coloneqq \tuplesketchx^B( \objectsketchx^B( \theta_1 ), ..., \objectsketchx^B( \theta_k ) )
\]

Here is our more nuanced attribute recovery claim.
\begin{restatable}{claim}{battr}\label{clm:b-attr}
  Prototype B has the \propattr{} property, which is as follows. Consider an object $\thetastar$ with weight $w_\thetastar$. Then (i) if no other objects are produced by the same module in the sketch, with high probability we can recover $w_\theta$ up to $O(\frac{\sqrt{\log N}}{w_\thetastar \sqrt{\dsketch}})$ $\ell_\infty$-error and (ii) even if other objects are output by the same module, if we know $\thetastar$'s index in the overall sketch, with high probability we can recover $w_\theta$ up to $O(\frac{\sqrt{\log N}}{w_\thetastar \sqrt{\dsketch}})$ $\ell_\infty$-error.
\end{restatable}

Sketches are still similar, and due to the $e_1$ we inserted into the object sketch, two objects produced by the same module now always have a base amount of similarity.
\begin{restatable}{claim}{bsim}\label{clm:b-sim}
  Prototype B has the \propsim{} property, just as in Claim~\ref{clm:a-sim}, except in case (ii) the dot-product is at least $\frac14 w_\thetastar \bar{w}_\thetastar - O(\sqrt{\log N} / \sqrt{\dsketch})$, in case (iii) the sketches are at most $\epsilon / 2$ from each other in $\ell_2$ distance, and we have an additional case (iv) if one sketch has an object $\thetastar$ of weight $w_\thetastar$ and the other sketch has an object $\bar{\theta}^\star$ of weight $\bar{w}_{\bar{\theta}^\star}$ and both objects are produced by the same module, then with high probability they have at least $\frac{1}{16} w_\thetastar \bar{w}_\thetastar - O(\sqrt{\log N} / \sqrt{\dsketch})$.
\end{restatable}

Finally, we can recover some summary statistics.
\begin{restatable}{claim}{bstat}\label{clm:b-stat}
  Prototype B has the \propstat{} property, which is as follows. Consider a module $\Mstar$, and suppose all objects produced by $\Mstar$ have the same weight in the overall sketch. Then (i) we can estimate the number of objects produced by $\Mstar$ up to an additive $O(\frac{\sqrt{\log N}}{\wstar \sqrt{\dsketch}})$ and (ii) we can estimate the summed attribute vector of objects produced by $\Mstar$ up to $O(\frac{\sqrt{\log N}}{\wstar \sqrt{\dsketch}})$ $\ell_\infty$-error.
\end{restatable}

\psuedosection{Last Steps: Handling Deep Networks} We conclude by describing how to get from prototype B to the sketch we presented in Section~\ref{sec:final-sketch}. We have two assumptions left to relax: (ii) and (iii). We gloss over how to relax assumption (iii) here, since the push back comes from the requirements of dictionary learning in Section~\ref{sec:dict_learning}, and the exact choice of random matrices is not so illuminating for this set of results. Instead, we focus on how to relax assumption (ii) and make our sketches recursive. As a first point, we'll be keeping the tuple sketch from prototype B:
\[ 
  \tuplesketchx(s_1, ..., s_k, w_1, ..., w_k) \coloneqq \sum_{i=1}^k w_i \ttransparentx{R_i} s_i \splitter{\qquad ( = \tuplesketchx^B(s_1, ..., s_k, w_1, ..., w_k) )}{}
\]

The main difference is that we now want object sketches to additionally capture information about their input objects. We define two important subsketches for each object, one to capture its attributes and one to capture its inputs. The attribute subsketch of an object what was the object sketch in prototype B:
\[
  \attributesketchx(\theta) \coloneqq \tfrac12 R_{M(\theta), 1} x_\theta + \tfrac12 R_{M(\theta), 2} e_1 \splitter{\qquad ( = \objectsketchx^B (\theta) )}{}
\]

The input subsketch is just a tuple sketch:
\[
  \inputsketchx(\theta) \coloneqq \tuplesketchx\left( \objectsketchx(\theta_1), ..., \objectsketchx(\theta_k), w_1, ..., w_k \right)
\]

The object sketch just tuple sketches together the two subsketches and applies a transparent matrix. The idea is that we might both want to query over the input object's module, or both the current object's module and the input object's module (e.g. summary statistics about all edges versus just edges that wound up in a cat). The transparent matrix means the information gets encoded both ways.
\[
  \objectsketchx(\theta) \coloneqq \ttransparentx{R_{M(\theta), 0}} \tuplesketchx\left( \attributesketchx(\theta), \inputsketchx(\theta), \tfrac12, \tfrac12 \right)
\]

One nice consequence of all this is that we can now stop special-casing the overall sketch, since we have defined the notion of input subsketches. Note that we \emph{do not want} the overall sketch to be the output pseudo-object's object sketch, because that would incorporate a $R_{\text{output}, 2} e_1$, which in turn gives all sketches a base amount of dot-product similarity that breaks the current \propsim{} property.
\[
  \overallsketchx \coloneqq \inputsketchx(\text{output pseudo-object})
\]
}{}
\section{Learning Modular Deep Networks via Dictionary Learning}
\label{sec:dict_learning}

In this section, we demonstrate that our sketches carry enough information to learn the network used to produce them. Specifically, we develop a method for training a new network based on (input, output, sketch) triples obtained from a teacher modular deep network. Our method is powered by a novel \emph{dictionary learning} procedure. Loosely speaking, dictionary learning tries to solve the following problem. There is an unknown dictionary matrix $R$, whose columns are typically referred to as atoms. There is also a sequence of unknown sparse vectors $x^{(k)}$; we only observe how they combine the atoms, i.e., $\{y^{(k)} = R x^{(k)}\}$. We want to use these observations $y^{(k)}$ to recover the dictionary matrix $R$ and the original sparse vectors $x^{(k)}$.

While dictionary learning has been well-studied in the literature from both an applied and a theoretical standpoint, our setup differs from known theoretical results in several key aspects. The main complication is that since we want to apply our dictionary learning procedure recursively, our error in recovering the unknown vectors $x^{(k)}$ will become noise on the observations $y^{(k)}$ in the very next step. Note that classical dictionary learning can only recover the unknown vectors $x^{(k)}$ up to permutation and coordinate-wise sign. To do better, we will carefully engineer our distribution of dictionary matrices $R$ to allow us to infer the permutation (between the columns of a matrix) and signs, which is needed to recurse. Additionally, we want to allow very general unknown vectors $x^{(k)}$. Rather than assuming sparsity, we instead make the weaker assumption that they have bounded $\ell_2$ norm. We also allow for distributions of $x^{(k)}$ which make it impossible to recover all columns of $R$; we simply recover a subset of essential columns.

With this in mind, our desired dictionary learning result is Theorem~\ref{thm:dictionary-learning}. We plan to apply this theorem with $N$ as thrice the maximum between the number of modules and the number of objects in any communication graph, $\utildeS$ as the number of samples necessary to learn a module, and $H$ as three times the depth of our modular network. Additionally, we think of $\epsilon_1$ as the tolerable input $\ell_\infty$ error while $\epsilon_H$ is the desired output $\ell_\infty$ error after recursing $H$ times.
\begin{restatable}{theorem}{recursabledictionarylearning}[Recursable Dictionary Learning]
\label{thm:dictionary-learning}
  There exists a family of distributions $\{\distribution(b, q, \dsketch)\}$ which produce $\dsketch \times \dsketch$ matrices satisfying the following. For any positive integer $N$, positive constant $H$, positive real $\epsilon_H$, base number of samples $S \le \poly(N)$, block size $b \ge \poly(\log N, \log \dsketch, 1/\epsilon_H)$, nonzero block probability $q \ge \poly(\log N, \log \dsketch, 1/\epsilon_H) / \sqrt{\dsketch}$, and dimension $\dsketch \ge \poly(1/\epsilon_H, \log N, \log S)$, there exists a sequence of $\ell_\infty$ errors $(0 <) \epsilon_1 \le \epsilon_2 \le \cdots \le \epsilon_{H-1} (\le \epsilon_H)$ with $\epsilon_1 \ge \poly(\epsilon_H)$, such that the following is true. For any $h \in [H-1]$, let $S_h = S N^{h-1}$. For any unknown vectors $x^{(1)}, ..., x^{(S_h)} \in \rsketchn$ with $\ell_2$-norm at most $O(1)$, if we draw $R_1, ..., R_N \sim \distribution(\dsketch)$ and receive $S_h$ noisy samples $y^{(k)} \coloneqq [R_1 R_2 \cdots R_N] x^{(k)} + z^{(k)}_1 + z^{(k)}_\infty$ where each $z^{(k)}_1 \in \rsketch$ is noise with $\norm{z^{(k)}_1}{1} \le O(\sqrt{\dsketch})$ (independent of our random matrices) and each $z^{(k)}_\infty \in \rsketch$ is noise with $\norm{z^{(k)}_\infty}{\infty} \le \epsilon_h$ (also independent of our random matrices), then there is an algorithm which takes as input $h$, $y^{(1)}$, ..., $y^{(S_h)}$, runs in time $\poly(S_h, \dsketch)$, and with high probability outputs $\hat{R}_1, ..., \hat{R}_N, \hat{x}^{(1)}, ..., \hat{x}^{(S_h)}$ satisfying the following for some permutation $\pi$ of $[N]$:
  \begin{itemize}[nosep]
    \item for every $i \in [N]$ and $j \in [\dsketch]$, if there exists $k^\star \in [S_h]$ such that $\abs{x^{(k^\star)}_{(i-1)\dsketch + j}} \ge \epsilon_{h+1}$ then the $j^{th}$ column of $\hat{R}_{\pi(i)}$ is $0.2 \dsketch$-close in Hamming distance from the $j^{th}$ column of $R_i$.
    \item for every $k \in [S_h], i \in [N], j \in [\dsketch]$, $\abs{x^{(k)}_{(\pi(i)-1)\dsketch + j} - x^{(k)}_{(i-1)\dsketch + j}} \le \epsilon_{h+1}$.
  \end{itemize}
\end{restatable}

\subsection{Recursable Dictionary Learning Implies Network Learnability}

We want to use Theorem~\ref{thm:dictionary-learning} to learn a teacher modular deep network, but we need to first address an important issue. So far, we have not specified how the deep modular network decides upon its input-dependent communication graph. As a result, the derivation of the communication graph cannot be learned (it's possibly an uncomputable function!). When the communication graph is a fixed tree (always the same arrangement of objects but with different attributes), we can learn it. Note that any fixed communication graph can be expanded into a tree; doing so is equivalent to not re-using computation. Regardless of the communication graph, we can learn the input/output behavior of each module regardless of whether the communication graph is fixed.
\begin{restatable}{theorem}{applieddictionarylearning}
\label{thm:applied-dictionary-learning}
  If the teacher deep modular network has constant depth, then any module $M^\star$ which satisfies the following two properties:
  \begin{itemize}[nosep]
    \item (Robust Module Learnability) The module is learnable from $\left(\alpha = \poly(N)\right)$ input/output training pairs which have been corrupted by $\ell_\infty$ error at most a constant $\epsilon > 0$.
    \item (Sufficient Weight) In a $\left(\beta = \frac{1}{\poly(N)}\right)$-fraction of the inputs, the module produces an object and all of the input objects to that object have effective weight at least $w$.
  \end{itemize}
  can, with high probability, be learned from $\poly(N)$ overall sketches of dimension $\poly(1 / w, 1 / \epsilon) \log^2 N$.
  
  Suppose we additionally know that the communication graph is a fixed tree. We can identify the sub-graph of objects which each have effective weight $w$ in a $\left(\beta = \frac{1}{\poly(N)}\right)$-fraction of the inputs.
\end{restatable}

Theorem~\ref{thm:applied-dictionary-learning} is proved in \splitter{Appendix~\ref{subsec:dict_learning_appendix}}{the supplementary material}. The main idea is to just repeatedly apply our recursable dictionary learning algorithm and keep track of which vectors correspond to sketches and which vectors are garbage.

\subsection{Recursable Dictionary Learning: Proof Outline}
The main idea of our recursable dictionary learning algorithm is the following. The algorithm proceeds by iteratively examining each of the $\dsketch/b$ blocks of each of the $S_h$ received samples. For the $\ell$th block of the $i$th sample, denoted by $y^{(i)}[(\ell-1)b + 1: ~ \ell b + 1]$, it tries to determine whether it is dominated by the contribution of a single $(\sigma_s, \sigma_c, \sigma_m)$ (and is not composed of the superposition of more than one of these). To do so, it first normalizes this block by its $\ell_1$-norm and checks if the result is close to a point on the Boolean hypercube in $\ell_{\infty}$ distance. If so, it rounds (the $\ell_1$-normalized version of) this block to the hypercube; we here denote the resulting rounded vector by $\overline{y}^{(i)}[(\ell-1)b + 1: ~ \ell b + 1]$. We then count the number of blocks $\ell' \in [\dsketch/b]$ whose rounding $\overline{y}^{(i)}[(\ell'-1)b + 1: ~ \ell' b + 1]$ is close in Hamming distance to $\overline{y}^{(i)}[(\ell-1)b + 1: ~ \ell b + 1]$. If there are at least $0.8 q \dsketch$ of these, we add the rounded block $\overline{y}^{(i)}[(\ell-1)b + 1: ~ \ell b + 1]$ to our collection (if it's not close to an already added vector). We also cluster the added blocks in terms of their matrix signatures $\sigma_m$ in order to associate each of them to one of the matrices $R_1, \dots, R_N$. Using the matrix signatures as well as the column signatures $\{\sigma_c\}$ allows us to recover all the essential columns of the original matrices. The absolute values of the $x$ coordinates can also be inferred by adding the $\ell_1$-norm of a block when adding its rounding to the collection. The signs of the $x$ coordinates can also be inferred by first inferring the true sign of the block and comparing it to the sign of the vector whose rounding was added to the collection.


\subsection{The Block-Sparse Distribution $\distribution$}

We are now ready to define our distribution $\distribution$ on random (square) matrices $R$. It has the following parameters:
\begin{itemize}[nosep]
    \item The block size $b \in \mathbb{Z}^+$ which must be a multiple of $3$ and at least $3 \max (\lceil \log_2 N \rceil, \lceil \log_2 \dsketch \rceil + 3)$.
    \item The block sampling probability $q \in [0, 1]$; this is the probability that a block is nonzero.
    \item The number of rows/columns $\dsketch \in \mathbb{Z}^+$. It must be a multiple of $b$, as each column is made out of blocks.
\end{itemize}

Each column of our matrix is split into $\dsketch / b$ blocks of size $b$. Each block contains three sub-blocks: a random string, the matrix signature, and the column signature. To specify the column signature, we define an encoding procedure $\Enc$ which maps two bits $b_m, b_s$ and the column index into a $(b/3)$-length sequence. $\Enc: \{\pm 1\}^2 \times [\dsketch] \to \{\pm \frac{1}{\sqrt{\dsketch q}}\}^{b/3}$, which is presented as Algorithm~\ref{alg:encoding}. These two bits encode the parity of the other two sub-blocks, which will aid our dictionary learning procedure in recovery of the correct signs. The sampling algorithm which describes our distribution is presented as Algorithm~\ref{alg:block-random-distribution}.
\begin{figure}
\centering
\splitter{
  { 
  \scalebox{0.7}{
  \input{fig-block-random-matrix.tex}
  } 
  } 
} {
  { 
  \scalebox{0.45}{
  \input{fig-block-random-matrix.tex}
  } 
  } 
}
\caption{Example block-random matrix from distribution $\distribution(b, q, \dsketch)$. $\Diamond \coloneqq \dsketch / b$ denotes the number of blocks.}
\label{fig:block-random-matrix}
\end{figure}
\begin{algorithm}[h]
\caption{$\Enc(j, b_m, b_s)$}
\begin{algorithmic}[1]\label{alg:encoding}
  \STATE {\bf Input:} Column index $j \in [\dsketch]$, two bits $b_m, b_s \in \{\pm 1\}$, . \\
  \STATE{\bf Output:} A vector $\sigma_c \in \{\pm \frac{1}{\sqrt{\dsketch q}}\}^{b/3}$. \\
  \item[] 
  \STATE Set $\sigma_c$ to be the (zero-one) binary representation of $j-1$ (a $\lceil \log_2 \dsketch \rceil$-dimensional vector). \\
  \STATE Replace each zero in $\sigma_c$ with $-1$ and each one in $\sigma$ with $+1$. \\
  \STATE Prepend $\sigma_c$ with a $+1$, making it a $\left(\lceil \log_2 \dsketch \rceil + 1\right)$-dimensional vector. \\
  \STATE Append $\sigma_c$ with $(b_m, b_s)$, making it a $\left(\lceil \log_2 \dsketch \rceil + 3\right)$-dimensional vector.\\
  \STATE Append $\sigma_c$ with enough $+1$ entries to make it a $(b/3)$-dimensional vector. Note that $b \ge 3 \left(\lceil \log_2 \dsketch \rceil + 3\right)$. \\
  \STATE Divide each entry of $\sigma_c$ by $\sqrt{\dsketch q}$.
  \RETURN $\sigma_c$.
\end{algorithmic}
\end{algorithm}

\begin{algorithm}[H]
\caption{$\distribution(b, q, \dsketch)$}
\begin{algorithmic}[1]\label{alg:block-random-distribution}
  \STATE {\bf Input:} Block size $b \in \mathbb{Z}^+$, block sampling probability $q \in [0, 1]$, number of rows/columns $\dsketch$. \\
  \STATE{\bf Output:} A matrix $R \in \mathbb{R}^{\dsketch \times \dsketch}$. \\
  \item[] 
  \STATE Initialize $R$ to be the all-zeros $\mathbb{R}^{\dsketch \times \dsketch}$ matrix. \\
  \STATE Sample a ``matrix signature'' vector $\sigma_m$ from the uniform distribution over $\{\pm \frac{1}{\sqrt{\dsketch q}}\}^{b/3}$. \\
  \FOR{column $j \in [\dsketch]$}
    \STATE Sample a ``random string'' vector $\sigma_{s, j}$ from the uniform distribution over $\{\pm \frac{1}{\sqrt{\dsketch q}}\}^{b/3}$. \\
    \FOR{block $i \in [\dsketch/b]$}
      \STATE Sample three coin flips $f_{m, i, j}, f_{s, i, j}, f_{c, i, j}$ as independent Rademacher random variables (i.e. uniform over $\{\pm 1\}$). \\
      \STATE Compute two bits $f'_{m, i, j} \coloneqq f_{m, i, j} \cdot \text{sign}(\sigma_m[1])$ and $f'_{s, i, j} \coloneqq f_{s, i, j}s \cdot \text{sign}(\sigma_{s, j}[1])$. \\
      \STATE Compute a ``column signature'' vector $\sigma_{c, i, j} \coloneqq \Enc(j, f'_{m, i, j}, f'_{s, i, j})$. \\
      \STATE Sample $\eta_{i, j}$ to be a (zero-one) Bernoulli random variable which is one with probability $q$. \\
      \IF{$\eta_{i, j} = 1$}
        \STATE Set $R[b(i-1)+1:bi+1,j]$ to be $f_{s, i, j} \cdot \sigma_{s, j}$ concatenated with $f_{c, i, j} \cdot \sigma_{c, i, j}$ concatenated with $f_{m, i, j} \cdot \sigma_m$. \\
      \ENDIF
    \ENDFOR
  \ENDFOR
  \RETURN $R$.
\end{algorithmic}
\end{algorithm}

\splitter{
\section{Conclusion}

In this paper, we presented a sketching mechanism which captures several natural properties. Two potential areas for future work are our proposed use cases for this mechanism: adding a sketch repository to an existing modular network and using it to find new concepts, and using our sketches to learn a teacher network.
}{}

\bibliographystyle{icml2019}

\splitter{
\clearpage

\appendix

\section{Technical Machinery for Random Matrices}
\label{apx:tm}

In this appendix, we present several technical lemmas which will be useful for analyzing the random matrices that we use in our sketches.

\subsection{Notation}

\begin{definition}
  Suppose we have two fixed vectors $x, y \in \rsketch$ and random matrices $R_1, \ldots, R_k$ drawn independently from distributions $\distribution_1, ..., \distribution_k$. We have a random variable $Z$ which is a function of $x$, $y$, and $R_1, \ldots, R_k$. $Z$ is a $\delta$-nice noise variable with respect to $D$ if the following two properties hold:
  \begin{enumerate}
    \item (Centered Property)
          $\mathbb{E}_{R_1 \sim \distribution_1, \ldots R_k \sim \distribution_k}[Z] = 0$.
    \item ($\delta$-bounded Property)
          With high probability (over $R_1 \sim \distribution_1, \ldots R_k \sim \distribution_k$),
          $|Z| \le \delta \cdot \norm{x}{2} \cdot \norm{y}{2}$.
  \end{enumerate}
\end{definition}

\begin{definition}\label{def:desync}
  We say that a distribution $\distribution$ has the $\alpha_D$-Leaky $\delta_D$-Desynchronization Property if the following is true. If $R \sim \distribution$, then the random variable $Z = \inner{Rx}{y} - \alpha_D \inner{x}{y}$ is a $\delta_D$-nice noise variable. If $\alpha_D = 0$ then we simply refer to this as the $\delta_D$-Desynchronization Property.
\end{definition}

\begin{definition}\label{def:isometry}
  We say that a distribution $\distribution$ has the $\alpha_I$-scaled $\delta_I$-Isometry Property if the following is true. If $R \sim \distribution$ and $x = y$, then the random variable $Z = \inner{Rx}{Rx} - \alpha_I \inner{x}{x}$ is a $\delta_I$-nice noise variable. If $\alpha_I = 1$ then we simply refer to this as the $\delta_I$-Isometry Property.
\end{definition}

Since our error bounds scale with the $\ell_2$ norm of vectors, it will be useful to use isometry to bound the latter using the following technical lemma.
\begin{lemma}\label{lem:length-isometry}
  If $\distribution$ satisfies the $\alpha_I$-scaled $\delta_I$-Isometry Property, then with high probability:
  \[
    \norm{Rx}{2}^2 \le (\alpha_I + \delta_I) \norm{x}{2}^2
  \]
\end{lemma}

\begin{proof}
  Consider the random variable:
  \[
    Z = \inner{Rx}{Rx} - \alpha_I \inner{x}{x}
  \]
  
  Since $\distribution$ satisfies the $\alpha_I$-scaled $\delta_I$-Isometry Property, $Z$ is a $\delta_I$-nice random variable. Hence with high probability:
  \begin{align*}
    \inner{Rx}{Rx} - \alpha_I \inner{x}{x} &\le
      \delta_I \norm{x}{2}^2 \\
    \inner{Rx}{Rx} &\le
      (\alpha_I + \delta_I) \norm{x}{2}^2
  \end{align*}
  
  This completes the proof.
\end{proof}

While isometry compares $\inner{Rx}{Rx}$ and $\inner{x}{x}$, sometimes we are actually interested the more general comparison between $\inner{Rx}{Ry}$ and $\inner{x}{y}$. The following technical lemma lets get to the general case at the cost of a constant blowup in the noise. 
\begin{lemma}\label{lem:advanced-isometry}
  If $\distribution$ satisfies the $\alpha_I$-scaled $\delta_I$-Isometry Property, then the random variable $Z = \inner{Rx}{Ry} - \alpha_I \inner{x}{y}$ is a $(3\delta_I)$-nice noise variable.
\end{lemma}

\begin{proof}
  This proof revolves around the scalar fact that $(a + b)^2 = a^2 + 2ab + b^2$, so from squaring we can get to products.
  
  This claim is trivially true if $x$ or $y$ is the zero vector. Hence without loss of generality, we can scale $x$ and $y$ to unit vectors, since being a nice noise variable is scale invariant.
  
  We invoke the $\alpha_I$-scaled $\delta_I$-bounded Isometry Property three times, on $x_1 = x + y$, $x_2 = x$, and $x_3 = y$. This implies that the following three noise variables are $\delta_I$-nice:
  \begin{align*}
    Z_1 &= \inner{R(x + y)}{R(x + y)} - \alpha_I \inner{x + y}{x + y} \\
    Z_2 &= \inner{Rx}{Rx} - \alpha_I \inner{x}{x} \\
    Z_3 &= \inner{Ry}{Ry} - \alpha_I \inner{y}{y}
  \end{align*}
  
  We now relate our four noise variables of interest.
  \begin{align*}
    Z_1
      &= \left[
           \inner{Rx}{Rx} + 2\inner{Rx}{Ry} + \inner{Ry}{Ry}
         \right]
      \\ &\phantom{ = }
         - \alpha_I \left[
           \inner{x}{x} + 2\inner{x}{y} + \inner{y}{y}
         \right] \\
      &= Z_2 + 2 Z + Z_3 \\
    Z &= \frac12 Z_1 - \frac12 Z_2 - \frac12 Z_3
  \end{align*}
  Hence, $Z$ is centered. We can bound it by analyzing the $\ell_2$ lengths of the vectors that we applied isometry to. With high probability:
  \begin{align*}
    |Z_1| &\le \delta_I \norm{x + y}{2}^2 \\
          &\le \delta_I \left(
                          \norm{x}{2} + \norm{y}{2}
                        \right)^2 \\
          &\le 4 \delta_I \\
    |Z_2| &\le \delta_I \norm{x}{2}^2 \\
          &\le \delta_I \\
    |Z_3| &\le \delta_I \norm{y}{2}^2 \\
          &\le \delta_I
  \end{align*}
  Hence $Z$ is also $(3\delta_I)$-bounded, making it a $(3\delta_I)$-noise variable, completing the proof.
\end{proof}

\subsection{Transparent Random Matrices}

For our sketch, we sometimes want to both rotate a vector and pass it through as-is. For these cases, the distribution $\transparentdist$ is useful and defined as follows. We draw $R$ from $D$ and return $\bar{R} = \ptransparentx{R}$. This subsection explains how the desynchronization and isometry of $\distribution$ carry over to $\transparentdist$. This technical lemma shows how to get desynchronization.
\begin{lemma}\label{lem:transparent-desync}
  If $\distribution$ satisfies the $\delta_D$-Desynchronization Property, then $\transparentdist$ satisfies the $(1 / 2)$-leaky $(\delta_D / 2)$-Desynchronization Property.
\end{lemma}

\begin{proof}
  We want to show that if $\bar{R}$ is drawn from $\transparentdist$, then the following random variable $Z$ is a $(\delta_D / 2)$-nice noise variable.
  \begin{align*}
    Z &= \inner{\bar{R}x}{y}
         - \frac12 \inner{x}{y} \\
      &= x^T \ptransparentx{R^T} y
         - \frac12 \inner{x}{y} \\
      &= \frac12 x^T y
         + \frac12 x^T R^T y
         - \frac12 \inner{x}{y} \\
      &= \frac12 \underbrace{\left[
           x^T R^T y
         \right]}_{\text{Let this be $Z'$.}}
  \end{align*}
  
  Since $\distribution$ satisfies the $\delta_D$-Desynchronization Property, we know $Z'$ is a $\delta_D$-nice noise variable. Our variable $Z$ is just a scaled version of $Z'$, making it a $(\delta_D / 2)$-nice noise variable. This completes the proof.
\end{proof}

This next technical lemma shows how to get isometry.
\begin{lemma}\label{lem:transparent-isometry}
  If $\distribution$ satisfies the $\delta_D$-Desynchronization Property and the $\delta_I$-Isometry Property, then $\transparentdist$ satisfies the $\frac12$-scaled $(\frac12 \delta_D + \frac14 \delta_I)$-Isometry Property.
\end{lemma}

\begin{proof}
  We want to show that if $\bar{R}$ is drawn from $\transparentdist$, and $x = y$, then the following random variable $Z$ is a
  $\left(\delta_D + \frac12 \delta_I\right)$-nice noise variable:
  \begin{align*}
    Z &= \inner{\bar{R}x}{\bar{R}x} - \frac12 \inner{x}{x} \\
      &= x^T \ptransparentx{R^T} \ptransparentx{R} x - \frac12 x^T x \\
      &= \frac14
         \left(
           x^T x + x^T R x + x^T R^T x + x^T R^T R x
         \right)
         - \frac12 x^T x \\
      &= \frac12 \underbrace{\inner{Rx}{x}}_{\text{Let this be $Z_1$.}}
         + \frac14 \underbrace{\left(
           \inner{Rx}{Rx} - \inner{x}{x}
         \right)}_{\text{Let this be $Z_2$.}}
  \end{align*}
  Since $\distribution$ satisfies the $\delta_D$-Desynchronization Property, we know $Z_1$ is a $\delta_D$-nice noise variable. Since $\distribution$ satisfies the $\delta_I$-Isometry Property, we know $Z_2$ is a $\delta_I$-nice noise variable.
  
  This makes our variable $Z$ a $(\frac12 \delta_D + \frac14 \delta_I)$-nice noise variable, as desired.
\end{proof}

\subsection{Products of Random Matrices}

For our sketch, we wind up multiplying matrices. In this subsection, we try to understand the properties of the resulting product. The distribution $\productdist$ is as follows. We independently draw $R_1 \sim \distribution_1, R_2 \sim \distribution_2$ and then return:
\[
  \tilde{R} = R_2 R_1
\].
This technical lemma says that the resulting product is desynchronizing.
\begin{lemma}\label{lem:product-desync}
  If $\distribution_1$ satisfies the $\delta_{D, 1}$-Desynchronization Property, $\distribution_1$ satisfies the $\delta_{I, 1}$-Isometry Property, and $\distribution_2$ satisfies the $\delta_{D, 2}$-leaky $\alpha_{D, 2}$-Desynchronization Property, then $\productdist$ satisfies the $\left[ \delta_{D, 1} + \delta_{D, 2} \sqrt{1 + \delta_{I, 1}} \right]$-Desynchronization Property.
\end{lemma}

\begin{proof}
  We want to show that if $\tilde{R}$ is drawn from $\productdist$, then the following random variable $Z$ is a $\left[ \delta_{D, 1} + \delta_{D, 2} \sqrt{1 + \delta_{I, 1}} \right]$-nice noise variable.
  \begin{align*}
    Z &= \inner{\tilde{R}x}{y} - \inner{x}{y} \\
      &= \inner{R_2 R_1 x}{y} - \inner{x}{y}
  \end{align*}
  
  We now define some useful (noise) variables.
  \begin{align*}
    Z_1 &\coloneqq \inner{R_1 x}{y} - \inner{x}{y} \\
    Z_2 &\coloneqq \inner{R_2 R_1 x}{y} - \inner{R_1 x}{y} \\
    Z &= Z_1 + Z_2
  \end{align*}
  
  From our assumptions about $\distribution_1$ and $\distribution_2$, we know that $Z_1$ is a $\delta_{D, 1}$-nice noise variable and $Z_2$ is a $\delta_{D, 2}$-nice noise variable. By Lemma~\ref{lem:length-isometry} we know that with high probability, $\norm{R_1 x}{2}^2 \le 1 + \delta_{I, 1}$. Hence, with high probability:
  \begin{align*}
    \abs{Z_1} &\le \delta_{D, 1} \norm{x}{2} \norm{y}{2} \\
    \abs{Z_2} &\le \delta_{D, 2}
                   \sqrt{1 + \delta{I, 1}} \norm{x}{2} \norm{y}{2} \\
    \abs{Z} &\le \abs{Z_1} + \abs{Z_2} \\
            &\le \left[
                   \delta_{D, 1} + \delta_{D, 2} \sqrt{1 + \delta_{I, 1}}
                 \right]  \norm{x}{2} \norm{y}{2}
  \end{align*}
  
  Hence $Z$ is a $\left[ \delta_{D, 1} + \delta_{D, 2} \sqrt{1 + \delta_{I, 1}} \right]$-nice noise variable, completing the proof.
\end{proof}

This technical lemma says that the resulting product is isometric.
\begin{lemma}\label{lem:product-isometry}
  If $\distribution_1$ satisfies the $\alpha_{I, 1}$-scaled $\delta_{I, 1}$-Isometry Property and $\distribution_2$ satisfies the $\alpha_{I, 2}$-scaled $\delta_{I, 2}$-Isometry Property, then $\productdist$ satisfies the $\left(\alpha_{I, 1} \alpha_{I, 2}\right)$-scaled $\left[ \delta_{I, 2} (1 + \delta_{I, 1}) + \alpha_{I, 2} \delta_{I, 1} \right]$-bounded Isometry Property.
\end{lemma}

\begin{proof}
  We want to show that if $\tilde{R}$ is drawn from $\productdist$, and $x = y$, then the following random variable $Z$ is a $\left[ \delta_{I, 2} (1 + \delta_{I, 1}) + \alpha_{I, 2} \delta_{I, 1} \right]$-nice noise variable.
  \begin{align*}
    Z &\coloneqq \inner{\tilde{R}x}{\tilde{R}x}
         - \alpha_{I, 1} \alpha_{I, 2} \inner{x}{x} \\
      &= x^T \tilde{R}^T \tilde{R} x - \alpha_{I, 1} \alpha_{I, 2} x^T x \\
      &= x^T R_1^T R_2^T R_2 R_1 x - \alpha_{I, 1} \alpha_{I, 2} x^T x
  \end{align*}
  
  We now define some useful (noise) variables.
  \begin{align*}
    x_0 &\coloneqq x \\
    x_1 &\coloneqq R_1 x_0 \\
    x_2 &\coloneqq R_2 x_1 \\
    Z_1 &\coloneqq \inner{R_1 x_0}{R_1 x_0} - \alpha_{I, 1} \inner{x_0}{x_0} \\
        &= \norm{x_1}{2}^2 - \alpha_{I, 1} \norm{x_0}{2}^2 \\
    Z_2 &\coloneqq \inner{R_2 x_1}{R_2 X_1} - \alpha_{I, 2} \inner{x_1}{x_1} \\
        &= \norm{x_2}{2}^2 - \alpha_{I, 2} \norm{x_1}{2}^2
  \end{align*}
  
  $Z$ is just a weighted sum of our random variables:
  \begin{align*}
    Z_2 + \alpha_{I, 2} Z_1
      &= \norm{x_2}{2}^2 - \alpha_{I, 2} \norm{x_1}{2}^2
      \\ &\phantom{ = }
         + \alpha_{I, 2} \norm{x_1}{2}^2 - \alpha_{I, 1} \alpha_{I, 2} \norm{x_0}{2}^2 \\
      &= \norm{x_2}{2}^2 - \alpha_{I, 1} \alpha_{I, 2} \norm{x_0}{2}^2 \\
      &= Z
  \end{align*}
  
  Since $\distribution_1$ satisfies the $\alpha_{I, 1}$-scaled $\delta_{I, 1}$-Isometry Property we know that $Z_1$ is a $\delta_{I, 1}$-nice noise variable. Similarly, we know $Z_2$ is a $\delta_{I, 2}$-nice noise variable. As a result, $Z$ is centered; it remains to bound it. By Lemma~\ref{lem:length-isometry}, we know that with high probability $\norm{x_1}{2}^2 \le 1 + \delta_{I, 1}$. Hence with high probability:
  \begin{align*}
    \abs{Z_1} &\le \delta_{I, 1} \norm{x_0}{2}^2 \\
    \abs{Z_2} &\le \delta_{I, 2} (1 + \delta_{I, 1}) \norm{x_0}{2}^2 \\
    \abs{Z} &\le \abs{Z_2} + \alpha_{I, 2} \abs{Z_1} \\
            &\le \delta_{I, 2} (1 + \delta_{I, 1}) +
                 \alpha_{I, 2} \delta_{I, 1}
  \end{align*}
  Hence $Z$ is a $\left[ \delta_{I, 2} (1 + \delta_{I, 1}) + \alpha_{I, 2} \delta_{I, 1}\right]$-nice noise variable, completing the proof.
\end{proof}

\subsection{Applying Different Matrices}

For the analysis of our sketch, we sometimes compare objects that have been sketched using different matrices. The following technical lemma is a two matrix variant of desynchronization for this situation.
\begin{lemma}\label{lem:diff-desync}
  Suppose that we have a distribution $\distribution$ which satisfies the $\alpha_I$-scaled $\delta_I$-Isometry Property and the $\alpha_D$-leaky $\delta_D$-Desynchronization Property. Suppose we independently draw $R_1 \sim \distribution$ and $R_2 \sim \distribution$. Then the random variable
  \[
    Z = \inner{R_1 x}{R_2 y} - \alpha_D^2 \inner{x}{y}
  \]
  is a $(\delta_D \sqrt{\alpha_I + \delta_I} + \alpha_D \delta_D)$-nice noise variable.
\end{lemma}

\begin{proof}
  We define useful (noise) variables $Z_1$ and $Z_2$.
  \begin{align*}
    x' &\coloneqq R_1 x \\
    Z_1 &\coloneqq \inner{x'}{R_2 y} - \alpha_D \inner{x'}{y} \\
    Z_2 &\coloneqq \inner{R_1 x}{y} - \alpha_D \inner{x}{y} \\
    Z &= Z_1 + \alpha_D Z_2
  \end{align*}
  
  Since $\distribution$ satisfies the $\alpha_D$-leaky $\delta_D$-Desynchronization Property, we know that $Z_1$ and $Z_2$ are $\delta_D$-nice noise variables. Since $D$ satisfies the $\alpha_I$-scaled $\delta_I$-Isometry Property, we can apply Lemma~\ref{lem:length-isometry} to show that with high probability, $\norm{R_1 x}{2}^2 \le (\alpha_I + \delta_I)$. Hence with high probability:
  \begin{align*}
    \abs{Z_1} &\le \delta_D \sqrt{\alpha_I + \delta_I} \\
    \abs{Z_2} &\le \delta_D \\
    \abs{Z} &\le \abs{Z_1} + \alpha_D \abs{Z_2} \\
             &\le \delta_D \sqrt{\alpha_I + \delta_I} + \alpha_D \delta_D
  \end{align*}
  
  Hence $Z$ is a $(\delta_D \sqrt{\alpha_I + \delta_I} + \alpha_D \delta_D)$-nice noise variable, as desired. This completes the proof.
\end{proof}
\section{Properties of the Block-Random Distribution}\label{apx:block-random}

In this section, we prove that our block-random distribution over random matrices satisfies the isometry and desynchronization properties required by our other results. In particular, our goal is to prove the following two theorems:

\begin{theorem}
\label{thm:block-random-isometry}
  Suppose that $q$ is $\Omega\left(\frac{\sqrt{b \log N}}{\sqrt{\dsketch}}\right)$ and $\dsketch$ is $\poly(N)$. Then there exists a $\delta = O\left(\frac{\sqrt{b \log N}}{\sqrt{\dsketch}}\right)$ such that the block-random distribution $\distribution(b, q, \dsketch)$ satisfies the $\delta$-Isometry Property.
\end{theorem}

\begin{theorem}
\label{thm:block-random-desync}
  Suppose that $q$ is $\Omega\left(\frac{\sqrt{b \log N}}{\sqrt{\dsketch}}\right)$. Then there exists a $\delta = O\left(\frac{\sqrt{b \log N}}{\sqrt{\dsketch}}\right)$ such that the block-random distribution $\distribution(b, q, \dsketch)$ satisfies the $O(\frac{\log N}{\sqrt{\dsketch}})$-Desynchronization Property.
\end{theorem}

Note that instantiating the theorems with the minimum necessary $b$ gives us good error bounds.

\begin{corollary}
\label{cor:block-random-isometry-desync}
  Suppose that $b = 3 \max \left( \lceil \log_2 N \rceil, \lceil \log_2 \dsketch \rceil + 3\right)$, $q = \left\lceil \frac{\sqrt{(\log N + \log \dsketch) \log N}}{\sqrt{\dsketch}} \right\rceil$, and $\dsketch$ is $\poly(N)$. Then the block-random distribution $\distribution(b, q, \dsketch)$ satisfies the $\Otilde\left(\frac{1}{\sqrt{\dsketch}}\right)$-Isometry Property and the $\Otilde\left(\frac{1}{\sqrt{\dsketch}}\right)$-Desynchronization Property.
\end{corollary}

\subsection{Isometry}

In this subsection, we will prove Theorem~\ref{thm:block-random-isometry}. Our proof essentially follows the first analysis presented by Kane and Nelson for their sparse Johnson-Lindenstrauss Transform ~\cite{Kane:2014:SJT:2578041.2559902}. Our desired result differs from theirs in several ways, which complicates our version of the proof:
\begin{itemize}[nosep]
  \item Their random matrices do not have block-level patterns like ours do. We handle this by using the small size of our blocks.
  \item Their random matrices have a fixed number of non-zeros per column, but the non-zeros within one of our columns are determined by independent Bernoulli random variables: $\eta_{i, j}$. We handle this by applying a standard concentration bound to the number of non-zeros.
  \item Some of our sub-block patterns depend on the $\pm 1$ coin flips for other sub-blocks, ruining independence. We handle this by breaking our matrix into three sub-matrices to be analyzed separately; see Figure~\ref{fig:block-random-submatrix}.
\end{itemize}

For the sake of completeness, we reproduce the entire proof here. The high-level proof plan is to use Hanson-Wright inequality~\cite{hanson1971bound} to bound the $\ell^{th}$ moment of the noise, and then use a Markov's inequality to convert the result into a high probability bound on the absolute value of the noise.

Suppose we draw a random matrix $R \in \mathbb{R}^{\dsketch \times \dsketch}$ from our distribution $\distribution(b, q, \dsketch)$. We split $R$ into three sub-matrices: $R_s \in \mathbb{R}^{\dsketch/3 \times \dsketch}$ containing the random string sub-blocks, $R_m \in \mathbb{R}^{\dsketch/3 \times \dsketch}$ containing the matrix signature sub-blocks, and $R_c \in \mathbb{R}^{\dsketch/3 \times \dsketch}$ containing the column signature sub-blocks.

\begin{figure}
\centering
{ 
\scalebox{0.7}{
\newcommand{\m}{\Diamond}
\(
\left[
\begin{array}{r@{}l|r@{}l|r@{}l|r@{}l|c|r@{}l}
  \rowcolor{green!20}
  f_{s, 1, 1} &\cdot \sigma_{s, 1}    & f_{s, 1, 2} &\cdot \sigma_{s, 2}    & f_{s, 1, 3}  &\cdot \sigma_{s, 3}     &              &\vec{0}                 & \cdots & f_{s, 1, \dsketch} &\cdot \sigma_{s, \dsketch}    \\
  f_{c, 1, 1} &\cdot \sigma_{c, 1, 1} & f_{c, 1, 2} &\cdot \sigma_{c, 1, 2} & f_{c, 1, 3}  &\cdot \sigma_{c, 1, 3}  &              &\vec{0}                 & \cdots & f_{c, 1, \dsketch} &\cdot \sigma_{c, 1, \dsketch} \\
  f_{m, 1, 1} &\cdot \sigma_m         & f_{m, 1, 2} &\cdot \sigma_m         & f_{m, 1, 3}  &\cdot \sigma_m          &              &\vec{0}                 & \cdots & f_{m, 1, \dsketch} &\cdot \sigma_m                \\
  \hline
  \rowcolor{green!20}
              &\vec{0}                & f_{s, 2, 2} &\cdot \sigma_{s, 2}    &              &\vec{0}                 &              &\vec{0}                 & \cdots &                    &\vec{0} \\
              &\vec{0}                & f_{c, 2, 2} &\cdot \sigma_{c, 2, 2} &              &\vec{0}                 &              &\vec{0}                 & \cdots &                    &\vec{0} \\
              &\vec{0}                & f_{m, 2, 2} &\cdot \sigma_m         &              &\vec{0}                 &              &\vec{0}                 & \cdots &                    &\vec{0} \\
  \hline
  \rowcolor{green!20}
  f_{s, 3, 1} &\cdot \sigma_{s, 1}    & f_{s, 3, 2} &\cdot \sigma_{s, 2}    & f_{s, 3, 3}  &\cdot \sigma_{s, 3}     & f_{s, 3, 4}  &\cdot \sigma_{s, 4}     & \cdots &                    &\vec{0} \\
  f_{c, 3, 1} &\cdot \sigma_{c, 3, 1} & f_{c, 3, 2} &\cdot \sigma_{c, 3, 2} & f_{c, 3, 3}  &\cdot \sigma_{c, 3, 3}  & f_{c, 3, 4}  &\cdot \sigma_{c, 3, 4}  & \cdots &                    &\vec{0} \\
  f_{m, 3, 1} &\cdot \sigma_m         & f_{m, 3, 2} &\cdot \sigma_m         & f_{m, 3, 3}  &\cdot \sigma_m          & f_{m, 3, 4}  &\cdot \sigma_m          & \cdots &                    &\vec{0} \\
  \hline
  \rowcolor{green!20}
  f_{s, 4, 1} &\cdot \sigma_{s, 1}    & f_{s, 4, 2} &\cdot \sigma_{s, 2}    & f_{s, 4, 3}  &\cdot \sigma_{s, 3}     &              &\vec{0}                 & \cdots & f_{s, 4, \dsketch} &\cdot \sigma_{s, \dsketch} \\
  f_{c, 4, 1} &\cdot \sigma_{c, 4, 1} & f_{c, 4, 2} &\cdot \sigma_{c, 4, 2} & f_{c, 4, 3}  &\cdot \sigma_{c, 4, 3}  &              &\vec{0}                 & \cdots & f_{c, 4, \dsketch} &\cdot \sigma_{c, 4, \dsketch} \\
  f_{m, 4, 1} &\cdot \sigma_m         & f_{m, 4, 2} &\cdot \sigma_m         & f_{m, 4, 3}  &\cdot \sigma_m          &              &\vec{0}                 & \cdots & f_{m, 4, \dsketch} &\cdot \sigma_m \\
  \hline
              &\vdots                 &             &\vdots                 &              &\vdots                  &              &\vdots                  & \ddots &                    &\vdots \\
  \hline
  \rowcolor{green!20}
              &\vec{0}                &             &\vec{0}                & f_{s, \m, 3} &\cdot \sigma_{s, 3}     & f_{s, \m, 4} &\cdot \sigma_{s, 4}     & \cdots &                    &\vec{0} \\
              &\vec{0}                &             &\vec{0}                & f_{c, \m, 3} &\cdot \sigma_{c, \m, 3} & f_{c, \m, 4} &\cdot \sigma_{c, \m, 4} & \cdots &                    &\vec{0} \\
              &\vec{0}                &             &\vec{0}                & f_{m, \m, 3} &\cdot \sigma_m          & f_{m, \m, 4} &\cdot \sigma_m          & \cdots &                    &\vec{0} \\
\end{array}
\right]
\)
} 
} 
\caption{The proof of Theorem~\ref{thm:block-random-isometry} involves three sub-matrices of a random block-random matrix. The first of these sub-matrices, $R_s$, is highlighted here in green. It consists of all the random string sub-blocks. Just as in Figure~\ref{fig:block-random-matrix}, $\Diamond \coloneqq \dsketch / b$ is shorthand for the number of blocks.}
\label{fig:block-random-submatrix}
\end{figure}

We fix $\delta = O\left(\frac{\log N}{\sqrt{\dsketch}}\right)$. Consider any one of these sub-matrices, $R_\star$. We wish to show that for any unit vector $x \in \rsketch$, with high probability over $R \sim \distribution(b, q, \dsketch)$:
\[
  \norm{R_\star x}{2}^2 \in \left[\frac13 (1 - \delta), \frac13 (1 + \delta)\right]
\]

Since this would imply that with high probability (due to union-bounding over the three sub-matrices):
\begin{align*}
  \norm{R x}{2}^2 &= \sum_{\star \in \{s, c, m\}} \norm{R_\star x}{2}^2 \\
                  &\in \left[ 1 - \delta, 1 + \delta \right]
\end{align*}

We will use the variant of the Hanson-Wright inequality presented by Kane and Nelson~\cite{Kane:2014:SJT:2578041.2559902}.
\begin{theorem}[Hanson-Wright Inequality~\cite{hanson1971bound}]
\label{thm:hanson-wright}
  Let $z = (z_1, ..., z_n)$ be a vector of i.i.d. Rademacher $\pm 1$ random variables. For any symmetric $B \in \mathbb{R}^{n \times n}$ and $\ell \ge 2$,
  \[
    \E \abs{z^T B z - \trace(B)}^\ell \le C^\ell \cdot \max \left\{ \sqrt{\ell} \cdot \norm{B}{F}, \ell \cdot \norm{B}{2} \right\}^\ell
  \]
  for some universal constant $C > 0$ independent of $B, n, \ell$.
\end{theorem}
Note that in the above theorem statement, $\norm{B}{F}$ is the Frobenius norm of matrix $B$ (equal to the $\ell_2$ norm of the flattened matrix) while $\norm{B}{2}$ is the operator norm of matrix $B$ (equal to the longest length of $A$ applied to any unit vector).

Our immediate goal is now to somehow massage $\norm{R_\star x}{2}^2$ into the quadratic form $z^T B z$ that appears in the inequality. We begin by expanding out the former. Note that for simplicity of notation, we'll introduce some extra indices: we use $\sigma_{s, i, j}$ to denote $\sigma_{s, j}$ and $\sigma_{m, i, j}$ to denote $\sigma_m$ so that all three types of sub-block patterns can be indexed the same way. Also, we'll use $\Diamond \coloneqq \dsketch / b$ to denote the number of blocks.
\begin{align*}
  R_\star x
    &= \sum_{j=1}^{\dsketch}
       \left[ \begin{array}{c}
         \eta_{1, j} f_{\star, 1, j} \sigma_{\star, 1, j} \\
         \eta_{2, j} f_{\star, 2, j} \sigma_{\star, 2, j} \\
         \vdots \\
         \eta_{\Diamond, j} f_{\star, \Diamond, j} \sigma_{\star, \Diamond, j}
       \end{array} \right]
       x_j \\
  \norm{R_\star x}{2}^2
    &= x^T R^T R x \\
    &= \sum_{i=1}^{\Diamond} \sum_{j=1}^{\dsketch} \sum_{j'=1}^{\dsketch}
         \left( \eta_{i, j} f_{\star, i, j} \sigma_{\star, i, j} x_j \right)^T
         \left( \eta_{i, j'} f_{\star, i, j'} \sigma_{\star, i, j'} x_{j'} \right) \\
    &= \sum_{i=1}^{\Diamond} \sum_{j=1}^{\dsketch} \sum_{j'=1}^{\dsketch}
         \eta_{i, j} \eta_{i, j'} f_{\star, i, j} f_{\star, i, j'} x_j x_{j'} \inner{\sigma_{\star, i, j}}{\sigma_{\star, i, j'}}
\end{align*}

As before, this can also be expressed as a quadratic form in the $f$ variables. In particular, let the $\Diamond \dsketch \times \Diamond \dsketch$ block-diagonal matrix $B_\star$ is as follows. There are $\Diamond$ blocks $B_{\star, i}$, each $\dsketch \times \dsketch$ in shape. The $(j, j')^{th}$ entry of the $i^{th}$ block is $\eta_{i, j} \eta_{i, j'} x_j x_{j'} \inner{\sigma_{\star, i, j}}{\sigma_{\star, i, j'}}$. As a result, $\norm{R_\star x}{2}^2 = f_\star^T B_\star f_\star$, where $f_\star$ is the $\Diamond \dsketch$-dimensional vector $\left(f_{\star, 1, 1}, f_{\star, 1, 2}, ...\right)$.

It is important to note for the sake of the Hanson-Wright inequality that the matrix $B_\star$ is random, but it \emph{does not depend} on the $f$ variables. Of course, $B_c$ depends on the $f$ variables involved in $B_s$ and $B_m$ by construction, but not its own (and this is why we split our analysis over the three submatrices).

All that remains before we can effectively apply the Hanson-Wright inequality is to bound the trace, Frobenius norm, and operator norm of this matrix $B_\star$. We begin with the trace, which is a random variable depending on the zero-one Bernoulli random variables $\eta$.
\begin{align*}
  \trace(B_\star)
    &= \sum_{i=1}^\Diamond \sum_{j=1}^{\dsketch}
         \eta_{i, j}^2 x_j^2 \norm{\sigma_{\star, i, j}}{2}^2 \\
    &= \sum_{i=1}^\Diamond \sum_{j=1}^{\dsketch}
         \eta_{i, j} x_j^2 (b/3)(\frac{1}{\dsketch q}) \\
    &= \sum_{i=1}^\Diamond \sum_{j=1}^{\dsketch}
         \eta_{i, j} x_j^2 \frac{1}{3 \Diamond q}
\end{align*}

Let the random variable $X_{i, j} \coloneqq \eta_{i, j} x_j^2 \frac{1}{3 \Diamond q}$. Since $x$ was a unit vector, we know that the expected value of $\sum_i \sum_j X_{i, j} = \frac13$. Additionally, note that:
\begin{align*}
  \sum_{i=1}^\Diamond \sum_{j=1}^{\dsketch} \left(x_j^2 \frac{1}{3 \Diamond q} \right)^2
    &= \frac{1}{9 \Diamond q^2} \sum_{j=1}^{\dsketch} x_j^4
    \le \frac{1}{9 \Diamond}
\end{align*}

By Hoeffding's Inequality~\cite{hoeffding1963probability} we the probability that this sum (i.e. the trace) differs too much from its expectation:
\begin{align*}
  \Pr\left[ \abs{\trace(B_\star) - \frac13} \ge \frac{\delta}{6} \right]
    &\le \exp\left(\frac{-2(\delta/6)^2}{\sum_{i=1}^\Diamond \sum_{j=1}^{\dsketch} \left(x_j^2 \frac{1}{3 \Diamond q} \right)^2}\right) \\
    &\le \exp\left(\frac{-3 \Diamond(\delta)^2}{2}\right)
\end{align*}

Hence there exists a sufficiently large $\delta = O\left(\frac{\sqrt{\log N}}{\sqrt{\Diamond}}\right)$ so that we will have with high probability that the trace is within $\frac{\delta}{6}$ of $\frac13$.

Next, we want a high probability bound on the Frobenius norm of $B_\star$.
\begin{align*}
  \norm{B_\star}{F}^2
    &= \sum_{i=1}^\Diamond \sum_{j=1}^{\dsketch} \sum_{j'=1}^{\dsketch}
         \left( \eta_{i, j} \eta_{i, j'} x_j x_{j'} \inner{\sigma_{\star, i, j}}{\sigma_{\star, i, j'}} \right)^2 \\
    &= \sum_{i=1}^\Diamond \sum_{j=1}^{\dsketch} \sum_{j'=1}^{\dsketch}
         \eta_{i, j} \eta_{i, j'} x_j^2 x_{j'}^2 \inner{\sigma_{\star, i, j}}{\sigma_{\star, i, j'}}^2 \\
    &\le \sum_{i=1}^\Diamond \sum_{j=1}^{\dsketch} \sum_{j'=1}^{\dsketch}
         \eta_{i, j} \eta_{i, j'} x_j^2 x_{j'}^2 \left(\frac{1}{\dsketch q} b/3\right)^2 \\
    &= \frac{1}{9\Diamond^2 q^2} \sum_{j=1}^{\dsketch} \sum_{j'=1}^{\dsketch}
         x_j^2 x_{j'}^2 \sum_{i=1}^\Diamond \eta_{i, j} \eta_{i, j'}
\end{align*}

We want to show that for any two columns $j, j'$ there aren't too many matching non-zero blocks. The number of matching blocks is the sum of $\Diamond$ i.i.d. Bernoulli random variables with probability $q^2$. By a Hoeffding Bound~\cite{hoeffding1963probability}, we know that for all $t \ge 0$:
\[
  \Pr\left[\sum_{i=1}^{\Diamond} \eta_{i, j} \eta_{i, j'} \ge \Diamond q^2 + t \right] \le \exp \left( -2t^2 / \Diamond \right)
\]
So the desired claim is true with high probability (i.e. polynomially small in both $N$ and $\dsketch$) as long as we can choose $t$ to be on the order of $\frac{\sqrt{\log (N \dsketch)}}{\sqrt{\Diamond}}$. However, we want this to be $O(\Diamond q^2)$. Since by assumption $q$ is $\Omega\left(\frac{\sqrt{\log N}}{\sqrt{\Diamond}}\right)$, we actually just want it to be $O(\log N)$. This is true since we assumed $\dsketch$ was $\poly(N)$ and $\Diamond \ge 1$.

We can now safely union-bound over all pairs of columns and finish bounding the Frobenius norm (with high probability):
\begin{align*}
  \norm{B_\star}{F}^2
    &\le \frac{1}{9\Diamond^2 q^2} \underbrace{\sum_{j=1}^{\dsketch} \sum_{j'=1}^{\dsketch} x_j^2 x_{j'}^2}_{\norm{x}{2}^4 = 1} O(\Diamond q^2) \\
    &\le O\left(\frac{1}{\Diamond}\right)
\end{align*}

Finally, we want a high probability bound on the operator norm of $B_\star$. The operator norm is just its maximum eigenvalue, and its eigenvalues are the eigenvalues of its blocks. We can write each block as a rank-$(b/3)$ decomposition $\sum_{k=1}^{b/3} u_{i, k} u_{i, k}^T$ by breaking up the inner products by coordinate (each yielding a vector $u_{i, k}$). Specifically, the vector $u_{i, k} \in \mathbb{R}^{\dsketch}$ is defined by $(u_{i, k})_j = \eta_{i, j} x_j \sigma_{\star, i, j}[k]$.
\begin{align*}
  \norm{B_\star}{2}
    &= \max_i \norm{B_{\star, i}}{2} \\
    &= \max_i \norm{\sum_{k=1}^{b/3} u_{i, k} u_{i, k}^T}{2} \\
    &\le \max_i \sum_{k=1}^{b/3} \norm{u_{i, k} u_{i, k}^T}{2} \\
    &= \max_i \sum_{k=1}^{b/3} \norm{u_{i, k}}{2}^2 \\
    &\le \max_i \sum_{k=1}^{b/3} \frac{1}{\dsketch q} \norm{x}{2}^2 \\
    &= \frac{b}{3} \frac{1}{\dsketch q} \\
    &= \frac{1}{3 \Diamond q}
\end{align*}

Applying the Hanson-Wright inequality, Theorem~\ref{thm:hanson-wright}, we know that with high probability:
\[
  \E \abs{z^T B z - \frac13 \pm \frac{\delta}{2}}^\ell \le C^\ell \cdot \max \left\{ \sqrt{\ell} \cdot O(1/\Diamond), \ell \cdot O(1/(\Diamond q)) \right\}^\ell
\]

Applying Markov's inequality for $\ell$ an even integer:
\begin{align*}
  \Pr \left[ \abs{z^T B z - \frac13 \pm \frac{\delta}{2}} > \frac{\delta}{6} \right]
    &< \left(\frac{6}{\delta}\right)^{\ell} 
      C^\ell \cdot \max \left\{ \sqrt{\ell} \cdot O(1/\sqrt{\Diamond}), \ell \cdot O(1/(\Diamond q)) \right\}^\ell \\
  \Pr \left[ \abs{z^T B z - \frac13} > \frac{\delta}{3} \right]
    &< \left(\frac{6C}{\delta} \max \left\{ \sqrt{\ell} \cdot O(1/\sqrt{\Diamond}), \ell \cdot O(1/(\Diamond q)) \right\} \right)^{\ell} \\
\end{align*}

We'll need to choose a sufficiently large $\ell$ on the order of $\log N$:
\begin{align*}
  \Pr \left[ \abs{z^T B z - \frac13} > \frac{\delta}{3} \right]
    &< \left(\frac{6C}{\delta} \max \left\{ O\left(\frac{\sqrt{\log N}}{\sqrt{\Diamond}}\right), O\left(\frac{\log N}{\Diamond q}\right) \right\} \right)^{\Theta(\log N)} \\
    &< \left(\frac{6C}{\delta} O\left(\frac{\sqrt{\log N}}{\sqrt{\Diamond}}\right) \right)^{\Theta(\log N)}
\end{align*}

We again used that $q$ is $\Omega\left(\frac{\sqrt{\log N}}{\sqrt{\Diamond}}\right)$. We conclude by choosing $\delta = O\left(\frac{\log N}{\Diamond}\right)$, to make the base of the left-hand side a constant and achieve our polynomially small failure probability. This completes the proof of Theorem~\ref{thm:block-random-isometry}.

\subsection{Desynchronization}\label{sec:desynch}

In this subsection, we will prove Theorem~\ref{thm:block-random-desync}. Suppose we draw a random matrix $R \in \mathbb{R}^{\dsketch \times \dsketch}$ from our distribution $\distribution(b, q, \dsketch)$, and we have two unit vectors $x, y \in \rsketch$. For convenience, we will index $x$ using $[\dsketch]$ and $y$ using $\{s, c, m\} \times [\Diamond] \times [b/3]$.

As in the previous section, we split $R$ into $R_s, R_c, R_m$ (see Figure~\ref{fig:block-random-submatrix}) so that our random variables will be independent of the sub-block patterns. For one sub-matrix:
\[
  y^T R_\star x
    = \sum_{i=1}^{\Diamond} \sum_{j=1}^{\dsketch}
        \eta_{i, j} f_{\star, i, j} \inner{\sigma_{\star, i, j}}{y_{\star, i}} x_j
\]

We view this as a sum of $\Diamond \dsketch$ random variables, which we will apply Bernstein's Inequality to bound. Note that:
\begin{align*}
  \abs{\eta_{i, j} f_{\star, i, j} \inner{\sigma_{\star, i, j}}{y_{\star, i}} x_j}
    &\le \frac{b}{3} \frac{1}{\dsketch q} \\
    &= \frac{1}{3\Diamond q} \\
  \sum_{i=1}^{\Diamond} \sum_{j=1}^{\dsketch} \E \left[ \eta_{i, j}^2 f_{\star, i, j}^2 \inner{\sigma_{\star, i, j}}{y_{\star, i}}^2 x_j^2 \right]
    &= q \sum_{i=1}^{\Diamond} \sum_{j=1}^{\dsketch} \inner{\sigma_{\star, i, j}}{y_{\star, i}}^2 x_j^2 \\
    &\le q \sum_{i=1}^{\Diamond} \sum_{j=1}^{\dsketch} \norm{\sigma_{\star, i, j}}{2}^2 \norm{y_{\star, i}}{2}^2 x_j^2 \\
    &= q \sum_{i=1}^{\Diamond} \sum_{j=1}^{\dsketch} \frac{b}{3} \frac{1}{\dsketch q} \norm{y_{\star, i}}{2}^2 x_j^2 \\
    &= \frac{1}{3\Diamond}
\end{align*}

Combining these with Bernstein's Inequality yields:
\[
  \Pr\left[y^T R_\star x > \frac{\delta}{3}\right]
    \le \exp \left(
          -\frac{\delta^2/18}{\frac{1}{3\Diamond} + \frac{1}{9\Diamond q} \frac{\delta}{3}}
        \right)
\]

We note that by assumption $q = \Omega\left(\frac{\sqrt{\log N}}{\sqrt{\Diamond}}\right)$ and hence we can also choose a sufficiently large $\delta = O\left(\frac{\sqrt{\log N}}{\sqrt{\Diamond}}\right)$ to make this probability polynomially small in $N$. This completes the proof.
\section{Proof Sketches for Sketch Prototypes}
\label{apx:prototype-sketch}

In this appendix, we present proof sketches for our sketch prototypes.

\subsection{Proof Sketches for Prototype A}

Recall Claim~\ref{clm:a-attr}, which we will now sketch a proof of.

\aattr*

\begin{proofsketch}
  We multiply our overall sketch with $R_{M(\thetastar)}^T = R_{M(\thetastar)}^{-1}$. This product is $\sum_\theta w_\theta R_{M(\thetastar)}^{-1} R_{M(\theta)} x_\theta$. Our signal term occurs when $\theta = \thetastar$ and is $w_\thetastar x_\thetastar$. We think of the other terms as noise; we know from our orthonormal matrix properties that with high probability they each perturb any specific coordinate by at most $O(w_\theta \sqrt{\log N} / \sqrt{\dsketch})$. Taken together, no coordinate is perturbed by more than $O(\sqrt{\log N} / \sqrt{\dsketch})$. Dividing both our signal term and noise by $w_\thetastar$, we get the desired claim.
\end{proofsketch}

Recall Claim~\ref{clm:a-sim}, which we will now sketch a proof of.

\asim*

\begin{proofsketch}
  Consider two overall sketches $s = \sum_\theta w_\theta R_{M(\theta)} x_\theta$ and $\bar{s} = \sum_{\bar{\theta}} \bar{w}_{\bar{\theta}} R_{M(\bar{\theta})} x_{\bar{\theta}}$. Their dot-product can be written as $\sum_\theta \sum_{\bar{\theta}} w_\theta \bar{w}_{\bar{\theta}} x_\theta^T R_{M(\theta)}^T R_{M(\bar{\theta})} x_{\bar{\theta}}$.
  
  First, let us suppose we are in case (i) and they share no modules. Since orthonormal matrices are desynchronizing, each term has magnitude at most $O(w_\theta \bar{w}_{\bar{\theta}} \sqrt{\log N} / \sqrt{\dsketch})$. Since the sum of each set of weights is one, the total magnitude is at most $O(\sqrt{\log N} / \sqrt{\dsketch})$. This completes the proof of case (i).
  
  Next, let us suppose we are in case (ii) and they share an object $\thetastar$. Since orthonormal matrices are isometric, the term $\theta = \bar{\theta} = \thetastar$ has a value of $w_\thetastar \bar{w}_\thetastar$. We can analyze the remaining terms as in case (i); they have total magnitude at most $O(\sqrt{\log N} / \sqrt{\dsketch})$. The worst-case is that the noise is negative, which yields the claim for case (ii).
  
  We need to analyze case (iii) a little differently. Since our objects are paired up between the sketches, we refer to the objects as $\theta_1, \ldots, \theta_k$ (in the first sketch) and $\bar{\theta}_1, \ldots, \bar{\theta}_k$ (in the second sketch. Because they are close in $\ell_2$ distance, we can write $x_{\bar{\theta}_i} = x_{\theta_i} + \epsilon_i$, where each $\epsilon_i$ is a noise vector with $\ell_2$ length at most $\epsilon$.

  We can write our second sketch as $\bar{s} = \sum_i w_{\theta_i} R_{M(\theta_i)} \left(x_{\theta_i} + \epsilon_i\right)$ (by assumption, the paired objects had the same weight and were produced by the same module). But then $\bar{s} = s + \sum_i w_{\theta_i} R_{M(\theta_i)} \epsilon_i$! Since orthonormal matrices are isometric and the weights sum to one, $\bar{s}$ is at most $\epsilon$ away from $s$ in $\ell_2$ norm, as desired.
\end{proofsketch}

\subsection{Proof Sketches for Prototype B}

Recall Claim~\ref{clm:b-attr}, which we will now sketch a proof of.

\battr*

\begin{proofsketch}
  We begin with the easier case (i). For case (i), we attempt to recover $x_\thetastar$ via multiplying the sketch by $R_{M(\thetastar), 1}^T$. The signal component of our sketch has the form $\ptransparentx{R_i} \frac12 R_{M(\thetastar), 1}^T x_\thetastar$. The $R_i / 2$ part can be considered noise, and the $I / 2$ part gives us a quarter of the signal we had when analyzing sketch A. The noise calculation is the same as before because $R_{M(\thetastar)}$ does not occur anywhere else, so we get the same additive error (up to a constant hidden in the big-oh notation).
  
  For the harder case (ii), we multiply the sketch by $R_{M(\thetastar)}^T R_i^T$ instead. This recovers the $R_i / 2$ part instead of the $I / 2$ part, but everything else is the same as the previous case.
\end{proofsketch}

Recall Claim~\ref{clm:b-sim}, which we will now sketch a proof of.

\bsim*

\begin{proofsketch}
  Case (i) is the same as before; the desynchronizing property of orthonormal matrices means that taking the dot-product of module sketches, even when transformed by $I/2$ or $R_i/2$, results in at most $O(\sqrt{\log N} / \sqrt{\dsketch})$ noise.
  
  Regarding case (ii), our new signal term is weaker by a factor of four. Notice that for both sketches, the object sketch $\objectsketchx(\thetastar)$ is still identical. We lose our factor of four when considering that a $\ptransparentx{R_i}$ has been added into the tuple sketch; the $I / 2$ lets through the information we care about, and compounded over two sketches results in a factor four loss. If the object were located at the same index in both sketches, we would only lose a factor of two.
  
  Regarding case (iii), the new module sketch is only half-based on attributes; the other half agrees if the objects have the same module. As a result, changing the attributes has half the effect as before.
  
  Case (iv) is similar to case (i), except only the $e_1$ component of the module sketch is similar, resulting in an additional factor two loss from each sketch, for a final factor of sixteen.
\end{proofsketch}

Recall Claim~\ref{clm:b-stat}, which we will now sketch a proof of.

\bstat*

\begin{proofsketch}
  It is more intuitive to begin with case (ii). Consider what would happen if we performed case (i) of \propattr{} as in Claim~\ref{clm:b-attr}. That claim assumed that there were no other objects produced by the same module, but if there were (and they had the same weight $\wstar$), then their attribute vectors would be added together. This exactly yields the current case (ii).
  
  Next, for case (i), we simply observe that the object sketch treats $x_\theta$ the same way as it treats $e_1$, so we could use the same idea to recover the sum, over all objects with the same module, of $e_1$. That is exactly the number of such modules times $e_1$, so looking at the first coordinate gives us the desired estimate for the current case (i).
\end{proofsketch}

\nuke{
\subsection{Proof Sketches for Prototype C}

Recall Claim~\ref{clm:c-attr}, which we will now sketch a proof of.

\cstat*

\begin{proofsketch}
  We begin by proving the easier case (ii). For case (ii), we attempt to recover $x_\thetastar$ via multiplying the sketch by $R_{M(\thetastar), 1}^T$. The signal component of our sketch has the form $\matrixprod{j=1}{h} \left[ \ptransparentx{R_{i_{2j-1}}} \ptransparentx{R_{M(\theta_j), 0}}
  \ptransparentx{R_{i_{2j}}} \right] R_{M(\thetastar), 1} x_\thetastar$ where the $\theta_j$ are objects that recursively use $x_\theta$ and $i_1, ..., i_{2h}$ are the indices into our various tuple sketches. We have already argued that considering only the $I / 2$ part of each of these splits results in an additional loss factor of $2^{3h}$ versus that of sketch B.
  
  For the harder case (i), we multiply the sketch by $R_{M(\thetastar)}^T R_{i_{2h}}^T \cdots, R_{i_1}^T$, using the $R / 2$ parts and not the $I / 2$ parts, again for a total loss of $2^{3h}$ compared to sketch B.
\end{proofsketch}

Recall Claim~\ref{clm:c-sim}, which we will now sketch a proof of.

\csim*

\begin{proofsketch}
  For cases (i), (ii), and (iv), we modify our previous analysis by going through the layers of our sketch. We have already argued how this yields effective weight times $2^{3h}$.
  
  Case (iii) only relied on the fact that as the differences propagate up the sketch, they only shrink in magnitude.
\end{proofsketch}

Recall Claim~\ref{clm:c-stat}, which we will now sketch a proof of.

\cstat*
}
\section{Proofs for Final Sketch}
\label{apx:final-sketch}

In this appendix, we present proofs for the properties of our final sketch. Recall that our final sketch is defined as follows.
\finalsketchequations

If we unroll these definitions, then the overall sketch has the form:
\[
  \overallsketchx
    = \sum_\theta w_\theta
        \matrixprod{j=1}{3h(\theta)}
          \left[
            \ptransparentx{R_{\theta, j}}
          \right]
        \objectsketchx(\theta)
\]
where $h(\theta)$ is the depth of object $\theta$ and $\ptransparentx{R_{\theta, j}}$ is the $j^{th}$ matrix when walking down the sketch tree from the overall sketch to the $\theta$ object sketch (i.e. alternating between input index, module matrix, and a one or two index into module versus inputs).

\psuedosection{Technical Assumptions} For this section, we need the following technical modification to the sketch. The $R_i$ matrices for the tuple operators are \emph{redrawn at every depth level} of the sketch, where the depth starts at one at the overall sketch and increases every time we take a tuple sketch. Additionally, we assume that we can guarantee $\delta \le \frac{1}{4H}$ by appropriately increasing the sketch dimension (in the typical case where $H$ is constant, this just increases the sketch dimension by a constant); this section simply accepts a $\delta$ guarantee derived in Appendix~\ref{apx:block-random} and does not tinker with sketch parameters.

\subsection{Proofs for \propattr{}}

\begin{theorem}\label{thm:final-attr}
  Our final sketch has the \propattr{} property, which is as follows. Consider an object $\thetastar$ which lies at depth $h(\thetastar)$ in the overall sketch and with effective weight $w_\thetastar$.
  \begin{enumerate}
    \item[(i)] Suppose no other objects in the overall sketch are also produced by $M(\thetastar)$. We can produce a vector estimate of the attribute vector $x_\thetastar$, which with high probability has an additive error of at most $O(2^{3h(\thetastar)} h(\thetastar) \delta / w_\thetastar)$ in each coordinate.
    \item[(ii)] Suppose we know the sequence of input indices to get to $\thetastar$ in the sketch. Then even if other objects in the overall sketch are produced by $M(\thetastar)$, we can still produce a vector estimate of the attribute vector $x_\thetastar$, which with high probability has an additive error of at most $O(2^{3h(\thetastar)} h(\thetastar) \delta / w_\thetastar)$ in each coordinate.
  \end{enumerate}
\end{theorem}

\begin{proof}
  We begin by proving the easier case (i). Suppose that our overall sketch is $s$. Let $\beta = 2^{3h(\thetastar)+1} / w_\thetastar$, and imagine that we attempt to recover the $j^{th}$ coordinate of $x_\thetastar$ by computing $\beta e_j^T R_{M(\thetastar), 1}^T s$, which has the form:
  \[
    \sum_\theta \beta w_\theta
      e_j^T R_{M(\thetastar), 1}^T
      \matrixprod{j=1}{3h(\theta)}
        \left[
          \ptransparentx{R_{\theta, j}}
        \right]
      \objectsketchx(\theta)
  \]
  
  When $\theta = \thetastar$, we get the term:
  \[
    2^{3h(\thetastar)+1}
    e_j^T R_{M(\thetastar), 1}^T
    \matrixprod{j=1}{3h(\thetastar)}
      \left[
        \ptransparentx{R_{\thetastar, j}}
      \right]
    \objectsketchx(\thetastar)
  \]

  If we expand out each transparent matrix, we would get $2^{3h(\thetastar)}$ sub-terms. Let's focus on the sub-term where we choose $I/2$ every time:
  \[
    2 e_j^T R_{M(\thetastar), 1}^T
    \left[\frac12 R_{M(\thetastar), 1} x_\thetastar + \frac12 R_{M(\thetastar), 2} e_1\right]
  \]
  
  By Lemma~\ref{lem:advanced-isometry}, we know that with high probability (note $e_j$ and $x_\thetastar$ are both unit vectors):
  \[
    \abs{2 e_j^T R_{M(\thetastar), 1}^T \times \frac12 R_{M(\thetastar), 1} x_\thetastar
         - e_j^T x_\thetastar}
    \le 3\delta = O(\delta)
  \]
  
  This half of the sub-term is (approximately) the $j^{th}$ coordinate of $x_\thetastar$, up to an additive error of $O(\delta)$. We control the other half using Lemma~\ref{lem:diff-desync}:
  \[
    \abs{2 e_j^T R_{M(\thetastar), 1}^T \times \frac12 R_{M(\thetastar), 2} e_1}
    \le \delta \sqrt{1 + \delta} = O(\delta)
  \]

  It remains to bound the additive error from the other sub-terms and the other terms.
  
  The other sub-terms have the form:
  \[
    2 e_j^T R_{M(\thetastar), 1}^T \tilde{R} \objectsketchx(\thetastar)
  \]
  where $\tilde{R}$ is a product of one to $3h(\thetastar)$ $R$ matrices. By Lemma~\ref{lem:length-isometry}, we know that with high probability $\objectsketchx(\thetastar)$ and $R_{M(\thetastar), 1} e_j$ have squared $\ell_2$ norm at most $(1 + \delta)$.
  
  We analyze $\tilde{R}$ by repeatedly applying Lemma~\ref{lem:product-desync}: we know that the distribution of one matrix satisfies $\delta$-Desynchronization and $\delta$-Isometry. Let $\gamma = \sqrt{1 + \delta}$. Then the product distribution of two matrices satisfies $\left[ (1 + \gamma) \delta \right]$-Desynchronization, the product distribution of three matrices satisfies $\left[ (1 + \gamma + \gamma^2) \delta \right]$-Desynchronization, and the product distribution of $3h(\thetastar)$ matrices satisfies $\left[ \frac{\gamma^{3h(\thetastar)} - 1}{\gamma - 1} \delta \right]$-Desynchronization. The last desynchronization is the largest, therefore no matter how many matrices are being multiplied, with high probability the absolute value of any other sub-term is bounded by:
  \begin{align*}
    &\phantom{{}={}}
    2 \frac{\gamma^{3h(\thetastar)} - 1}{\gamma - 1} \delta (1 + \delta)
    \\ &\le
    2 \frac{\gamma^{6h(\thetastar)} - 1}{\gamma^2 - 1} \delta (1 + \delta)
    \\ &\le
    2 \frac{(1 + \delta)^{3h(\thetastar)} - 1}{\delta} \delta (1 + \delta)
    \\ &\le
    2 (1 + O(h(\thetastar) \delta) - 1) (1 + \delta)
    \\ &\le
    O(h(\thetastar) \delta)
  \end{align*}
  Note that we used the fact that $(1 + x)^n \le 1 + O(xn)$ if $xn \le \frac12$.
  
  Hence, the absolute value of the sum of the other sub-terms is bounded by $O(2^{3h(\thetastar)} h(\thetastar) \delta)$.
  
  We now bound the additive error from the other terms, which have the form:
  \[
    \beta w_\theta
    e_j^T R_{M(\thetastar), 1}^T
    \matrixprod{j=1}{3h(\theta)}
      \left[ \ptransparentx{R_{\theta, j}} \right]
    \objectsketchx(\theta)
  \]
  
  The plan is to use the desynchronization of $R_{M(\thetastar), 1}$. To do so, we will need to bound the $\ell_2$ length of $\objectsketchx(\theta)$ as it gets multiplied by this sequence of matrices. By Lemma~\ref{lem:transparent-isometry} that the distribution of these matrices is $\frac12$-scaled $(\frac34 \delta)$-Isometric. By repeatedly applying Lemma~\ref{lem:length-isometry} we know that the squared $\ell_2$ length is upper bounded by $(\frac12 + \frac34 \delta)^{3h(\theta)} (1 + \delta)$. By our assumptions, this is upper-bounded by a constant:
  \begin{align*}
    &\phantom{{}\le{}}
    (\frac12 + \frac34 \delta)^{3h(\theta)} (1 + \delta) \\
      &\le (1 + \delta)^{3h(\theta) + 1}  \\
      &\le (1 + 1 / (2H))^{2H} \\
      &\le e
  \end{align*}
  
  Now, by the desynchronization of $R_{M(\thetastar), 1}$, we know that with high probability the absolute value of the term indexed by $\theta$ is at most $O(\beta w_\theta \delta)$. Hence with high probability the absolute value of all other terms is at most $O(\beta \delta)$. Plugging in our choice of $\beta$, it is at most $O(2^{3h(\thetastar) + 1} \delta / w_\thetastar)$.
  
  Hence we have shown that with high probability, we can recover the $j^{th}$ coordinate of $x_\thetastar$ with the desired additive error. This completes the proof of case (i).

  We now prove the harder case (ii). Again, we let $\beta = 2^{3h(\thetastar) + 1} / w_\thetastar$, and imagine that we attempt to recover the $j^{th}$ coordinate of $x_\thetastar$ by computing $\beta e_j^T R_{M(\thetastar), 1}^T \revmatrixprod{j=1}{3h(\thetastar)} \left[ R_{\thetastar, j}^T \right] s$, which has the form:
  \begin{align*}
    &\sum_\theta \beta w_\theta
      e_j^T R_{M(\thetastar), 1}^T
      \revmatrixprod{j=1}{3h(\thetastar)}
        \left[
          R_{\thetastar, j}^T
        \right]
    \\ &\qquad
      \matrixprod{j=1}{3h(\theta)}
        \left[
          \ptransparentx{R_{\theta, j}}
        \right]
      \objectsketchx(\theta)
  \end{align*}
  It is worth noting that we \emph{did not compute} the alternative product $\beta e_j^T R_{M(\thetastar), 1}^T \revmatrixprod{j=1}{3h(\thetastar)} \left[ \ptransparentx{R_{\thetastar, j}^T} \right] s$ because the transparent matrices in the alternative would allow through information about other objects produced by the same module.
  
  When $\theta = \thetastar$, we get the term:
  \begin{align*}
    & 2^{3h(\thetastar) + 1}
      e_j^T R_{M(\thetastar), 1}^T
      \revmatrixprod{j=1}{3h(\thetastar)}
        \left[
          R_{\thetastar, j}^T
        \right]
    \\ &\qquad
      \matrixprod{j=1}{3h(\thetastar)}
        \left[
          \ptransparentx{R_{\thetastar, j}}
        \right]
      \objectsketchx(\thetastar)
  \end{align*}
  
  If we expand out each transparent matrix, we would get $2^{3h(\thetastar)}$ sub-terms. Let's focus on the sub-term where we choose $R/2$ every time:
  \begin{align*}
    &2 e_j^T R_{M(\thetastar), 1}^T \tilde{R}^T \tilde{R}
    \left[
      \frac12 R_{M(\thetastar), 1} x_\thetastar +
      \frac12 R_{M(\thetastar), 2} e_1
    \right]
  \end{align*}
  where $\tilde{R} = \matrixprod{j=1}{3h(\thetastar)} [R_{\thetastar, j}]$.
  
  As before, we expect to get some signal from the first half $\frac12 R_{M(\thetastar), 1} x_\thetastar$ and some noise from the second half $\frac12 \R_{M(\thetastar), 2} e_1$. Let us begin with the first half. The matrices involved match exactly, so it is a matter of repeatedly applying Lemma~\ref{lem:product-isometry}. Note that all matrices involved are $\delta$-Isometric. Since at each step we multiply the isometry parameter by $(1 + \delta)$ and then add $\delta$, the isometry parameter after combining $3h(\thetastar)$ times is:
  \begin{align*}
    \sum_{i=0}^{3h(\thetastar)} \delta (1 + \delta)^i
      &= \delta
           \frac{(1 + \delta)^{3h(\thetastar) + 1} - 1}{1 + \delta - 1} \\
      &= (1 + \delta)^{3h(\thetastar) + 1} - 1 \\
      &\le 1 + O(h(\thetastar) \delta) - 1 \\
      &\le O(h(\thetastar) \delta)
  \end{align*}
  Note that we used the fact that $(1 + x)^n \le 1 + O(xn)$ if $xn \le \frac12$ along with our assumption that $\delta < 1 / (4H)$.
  
  Hence the product of all $(3h(\thetastar)+1)$ matrices is $O(h(\thetastar) \delta)$-Isometric. By Lemma~\ref{lem:advanced-isometry} this implies that with high probability:
  \begin{align*}
    \abs{
      \inner{\tilde{R} R_{M(\thetastar), 1} e_j}
            {\tilde{R} R_{M(\thetastar), 1} x_\thetastar}
      - \inner{e_j}{x_\thetastar}
    }
    \le O(h(\thetastar) \delta)
  \end{align*}

  We now control the noise half. By Lemma~\ref{lem:diff-desync}, with high probability:
  \[
    \abs{\inner{R_{M(\thetastar), 1} e_j}
               {R_{M(\thetastar), 2} e_1}
    }
    \le \delta \sqrt{1 + \delta} = O(\delta)
  \]
  We have already computed that the remaining $3h(\thetastar)$ matrices is $O(H\delta)$-Isometric and hence approximately preserves this. Lemma~\ref{lem:advanced-isometry} says that with high probability:
  \begin{align*}
    &\bigg| \inner{\tilde{R} R_{M(\thetastar), 1} e_j}
               {\tilde{R} R_{M(\thetastar), 2} e_1}
  \\ &\qquad
       - \inner{R_{M(\thetastar), 1} e_j}
               {R_{M(\thetastar), 2} e_1}
    \bigg| \le O(H \delta)
  \end{align*}
  By the triangle inequality, we now know that with high probability, the sub-term where we choose $R/2$ every time is at most $O(H \delta)$ additive error away from the $j^{th}$ coordinate of $x_\thetastar$.
  
  We now consider the other sub-terms. These other sub-terms are of the form:
  \[
    2 e_j^T R_{M(\thetastar), 1}^T \tilde{R}^T \tilde{R}'
    \left[
      \frac12 R_{M(\thetastar), 1} x_\thetastar +
      \frac12 R_{M(\thetastar), 2} e_1
    \right]
  \]
  where $\tilde{R}$ is still $\matrixprod{j=1}{3h(\thetastar)} [R_{\thetastar, j}]$ but $\tilde{R}'$ is now the product of a subsequence of those terms; we write it as $\matrixprod{j=1}{3h(\thetastar)} [S_{\thetastar, j}]$ where $S_{\thetastar, j}$ is either $R_{\thetastar, j}$ or the identity matrix.
  
  Next, we define $x_J \coloneqq \matrixprod{j=J}{3h(\thetastar)} [R_{\thetastar, j}] R_{M(\thetastar), 1} e_j$ and $x_J' \coloneqq \matrixprod{j=J}{3h(\thetastar)} [S_{\thetastar, j}] \objectsketchx(\theta)$. Additionally, when $J = 3h(\thetastar) + 1$, $x_J = R_{M(\thetastar), 1} e_j$ and $x_J' = \objectsketchx(\theta)$. Our plan is to induct backwards from $J = 3h(\thetastar) + 1$, controlling the dot product of the two.
  
  By Lemma~\ref{lem:length-isometry}, we know that all $x_J$ and $x_J'$ have squared $\ell_2$ length at most $(1 + \delta)^{3h(\thetastar)+ 1}$. Under our assumptions, this is less than $e$ and is hence a constant.
  
  Lemma~\ref{lem:diff-desync} and Lemma~\ref{lem:advanced-isometry} already imply that for our base case of $J = 3h(\thetastar) + 1$, the dot product of the two is at most $O(\delta)$ away from the $j^{th}$ coordinate of $x_\thetastar$.
  
  How does the dot product change from $J$ to $J - 1$? We either multiply $x_J$ by a random matrix, or both $x_J$ and $x_J'$ by the same random matrix. In the first case, the $\delta$-Desynchronization of our matrices tells us that the new dot-product will be zero times the old dot-product plus $O(\delta)$ error. In the second case, the $\delta$-Isometry of our matrices tells us that the new dot-product will be the old dot-product plus $O(\delta)$ error. Since the first case must occur \emph{at least once}, we know that the final dot product when $J = 1$ consists of no signal and $O(h(\thetastar) \delta)$ noise.
  
  Hence, summing up our $2^{3h(\thetastar)} - 1$ other sub-terms, we arrive at $O(2^{3h(\thetastar)} h(\thetastar) \delta)$ noise. It remains to consider the other terms.
  
  Unfortunately, we will need to analyze our other terms by breaking them into sub-terms as well, since otherwise we would need to consider the situation where $R$ is applied to one vector and $\ptransparentx{R}$ to the other. The other terms have the form:
  \begin{align*}
    & \beta w_\theta
      e_j^T R_{M(\thetastar), 1}^T
      \revmatrixprod{j=1}{3h(\thetastar)}
        \left[
          R_{\thetastar, j}^T
        \right]
    \\ &\qquad
      \matrixprod{j=1}{3h(\theta)}
        \left[
          \ptransparentx{R_{\theta, j}}
        \right]
      \objectsketchx(\theta)
  \end{align*}
  
  We break our terms into sub-terms, but \emph{only for the first $3h(\thetastar)$ choices}. Our analysis focuses on handling the case where $\theta$ is at a greater depth than $\thetastar$; we handle the other case by inserting identity matrices and splitting them into $I/2 + I/2$. We wind up with $2^{3h(\thetastar)}$ sub-terms. Such a sub-term has the form:
  \[
    (2 w_\theta / w_\thetastar) e_j^T R_{M(\thetastar), 1}^T \tilde{R}^T \tilde{R}' \tilde{R}''
    \objectsketchx(\theta)
  \]
  where
  \begin{align*}
    \tilde{R} &\coloneqq \matrixprod{j=1}{3h(\thetastar)} [R_{\thetastar, j}] \\
    \tilde{R}' &\coloneqq \matrixprod{j=1}{3h(\thetastar)} [S_{\theta, j}] \\
    \tilde{R}'' &\coloneqq
      \matrixprod{j=3h(\thetastar)+1}{3h(\theta)} \left[\transparentx{R_{\theta, j}}\right]
  \end{align*}
  and $S_{\theta, j}$ is either $R_{\theta, j}$ or the identity matrix.
  
  We define:
  \begin{align*}
    x_J &\coloneqq \matrixprod{j=J}{3h(\thetastar)} [R_{\thetastar, j}]
                   R_{M(\thetastar), 1} e_j \\
    x_J' &\coloneqq \matrixprod{j=J}{3h(\thetastar)} [S_{\theta, j}]
                   \tilde{R}'' \objectsketchx(\theta)
  \end{align*}
  When $J = 3h(\thetastar) + 1$, we define $x_J \coloneqq R_{M(\thetastar), 1} e_j$ and $x_J' = \tilde{R}'' \objectsketchx(\theta)$.
  
  As usual, Lemma~\ref{lem:length-isometry} guarantees that all $x_J$ and $x_J'$ have constant $\ell_2$ length. For our base induction case $J = \left[\max (3h(\thetastar), 3h(\theta)) + 1\right]$, we would like to know the dot product of $R_{M(\thetastar), 1} e_j$ with $\tilde{R}'' \objectsketchx(\theta)$.
  
  We first analyze the dot product of $R_{M(\thetastar), 1}$ with $\objectsketchx(\theta)$. Keep in mind that it may be the case that $M(\theta) = M(\thetastar)$. In the same module case, we know by Lemma~\ref{lem:advanced-isometry} and Lemma~\ref{lem:diff-desync} that we get a signal of the $j^{th}$ component of $x_\theta$ along with $O(\delta)$ noise. In the different module case, Lemma~\ref{lem:diff-desync} is enough to tell us that there is no signal and $O(\delta)$ noise. As a reminder, the goal is to show that this signal \emph{does not get through}, since we only want the $j^{th}$ component of $x_\thetastar$, not any other $x_\theta$.
  
  We know by Lemma~\ref{lem:transparent-desync} that our transparent matrices are $\frac12$-leaky $O(\delta)$-Desynchronizing. Hence after applying all the transparent matrices, we know that at most we have our original signal (the $j^{th}$ component of $x_\theta$ or nothing) and $O(\delta)$ noise (the noise telescopes due to the $\frac12$ leaking/scaling). This is the base case of our backwards induction, $J = 3h(\thetastar) + 1$.
  
  We now consider what happens to the dot product of $x_J$ and $x_J'$ as we move from $J$ to $J - 1$. We may either (a) multiply both by different random matrices, (b) multiply both by the same random matrix, or (c) multiply only one by a random matrix. Furthermore, we know that the sequences $R_{\thetastar, \cdot}$ and $R_{\theta, \cdot}$ differ at some point in the first $3h(\thetastar)$ matrices, so we must end up in case (a) or case (c) at least once.
  
  For case (a), we know by Lemma~\ref{lem:diff-desync} that the current dot-product will be multiplied by zero and $O(\delta)$ noise will be added. For case (b), we know by Lemma~\ref{lem:advanced-isometry} that the current dot-product will be multiplied by one and $O(\delta)$ noise will be added. For case (c), we know by $d$-Desynchronization that the current dot-product will be multiplied by zero and $O(\delta)$ noise will be added.
  
  Since case (a) or (c) must occur at least once, we know that we get no signal and at most $O(h(\thetastar) \delta)$ noise. Since each of our $2^{3h(\thetastar)}$ sub-terms has a multiplier of $(2 w_\theta / w_\thetastar)$ on top of the dot product of $x_1$ and $x_1'$, each term has noise bounded in magnitude (still with high probability) by $O(w_\theta 2^{3h(\thetastar)} h(\thetastar) \delta / w_\thetastar)$. Summing over all objects $\theta$, we get our final desired claim: with high probability, the total additive noise is bounded by $O(2^{3h(\thetastar)} h(\thetastar) \delta / w_\thetastar)$. This completes the proof.
\end{proof}

\psuedosection{Proof Notes} Although our proofs were written as if we extracted one attribute value at a time, we can extract them simultaneously by computing the vector $\beta R_{M(\thetastar), 1}^T s$, which is exactly a stacked version of the terms we were considering. Our technique is extendable to cases between (i) and (ii). For example, suppose we know that only one object in the overall sketch was both produced by the eye module and then subsequently used by the face module. We could recover its attributes by computing the vector $\beta R_{\text{eye}, 1}^T R_{\text{face}, 0}^T s$. Finally, the way we extract the attributes from the overall sketch is the same way we could extract attributes from a module sketch, meaning we do not need to know the precise sketch we are working with when attempting to extract information (although it will of course affect the error).

\subsection{Proofs for \propsim{}}

\begin{theorem}\label{thm:final-sim}
  Our final sketch has the \propsim{} property, which is as follows. Suppose we have two overall sketches $s$ and $\bar{s}$.
  \begin{enumerate}
    \item[(i)] If the two sketches share no modules, then with high probability they have at most $O(\delta)$ dot-product.
    \item[(ii)] If $s$ has an object $\thetastar$ of weight $w_\thetastar$ and $\bar{s}$ has an object $\bar{\theta}^\star$ of weight $\bar{w}_\thetastar$ and both objects are produced by the same module, then with high probability they have at least $\Omega(2^{-6h(\thetastar)} w_\thetastar \bar{w}_{\bar{\theta}^\star}) - O(\delta)$ dot-product.
    \item[(iii)] If the two sketches are identical except that the attributes of any object $\theta$ differ by at most $\epsilon$ in $\ell_2$ distance, then with probability one the two sketches are at most $\epsilon / 2$ from each other in $\ell_2$ distance.
  \end{enumerate}
\end{theorem}

\begin{proof}
  We prove cases (i) and (ii) together. Suppose that our overall sketches are $s, \bar{s}$. Then the dot-product of our sketches has the form:
  \begin{align*}
    &\sum_\theta \sum_\thetabar
        w_\theta \bar{w}_\thetabar
    \\ &\phantom{{}={}} \qquad
      \objectsketchx(\theta)^T
      \revmatrixprod{j=1}{3h(\theta)}
        \left[ \ptransparentx{R_{\theta, j}^T} \right]
    \\ &\phantom{{}={}} \qquad
      \matrixprod{j=1}{3h(\thetabar)}
        \left[ \ptransparentx{R_{\thetabar, j}} \right]
      \objectsketchx(\thetabar)
  \end{align*}
  
  Our proof plan is to analyze a single term of this double-summation at a time. Fix $\theta$ and $\thetabar$. We analyze this term with our toolkit of technical lemmas. The plan is to first understand $\inner{\objectsketchx(\theta)} {\objectsketchx(\thetabar)}$. There are two possibilities for this dot-product, depending on our two objects $\theta$ and $\thetabar$. If these objects come from different modules, then we will want a small dot-product. If these objects come from the same module, then we will want a significant dot-product.
  
  \psuedosection{Dot-Product Under Different Modules} Since the matrices are different, we know by Lemma~\ref{lem:diff-desync} that with high probability, all of the following are true:
  \begin{align*}
    \abs{\inner{
      R_{M(\theta), 1} x_\theta
    }{
      R_{M(\thetabar), 1} x_\thetabar
    }}
    \le \delta \sqrt{1 + \delta}
    &\le O(\delta) \\
    \abs{\inner{
      R_{M(\theta), 1} x_\theta
    }{
      R_{M(\thetabar), 2} e_1
    }}
    \le \delta \sqrt{1 + \delta}
    &\le O(\delta) \\
    \abs{\inner{
      R_{M(\theta), 2} e_1
    }{
      R_{M(\thetabar), 1} x_\thetabar
    }}
    \le \delta \sqrt{1 + \delta}
    &\le O(\delta) \\
    \abs{\inner{
      R_{M(\theta), 2} e_1
    }{
      R_{M(\thetabar), 2} e_1
    }}
    \le \delta \sqrt{1 + \delta}
    &\le O(\delta)
  \end{align*}
  Combining these statements and using the triangle inequality, we know that with high probability,
  \[
    \abs{\inner{\objectsketchx(\theta)}{\objectsketchx(\thetabar)}} \le O(\delta)
  \]
  
  \psuedosection{Dot-Product Under Same Module} Since the matrices are the same, we know by Lemma~\ref{lem:advanced-isometry} that with high probability, both of the following are true:
  \begin{align*}
    \abs{\inner{
      R_{M(\theta), 1} x_\theta
    }{
      R_{M(\thetabar), 1} x_\thetabar
    } - \inner{x_\theta}{x_\thetabar}}
    \le 3\delta
    &\le O(\delta) \\
    \abs{\inner{
      R_{M(\theta), 2} e_1
    }{
      R_{M(\thetabar), 2} e_1
    } - \inner{e_1}{e_1}}
    \le 3\delta
    &\le O(\delta)
  \end{align*}
  Also, we still know by Lemma~\ref{lem:diff-desync} that with high probability, both of the following are true:
  \begin{align*}
    \abs{\inner{
      R_{M(\theta), 1} x_\theta
    }{
      R_{M(\thetabar), 2} e_1
    }}
    \le \delta \sqrt{1 + \delta}
    &\le O(\delta) \\
    \abs{\inner{
      R_{M(\theta), 2} e_1
    }{
      R_{M(\thetabar), 1} x_\thetabar
    }}
    \le \delta \sqrt{1 + \delta}
    &\le O(\delta)
  \end{align*}
  Combining these statements with the triangle inequality, we know that with high probability,
  \[
    \abs{\inner{\objectsketchx(\theta)}{\objectsketchx(\thetabar)} -
         \frac14 \inner{x_\theta}{x_\thetabar} - \frac14 \inner{e_1}{e_1}} \le O(\delta)
  \]
  Since attribute vectors are nonnegative, we know that:
  \[
    \inner{\objectsketchx(\theta)}{\objectsketchx(\thetabar)} \ge \frac14 - O(\delta)
  \]
  
  We have successfully established our desired facts about the base dot-product and are now ready to consider what happens when we begin applying matrices of the form $\ptransparentx{R}$ to both vectors in this dot product. We begin by controlling the $\ell_2$ norm of the various vectors produced by gradually applying these matrices. By Lemma~\ref{lem:length-isometry}, we know that every vector of the form
  \[
    x_J = \matrixprod{j=J}{3h(\theta)}
      \left[ \ptransparentx{R_{\theta, j}} \right]
    \objectsketchx(\theta)
  \]
  where $J \in \{1, 2, \ldots, 3h(\theta)\}$ or of the form
  \[
    \bar{x}_J = \matrixprod{j=J}{3h(\bar{\theta})}
      \left[ \ptransparentx{R_{\bar{\theta}, j}} \right]
    \objectsketchx(\bar{\theta})
  \]
  where $J \in \{1, 2, \ldots, 3h(\thetabar)\}$ has squared $\ell_2$ norm at most $(\frac12 + \frac34 \delta)^{\max (3h(\theta), 3h(\thetabar))} (1 + \delta)$. By our assumptions, this is upper-bounded by a constant:
  \begin{align*}
    &\phantom{{}\le{}}
    (\frac12 + \frac34 \delta)^{\max (3h(\theta), 3h(\thetabar))} (1 + \delta) \\
      &\le (1 + \delta)^{\max (3h(\theta), 3h(\thetabar)) + 1}  \\
      &\le (1 + 1 / (2H))^{2H} \\
      &\le e
  \end{align*}
  We have shown that our unit vectors maintain constant $\ell_2$ norm throughout this process. This allows us to safely apply our desynchronization and isometry properties to any $x_J$ or $\bar{x}_J$.
  
  From here, we plan to induct in order of decreasing depth. For homogeneity of notation, we will define $x_J$ to be $\objectsketchx(\theta)$ when $J > 3h(\theta)$ and $\bar{x}_J$ to be $\objectsketchx(\thetabar)$ when $J > 3h(\thetabar)$. We induct backwards, beginning with $J = \max (3h(\theta), 3h(\bar{\theta})) + 1$ and ending with $J = 1$.
  
  In our base case, we know that these are two possibilities for $\inner{x_J}{\bar{x}_J}$. If we are in the different module case, then the magnitude of this quantity is $O(\delta)$. If we are in the same module case, then this quantity is at least $\frac14 - O(\delta)$.
  
  To get from $J$ to $J - 1$, there are three possibilities. We either (a) multiply both $x_J$ and $\bar{x}_J$ by the same matrix, (b) multiply $x_J$ and $\bar{x}_J$ by different matrices, or (c) multiply only one of them by a matrix. Note that case (c) happens when one object is at a different depth than the other object, so our analysis first peels off the additional matrices to make them the same depth, then proceeds. This is an important point, because tuple matrices are only shared across the same depth.
  
  In any one of these lettered cases, we know from Lemma~\ref{lem:transparent-desync} and Lemma~\ref{lem:transparent-isometry} that the distribution of the matrix(ces) is $\frac12$-leaky $O(\delta)$-Desynchronizing and $\frac12$-scaled $O(\delta)$-Isometric.
  
  In case (a), we apply Lemma~\ref{lem:advanced-isometry} to show that with high probability, the new dot product is $\frac12$ times the old dot product plus $O(\delta)$ additive noise. In case (b), we apply Lemma~\ref{lem:diff-desync} to show that with high probability, the new dot product is $\frac14$ times the old dot product plus $O(\delta)$ additive noise. In case (c), we use the fact that our distribution is $\frac12$-leaky $O(\delta)$-Desynchronizing to show that the new dot product is $\frac12$ times the old dot product, plus $O(\delta)$ additive noise.
  
  Hence, if the objects were produced by different modules, then we begin with $O(\delta)$ noise and at every step we reduce by at least a factor of $\frac12$ before adding $O(\delta)$ noise. The total noise telescopes into $O(\delta)$ total. If the objects were produced by the same module, then we begin with at least $\frac14$ signal and $O(\delta)$ noise. The noise sums as before into $O(\delta)$ total, and the signal falls by at most $\frac14$ per step, resulting in a final signal of $\Omega(2^{-6h(\theta)})$.
  
  However, our original term under consideration was actually a scaled version of what we just analyzed: $w_\theta \bar{w}_\thetabar x_1^T \bar{x}_1$. Hence in the different module case, there is $O(w_\theta \bar{w}_\thetabar \delta)$ noise, and in the same module case there is $\Omega(2^{-6h(\theta)} w_\theta \bar{w}_\thetabar)$ signal and $O(w_\theta \bar{w}_\thetabar \delta)$ noise.
  
  Hence for case (i) which we were originally trying to prove, since the effective weights $w_\theta$ sum to one and the effective weights $\bar{w}_\thetabar$ sum to one, we wind up, with high probability, at most $O(\delta)$ total noise, as desired. For case (ii) which we were originally trying to prove, we get at least $\Omega(2^{-6h(\thetastar)} w_\thetastar \bar{w}_{\bar{\thetastar}}) - O(\delta)$ dot product.
  
  We conclude by proving case (iii). Suppose that in sketch $\bar{s}$, the attributes of $\theta$ are $x_\theta + \epsilon_\theta$, where $\norm{\epsilon_\theta}{2} \le \epsilon$ for all objects $\theta$. Let's write the difference between the two sketches.
  \begin{align*}
    \bar{s} - s
      &= \sum_\theta
           w_\theta
           \matrixprod{j=1}{3h(\theta)}
             \left[ \ptransparentx{R_{\theta, j}} \right]
           \frac12 R_{M(\theta), 1} \epsilon_\theta
  \end{align*}
  
  Applying Lemma~\ref{lem:transparent-isometry}, we know that $\ptransparentx{R}$ is $\frac12$-scaled $\frac34 \delta$-Isometric. Repeatedly applying Lemma~\ref{lem:advanced-isometry}, we can conclude that with high probability:
  \begin{align*}
    &\phantom{{}\le{}}
    \norm{
      \matrixprod{j=1}{3h(\theta)}
        \left[ \ptransparentx{R_{\theta, j}} \right] R_{M(\theta), 1}
      \frac14 \epsilon_\theta
    }{2}^2 \\
      &\le (\frac12 + \frac34 \delta)^{3h(\theta)} (1 + \delta) \frac14 \epsilon^2 \\
      &\le (\frac12 + \frac38)^{3h(\theta) + 1} \frac14 \epsilon^2 \\
      &\le \frac14 \epsilon^2
  \end{align*}
  Taking the square root of the above and applying the triangle inequality yields the following.
  \begin{align*}
    \norm{\bar{s} - s}{2} &\le \sum_\theta w_\theta \frac12 \epsilon \\
      &\le \frac12 \epsilon
  \end{align*}
  This completes the proof of case (iii).
\end{proof}

\subsection{Proofs for \propstat{}}

\begin{theorem}\label{thm:final-stat}
  Our final sketch has the \propstat{} property, which is as follows. Consider a module $\Mstar$ which lies at depth $h(\Mstar)$, and suppose all the objects produced by $\Mstar$ has the same effective weight $\wstar$.
  \begin{enumerate}
    \item[(i)] \property{Frequency Recovery}. We can produce an estimate of the number of objects produced by $\Mstar$. With high probability, our estimate has an additive error of at most $O(2^{3h(\Mstar)} \delta / \wstar)$.
    \item[(ii)] \property{Summed Attribute Recovery}. We can produce a vector estimate of the summed attribute vectors of the objects produced by $\Mstar$. With high probability, our estimate has additive error of at most $O(2^{3h(\Mstar)} \delta / \wstar)$ in each coordinate.
  \end{enumerate}
\end{theorem}

\begin{proof}
  Let the overall sketch be $s$.

  We begin by proving case (i). As we did in case (ii) of \propattr{}, let $\beta = 2^{3h(\Mstar) + 1} / \wstar$. We compute $\beta e_1^T R_{\Mstar, 1}^T s$, which has the form:
  \[
    \sum_\theta \beta w_\theta
      e_1^T R_{\Mstar, 2}^T
      \matrixprod{j=1}{3h(\theta)}
        \left[
          \ptransparentx{R_{\theta, j}}
        \right]
      \objectsketchx(\theta)
  \]
  
  There are two types of terms in this sum: those objects produced by $\Mstar$, and those that are not. Consider some object $\thetastar$ produced by $\Mstar$. When $\theta = \thetastar$, we get the term:
  \[
    2^{3h(\thetastar)+1}
    e_1^T R_{\Mstar, 2}^T
    \matrixprod{j=1}{3h(\Mstar)}
      \left[
        \ptransparentx{R_{\thetastar, j}}
      \right]
    \objectsketchx(\thetastar)
  \]
  
  If we expand out each transparent matrix, we would get $2^{3h(\Mstar)}$ sub-terms. Again, focus on the sub-term where we choose $I / 2$ every time:
  \[
    2 e_1^T R_{\Mstar, 2}^T
    \left[\frac12 R_{\Mstar, 1} x_\thetastar + \frac12 R_{\Mstar, 2} e_1\right]
  \]
  
  By the $\delta$-Isometry property of our distribution, we know that with high probability (note $e_1$ is a unit vector):
  \[
    \abs{2 e_1^T R_{\Mstar, 2}^T \times \frac12 R_{\Mstar, 2} e_1
         - e_1^T e_1}
    \le \delta
  \]
  This half of the sub-term (up to additive error $O(\delta)$) contributes one to our estimate, representing one copy of an object produced by $\Mstar$. We control the other half of the sub-term using Lemma~\ref{lem:diff-desync}:
  \[
    \abs{2 e_1^T R_{\Mstar, 2}^T \times \frac12 R_{\Mstar, 1} x_\thetastar}
    \le \delta \sqrt{1 + \delta} = O(\delta)
  \]
  
  Next, we bound the additive error from the other sub-terms of such a term $\thetastar$. These other sub-terms have the form:
  \[
    2 e_1^T R_{\Mstar, 2}^T \tilde{R} \objectsketchx(\thetastar)
  \]
  where $\tilde{R}$ is a product of one to $3h(\Mstar)$ $R$ matrices. As before, we apply Lemma~\ref{lem:length-isometry} so that with high probability $\objectsketchx(\thetastar)$ and $R_{\Mstar, 2} e_1$ have squared $\ell_2$ norm at most $(1 + \delta)$, and then conclude using Lemma~\ref{lem:product-desync} that the distribution of $\tilde{R}$ satisfies $O(\delta)$-Desynchronization (this was the worst case). Hence with high probability the absolute value of any other sub-term was bounded by $O(\delta)$, and the absolute value of the sum of the other sub-terms for $\thetastar$ is bounded by $O(2^{3h(\Mstar)} \delta)$.
  
  There are at most $1 / \wstar$ such objects, so the total error from these terms is $O(2^{3h(\Mstar)} \delta / \wstar)$, as desired. It remains to show that the total error from the other terms is also $O(2^{3h(\Mstar)} \delta / \wstar)$. These other terms have the form:
  \[
    \beta w_\theta
    e_1^T R_{\Mstar, 2}^T
    \matrixprod{j=1}{3h(\theta)}
      \left[ \ptransparentx{R_{\theta, j}} \right]
    \objectsketchx(\theta)
  \]
  We prove this using the $\delta$-Desynchronization of $R_{\Mstar, 2}$. To do so, we will need to bound the $\ell_2$ length of $\objectsketchx(\theta)$ as it gets multiplied by this sequence of matrices. As before, we use Lemma~\ref{lem:transparent-isometry} to know that the distribution of these matrices is $\frac12$-scaled $(\frac34 \delta)$-Isometric, and then repeatedly apply Lemma~\ref{lem:product-isometry} to conclude that the squared $\ell_2$ length is upper bounded by a constant. Hence the $\delta$-Desynchronization of $R_{\Mstar, 2}$ implies that with high probability the absolute value of the term indexed by $\theta$ is at most $O(\beta w_\theta \delta)$. Hence  the absolute value of all other terms is at most $O(\beta \delta)$. Plugging in our choice of $\beta$, it is at most $O(2^{3h(\Mstar)} \delta / \wstar)$, as desired.
  
  We have finished proving case (i); with high probability, we can recover the number of objects produced by module $\Mstar$, with the desired additive error.
  
  We now prove case (ii). We do exactly what we did in case (i), except that instead of computing $\beta e_1^T R_{\Mstar, 2}^T s$ we compute $\beta e_j^T R_{\Mstar, 1}^T$. Instead of operating on the first coordinate of the $e_1$ in our object sketch, we now operate on the $j^{th}$ coordinate of the $x_\theta$ in our object sketch, and hence we can recover the $j^{th}$ coordinate of the sum of the attribute vectors of the appropriate objects to the same error as in case (i).
\end{proof}

\psuedosection{Proof Notes} As when we were reasoning about the \propattr{} property, we can simultaneously extract the entire summed attribute vector at the same time by computing the vector $\beta R_{\Mstar, 1}^T s$. Additionally, we focus on more precise groups of objects than all those output by a particular module. For example, if we cared about objects produced by the eye module and then subsequently used by the face module, we could recover their approximate frequency and summed attribute vector by computing $\beta e_1^T R_{\text{eye}, 2}^T R_{\text{face}, 0}^T s$ and $\beta R_{\text{eye}, 1}^T R_{\text{face}, 0}^T s$.

\subsection{Generalizing for \propsig{}}\label{subsec:sig-sketch}

In this subsection, we explain how to also satisfy the \propsig{} property through a small modification to our sketch. The key idea is that each object is given a ``magic number'' based on its attributes: $m_\theta$ is a random $(\log N)$-sparse unit vector whose entries are in $\{0, \frac{1}{\sqrt{\log N}}\}$. We revise the sketch of object $\theta$ to be $\frac13 R_{M(\theta), 1} x_\theta + \frac13 R_{M(\theta), 2} e_1 + \frac13 R_{M(\theta), 3} m_\theta$. The process of recovering an object signature is the same as recovering its attribute vector, except that the signature is quantized and therefore as long as the additive error is less than $\frac{1}{2\sqrt{\log N}}$, we can recover it exactly.

\subsection{Generalizing for \properase{}}\label{subsec:erase-sketch}

In this subsection, we explain how to generalize our previous proofs for the \properase{} property. We will need to better understand the behavior of our matrices when only a prefix of the result is considered. We can extend our definitions as follows:
\begin{definition}
  We say that a distribution $D$ over random $\dsketch \times \dsketch$ matrices satisfies the $d_{DE}$-prefix $\alpha_{DE}$-leaky $\delta_{DE}$-Desynchronization Property if the following is true. If $R \sim D$, then the random variable $Z = \sum_{i=1}^{d_{DE}} (Rx)_i (y)_i - \alpha_{DE} \inner{x}{y}$ is a $\delta_{DE}$-nice noise variable. If $\alpha_{DE} = 0$ then we simply refer to this as the $d_{DE}$-prefix $\delta_{DE}$-Desynchronization Property.
\end{definition}

\begin{definition}
  We say that a distribution $D$ over random $\dsketch \times \dsketch$ matrices satisfies the $d_{IE}$-prefix $\alpha_{IE}$-leaky $\delta_{IE}$-Isometry Property if the following is true. If $R \sim D$ and $x = y$, then the random variable $Z = \sum_{i=1}^{d_{IE}} (Rx)_i^2 - \alpha_{IE} \inner{x}{x}$ is a $\delta_{IE}$-nice noise variable. If $\alpha_{IE} = 1$ then we simply refer to this as the $d_{IE}$-prefix $\delta_{IE}$-Isometry Property.
\end{definition}

The next step is to extend our technical lemmas to reason about these prefixes. For example, we will again want to instead compare $\sum_{i=1}^{d_{IE}} (Rx)_i (Ry)_i$ and $\inner{x}{y}$ instead, so Lemma~\ref{lem:advanced-isometry} turns into the following:
\begin{lemma}\label{lem:erase-advanced-isometry}
  If $D$ satisfies the $d_{IE}$-prefix $\alpha_{IE}$-leaky $\delta_{IE}$-Isometry Property, then the random variable $Z = \sum_{i=1}^{d_{IE}} (Rx)_i (Ry)_i - \alpha_{IE} \inner{x}{y}$ is a $(3 \delta_{IE})$-nice noise variable.
\end{lemma}

\begin{proof}
  The proof is the same idea as before. The claim is trivially true if $x$ or $y$ is zero, and then without loss of generality we scale them to unit vectors.
  
  We invoke the $d_{IE}$-prefix $\alpha_{IE}$-leaky $\delta_{IE}$-Isometry Property three times, on $x_1 = x + y$, $x_2 = x$, and $x_3 = y$. This implies that the following three noise variables are $d_{IE}$-nice:
  \begin{align*}
    Z_1 &= \sum_{i=1}^{d_{IE}} (R(x + y))_i^2 - \alpha_{IE} \inner{x + y}{x + y} \\
    Z_2 &= \sum_{i=1}^{d_{IE}} (R(x))_i^2 - \alpha_I \inner{x}{x} \\
    Z_3 &= \sum_{i=1}^{d_{IE}} (R(y))_i^2 - \alpha_I \inner{y}{y}
  \end{align*}
  
  By construction, $Z = \frac12 Z_1 - \frac12 Z_2 - \frac12 Z_3$, so $Z$ is centered. We bound it by noting that $\norm{x + y}{2}^2 \le 4$. We know that with high probability,
  $\abs{Z_1} \le 4\delta_{IE}$, $\abs{Z_2} \le \delta_{IE}$, and $\abs{Z_3} \le \delta_{IE}$. By the triangle inequality, $Z$ is $(3\delta_{IE})$-bounded, making it a $(3\delta_{IE})$-noise variable and completing the proof.
\end{proof}

Using extended technical lemmas, it is possible to generalize our various properties to the case where we only have a prefix of the sketch. For example, the following claim generalizes case (i) of the \propattr{} property.
\begin{claim}\label{clm:final-erase}
  Suppose we have an overall sketch $s$ and we keep only a $\dsketch' \le \dsketch$ prefix of it, $s'$. Consider an object $\thetastar$ which lay at depth $h(\thetastar)$ and had effective weight $w_\thetastar$ in the original sketch $s$. Suppose no other objects in $s$ are also produced by $M(\thetastar)$. We can produce a vector estimate of the attribute vector $x_\thetastar$ from $s'$, which with high probability has an additive error of at most $O(2^{3h(\thetastar)} \delta_E / w_\thetastar)$ in each coordinate.
\end{claim}

\begin{proofsketch}
  We have an overall sketch $s$ and a version $s'$ where the last $(\dsketch - \dsketch')$ coordinates have been replaced with zeros. We compute $\beta e_j^T R_{M(\thetastar), 1}^T s'$. From our proof of case (i) of Theorem~\ref{thm:final-attr}, there are three things to analyze: the primary signal sub-term, the other sub-terms, and the other terms.
  
  The primary signal sub-term is now the following.
  \[
    \inneri{\dsketch'}
           {\frac12 R_{M(\thetastar), 1} x_\thetastar + \frac12 R_{M(\thetastar), 2} e_1}
           {2 R_{M(\thetastar), 1} e_j}
  \]
  
  Using $\dsketch'$-prefix isometry, we can show that the first half is a (scaled-down) of $e_j^T x_\thetastar$ plus (scaled-down) small noise. Using a generalization of Lemma~\ref{lem:diff-desync}, we can show that the second half is (scaled-down) small noise.
  
  When analyzing the other sub-terms, we use the $\dsketch'$-prefix desynchronization of the final matrix in $\tilde{R}$ to show the the sub-term is just (scaled-down) small noise. Finally, when analyzing the other terms, we use the $\dsketch'$-prefix desynchronization of $R_{M(\thetastar), 1}$ to show that the term is just (scaled-down) small noise.
\end{proofsketch}
\section{Proofs for Dictionary Learning}
\label{subsec:dict_learning_appendix}

\subsection{Proof of Network Learnability}

In this subsection, we prove Theorem~\ref{thm:applied-dictionary-learning}. Recall Theorem~\ref{thm:applied-dictionary-learning}:

\applieddictionarylearning*

\begin{proof}
  Our plan is to use Theorem~\ref{thm:dictionary-learning} to unwind our sketching. Precisely, we invoke it with $N$ as thrice the maximum between the number of modules and the number of objects in any communication graph, $\utildeS$ as $(\alpha / \beta) \log N = \poly(N)$, $H$ as three times the depth of our modular network, and $\epsilon_H$ as $\min(\epsilon / 4, 1 / 4) 2^{-H} w$. This yields a sketch dimension $\dsketch \le \poly(1 / \epsilon_H) \log N \log \poly(N)$ which is $\poly(1 / w, 1 / \epsilon) \log^2 N$ as claimed. Additionally, it yields a base number of samples $S = \poly(N)$ and a sequence of $\ell_\infty$ errors $\epsilon_1 \le \epsilon_2 \le \cdots \le \epsilon_{H-1} \le \epsilon_H$. These are guarantees on an algorithm $A$ which accepts an $h \in [3 \cdot \text{ network depth}]$ as well as $S_h = S N^{h-1}$ samples with up to $\epsilon_h$ $\ell_\infty$ error and $O(\sqrt{\dsketch})$ $\ell_1$ error.
  
  This base number of samples is why we will require $\poly(N)$ overall sketches. By construction, overall sketches are just input subsketches of the output psuedo-object. We know that no communication graph contains more than $N / 3$ objects by the definition of $N$, so we can write all the overall sketches in the form:
  \[
    \overallsketchx = \sum_{i=1}^{N} x_i + R_i x_i
  \]
  where $x_i$ is $(w_i / 2) \objectsketchx(\theta_i)$ if there actually was an $i^{th}$ input and the zero vector otherwise. We want to treat $\sum_i x_i$ as $\ell_1$ noise, but how much do they contribute? We can only tolerate them contributing at most $O(\sqrt{\dsketch})$ $\ell_1$ norm. In fact, we claim that any sketch vector has at most $O(1)$ $\ell_2$ norm, which would imply the desired statement by Cauchy-Schwartz and noticing that we are taking a convex combination of sketch vectors. We prove an $\ell_2$ bound on sketch vectors as in \splitter{Appendix~\ref{apx:final-sketch}}{the supplementary material}, but roughly the idea is that our error parameter $\delta$, which we know to be $O(\frac{\log N}{\sqrt{\dsketch}})$, needs to be less than the constant $\frac{1}{8H}$ (recall that our network depth is constant). Since our sketch dimension already scaled with $\log^2 N$, this is true up to making our sketch dimension a constant factor larger. We actually have no additional $\ell_\infty$ noise, which is under the $\epsilon_1$ requirement. Hence, we can safely apply $A$ and retrieve $(w_i / 2)$ times our object sketches (and some zero vectors), up to $\epsilon_2$ $\ell_\infty$ noise. Note that we now have a total of $\utildeS N$ vectors, since the procedure turns one vector into $N$ vectors.
  
  If any of our retrieved vectors have less than $\epsilon_2$ $\ell_\infty$ norm, we declare them garbage and any further vectors that dictionary learning recovers from them as garbage. When learning modules, we will not use any garbage vectors. Now, we have several approximate, scaled object sketches and several garbage vectors. We note that the scaled object sketches, by definition, are of the form:
  \[
    w \cdot \objectsketchx(\theta) = w \ptransparentx{R_{M(\theta), 0}} x \qquad \left( + \sum_{M \not = M(\theta)} R_{M, 0} \vec{0} \right)
  \]
  where $w$ is some weight from the previous step and $x$ is $w$ times a tuple sketch. Undoing the first matrix off an object sketch is exactly like the previous step; we handle the transparent matrix by noting that all sketches have constant $\ell_2$ norm with high probability and hence it's a matter of $\ell_1$ noise for our algorithm. Our $\epsilon_2$ $\ell_\infty$ noise turns into $\epsilon_3$ $\ell_\infty$ noise. The garbage vectors can be thought of as $\ell_\infty$ noise plus the zero vector, which also decomposes nicely:
  \[
    \vec{0} = \sum_M R_{M, 0} \vec{0}
  \]
  so the dictionary learning procedure can safely handle them. There are at most $N / 3$ matrices involved (we can imagine the problem as padded with zero matrices and additional zero vectors). We recover $\utildeS N^2$ vectors, which are attribute subsketches, input subsketches, and garbage vectors.
  
  The process of undoing attribute subsketches and input subsketches is why we need $N$ to be thrice the maximum between the number of modules and the number of objects in any communication graph. For attribute subsketches, there are up to two matrices for every module at this depth. For input subsketches, there is possibly a tuple matrix for every object at this depth. Still, there are at most $N$ matrices involved, so we can recover our $\utildeS N^3$ vectors, which are object attribute vectors, $e_1$'s, object sketches, and garbage vectors.
  
  Unlike previous steps, where we had one type of vector (and garbage vectors), we now have \emph{three types of vectors} that we want to distinguish; we can only recurse on object sketches because we cannot guarantee that object attribute vectors or $e_1$ can be written as the correct matrix-vector product form. We can get some helpful information by looking at which vectors produced which vectors in our dictionary learning procedure. By construction, we know that one object sketch gets turned into an attribute subsketch and an input subsketch. The former gets turned into an attribute vector and $e_1$ while the latter gets turned into a series of object sketches. For example, if the input subsketch produces at least three non-garbage vectors, then we know which subsketch was which (the attribute subsketch only produces up to two non-garbage vectors with high probability). Of course, this idea won't help us distinguish in general; an object might only have two input objects! The better test is to see which subsketch produces a vector that looks more like $e_1$. To begin with, it's worth noting that the two subsketches have the same scaling: the effective weight of object $\theta$, $w_\theta$, times $2^{-2h(\theta)+2}$.
  
  Now, let's consider how small the first coordinate of the (scaled) $e_1$ vector can be. It picks up an additional $1/2$ scaling factor, so the vector we recover is $w_\theta 2^{-2h(\theta)+1} e_1$ up to $\epsilon_{3h(\theta)-2}$ additional $\ell_\infty$ error. We know that $\epsilon_{3h(\theta)-2} < \epsilon_H < w 2^{-H}$, so the first coordinate is at least:
  \[
    w_\theta 2^{-2h(\theta)+1} - w 2^{-H}
  \]
  
  Now, let's consider how large the first (or any) coordinate of any object sketch could be. Due to \splitter{Theorem~\ref{thm:block-random-desync}}{the supplementary material}, we know that any (at most unit in length) vector has a first coordinate of at most $O(\frac{\log N}{\sqrt{\dsketch}})$ after one of our block random matrices is applied to it. Furthermore, applying the various transparent matrices $\ptransparentx{R}$ throughout our sketch increases this by at most a constant factor. The object sketch is just a convex combination of such vectors, so it also has at most $O(\frac{\log N}{\sqrt{\dsketch}})$ in its first coordinate. Since there is up to $w 2^{-H}$ additional $\ell_\infty$ error in the recovery process and this vector is scaled too, we know that the recovered object sketch vector has first coordinate at most:
  \[
    w_\theta 2^{-2h(\theta)+2} O(\tfrac{\log N}{\sqrt{\dsketch}}) + w 2^{-H}
  \]
  
  We require that $O(\tfrac{\log N}{\sqrt{\dsketch}}) < \frac18$. As already discussed, since our sketch dimension already scaled with $\log^2 N$, this amounts to making the sketch dimension at most a constant factor larger. Now, our recovered object sketch vector has first coordinate at most:
  \[
    w_\theta 2^{-2h(\theta)+1} / 4 + w 2^{-H}
  \]
  
  Whether our two cases are distinguishable still depends on how $\theta$'s effective weight $w_\theta$ relates to the goal weight $w$. We could clearly distinguish the two if  $w_\theta 2^{-2h(\theta)+1}$ was at least $4 w 2^{-H}$, since then the first case would yield a value of at least $3 w 2^{-H}$ and the second case a value of at most $2 w 2^{-H}$. Hence our test is the following: if some vector produced by the object sketch has a first coordinate of at least $3 w 2^{-H}$, then take the one with the largest first coordinate and decide that it was produced by the object attribute subsketch. If no vector produced by the object sketch has a first coordinate of at least $3 w 2^{-H}$, then swap them all for zero vectors and declare them garbage.
  
  Notice that if a recovered object sketch vector manages to have a first coordinate value of at least $3 w 2^{-H}$, then we know that $w_\theta 2^{-2h(\theta)+1}$ is at least $8 w 2^{-H}$ and hence the scaled $e_1$ still beats all the recovered object sketch vectors. On the other hand, all the vectors are declared garbage, then we know that $w_\theta 2^{-2h(\theta)+1} < 4 w 2^{-H}$ and hence $w_\theta < w$ (i.e. we don't need to recover this object or any of its children). Hence our procedure works and we can continue to safely recurse.
  
  The result of this procedure is that for each module (we can tell which vectors belong to which modules due to shared final matrices) we recover unordered pairs of scaled (attribute vector, $e_1$), at least every time the module produces an object with at least $w$ weight (possibly more). We'll consider the one with larger first coordinate to be $e_1$, and divide the other vector by the first coordinate of $e_1$ to unscale it.
  
  Note that both vectors have been scaled by $w_\theta 2^{-2h(\theta)+1}$ and then face at most $\epsilon_H$ additional $\ell_\infty$ noise. note that we chose a $\epsilon_H < (\epsilon / 4) 2^{-H} w$ and no vector could have passed our test unless $w_\theta 2^{-2h(\theta)+1} \ge 2 w 2^{-H}$, i.e. $w / w_\theta \le 2^{H-2h(\theta)}$. Combining these facts, we know that a single instance of the $\ell_\infty$ noise messes up any (unscaled) coordinate by at most:
  \begin{align*}
    2^{2h(\theta)-1} \epsilon_H / w_\theta
      &\le 2^{2h(\theta)-1} (\epsilon / 4) 2^{-H} w / w_\theta \\
      &\le 2^{2h(\theta)-1} (\epsilon / 4) 2^{-H} 2^{H-2h(\theta)} \\
      &= (\epsilon / 8)
  \end{align*}
  
  There are two cases: (i) we consider the correct vector to be $e_1$ and (ii) we consider the object attribute vector to be $e_1$. In case (i), our object attribute vector is scaled by a number which is within a multiplicative $(1 \pm \epsilon / 8)$ of correct. Additionally, its noise makes it off by an additive $\epsilon / 8$ in $\ell_\infty$ distance. Since we capped $\epsilon$ to be at most one, the multiplicative difference is at most $(9/8) / (1 - \epsilon / 8) - 9/8 \le \frac{9}{56} \epsilon$, for a total difference of $\frac{2}{7} \epsilon$ $\ell_\infty$ distance.
  
  In case (ii), the original attribute vector had to have first coordinate at least $1 - (\epsilon / 4)$. Since it has an $\ell_1$ norm of at most one, we know that it is at most $\epsilon / 4$ in $\ell_\infty$ distance from $e_1$. We imagine that the noised version of $e_1$ is it; by triangle inequality this choice is at most $\frac{3}{8} \epsilon$ from it. We again pick up at most a $\frac{9}{56} \epsilon$ error from incorrect scaling, for a total of at most $\frac{15}{28} \epsilon$ $\ell_\infty$ error.
  
  We conclude the proof of the first half of our theorem by noting that since we started with $(\alpha / \beta) \log N$ samples, with high probability we will have at least $\alpha$ samples where we recover an input/output pair for $M^\star$ within the robust module learnability conditions.

  Regarding fixed communication graphs, we can use this same algorithm to detect the entire path to an object with high probability (i.e. what modules produce each node along the path) if the object has effective weight $w$ along this path. This means we can identify the entire sub-graph of objects which have effective weight $w$ ``often enough''.
\end{proof}

\subsection{Proof of Recursable Dictionary Learning}

In this subsection, we prove Theorem~\ref{thm:dictionary-learning}, which we restate next.
\recursabledictionarylearning*


We point out that the $\ell_{\infty}$ bound on the recovery error in Theorem~\ref{thm:dictionary-learning} implies that the $x$ vectors can be recovered exactly if they are quantized and the recovery error $\epsilon_{h+1}$ is small enough.


\paragraph{Notation.}
We next provide some notation that will be used in the proof of Theorem~\ref{thm:dictionary-learning}. Let $H$ be a given positive constant and let $S \le \poly(N)$ be a base number of samples. Fix $h \in [H-1]$ and let $S_h = S N^{h-1}$.
\begin{itemize}
    \item For each sample index $i \in [S_h]$, we divide the $\dsketch$-dimensional real vector $y^{(i)}$ into $\dsketch/b$ contiguous blocks each of length $b$, with the $\ell$th block being $y^{(i)}[(\ell-1) b+1: ~ \ell b + 1]$ for each $\ell \in [\dsketch/b]$.
    \item For any two points $x,y \in \mathbb{R}^n$, recall that their $\ell_{\infty}$-distance is defined as
    $$ d_{\infty}(x,y) := \|y-x\|_{\infty} = \max_{i \in [n]} |y_i-x_i|.$$
    Moreover, we define the symmetric $\ell_{\infty}$-distance between $x$ and $y$ as $\min(d_{\infty}(x,y), d_{\infty}(x,-y))$.
    \item For any vector $y \in \mathbb{R}^n$ and any finite subset $L$ of $\mathbb{R}^n$, we denote the $\ell_{\infty}$-distance of $y$ from $L$ by
    $$d_{\infty}(y,L) := \min_{x \in L} \|y-x\|_{\infty}.$$
    If this distance is at least $\tau$, then we say that $y$ is $\tau$-far in $\ell_{\infty}$-distance from the set $L$.
    \item For any $y, y' \in \mathbb{R}^n$, we define their Hamming distance as $\Delta(y,y') := |\{j \in [n]: y_j \neq y'_{j}\}|$ and the symmetric Hamming distance as $\overline{\Delta}(y,y') = \min(\Delta(y,y'), \Delta(y,-y'))$.
\end{itemize}

\paragraph{Algorithm.}

\paragraph{Analysis.}

Theorem~\ref{thm:dictionary-learning} follows from Theorem~\ref{th:recovery_alg_guarantees} below which proves the guarantees on the operation of Algorithm~\ref{alg:recovery}.

\begin{theorem}[Guarantees of Algorithm~\ref{alg:recovery}]\label{th:recovery_alg_guarantees}
  Let $b$, $q$ and $\dsketch$ satisfy $b \ge \poly(\log N, \log \dsketch, 1/\epsilon_H)$, $q \ge \poly(\log N, \log \dsketch, 1/\epsilon_H) / \sqrt{\dsketch}$, and $\dsketch \ge \poly(1/\epsilon_H, \log N, \log S)$. For any unknown vectors $x^{(1)}, ..., x^{(S_h)} \in \mathbb{R}^{\dsketch \cdot N}$ with $\ell_2$-norm at most $O(1)$, if we draw $R_1, ..., R_N \sim \distribution(b, q, \dsketch)$ and receive $S_h$ noisy samples $y^{(k)} \coloneqq [R_1 R_2 \cdots R_N] x^{(k)} + z^{(k)}_1 + z^{(k)}_\infty$ where each $z^{(k)}_1 \in \rsketch$ is noise with $\norm{z^{(k)}_1}{1} \le O(\sqrt{\dsketch})$ (independent of our random matrices) and each $z^{(k)}_\infty \in \rsketch$ is noise with $\norm{z^{(k)}_\infty}{\infty} \le \epsilon_h$ (also independent of our random matrices), then Algorithm~\ref{alg:recovery}, taking as inputs $h$, $y^{(1)}$, $\dots$, $y^{(S_h)}$, and thresholds $\tau_1 = \Theta\bigg(\frac{\epsilon_{h+1}}{\sqrt{q \cdot \dsketch}}\bigg)$ and $\tau_2 = \Theta( \epsilon_{h+1}^2)$, runs in time $\poly(S_h, \dsketch)$ and outputs matrices $\hat{R}_1, \dots, \hat{R}_N \in \mathbb{R}^{\dsketch \times \dsketch}$ and vectors $\hat{x}^{(1)}, \dots, \hat{x}^{(S_h)} \in \mathbb{R}^{\dsketch \cdot N}$ that, with high probability, satisfy the following for some permutation $\pi$ of $[N]$:
  \begin{itemize}
    \item {\bf Matrix Recovery:} For every column $(i-1)\dsketch + j$ of the matrix $[R_1 R_2 \cdots R_N]$ (where $i \in [N]$ and $j \in [\dsketch]$) for which there exists a sample index $k^\star \in [S_h]$ such that $\abs{x^{(k^\star)}_{(i-1)\dsketch + j}} \ge \epsilon_{h+1}$, the $j$th column of $\hat{R}_{\pi(i)}$ is $0.2 \dsketch$-close in Hamming distance to the $j$th column of $R_i$. Moreover, all the other columns of $\hat{R}_{\pi(i)}$ are zero.
    \item {\bf Input Recovery:} For each $i \in [N]$ and $j \in [\dsketch]$, it is guaranteed that $|\hat{x}^{(k)}_{(\pi(i)-1)\dsketch + j} - x^{(k)}_{(i-1)\dsketch + j}| \le \epsilon_{h+1}$.
  \end{itemize}
\end{theorem}

We now define the permutation $\pi$ of $[N]$ that appears in the statement of Theorem~\ref{th:recovery_alg_guarantees}. For each $i \in [N]$, we let $\pi(i)$ be the first index $i^*$ set in line~\ref{line:i_star_def} of Algorithm~\ref{alg:recovery} for which $\hat{\sigma}_{i^*,m}$ is $10\epsilon_{h+1}$-close in Hamming distance to $\hat{\sigma}_{i,m}$. If no such $i^*$ exists, we define $\pi(i)$ in an arbitrary way that doesn't result in a collision.

\begin{algorithm}[H]
\caption{$\Dec(\overline{z}_c)$}
\begin{algorithmic}[1]\label{alg:decoding}
  \STATE {\bf Input:} A vector $\overline{z}_c$ on the hypercube $\{\pm \frac{1}{\sqrt{q \dsketch}}\}^{b/3}$. \\
  \STATE{\bf Output:} A pair $(j, f)$ where $j$ is an index in $[\dsketch]$ and $f$ is a sign. \\
  \item[] 
  \IF{the first coordinate of $\overline{z}_c$ is negative}
  \STATE Negate each coordinate of $\overline{z}_c$.
  \ENDIF
  \STATE Set $t$ to $\lceil \log_2 \dsketch \rceil$.
  \STATE Let $\sigma_c$ be the binary vector obtained from $\overline{z}_c[2:t+2]$ by replacing $-\frac{1}{\sqrt{q \dsketch}}$ by $0$ and $+\frac{1}{\sqrt{q \dsketch}}$ by $1$.
  \STATE Let $j$ be the non-negative integer whose binary representation is $\sigma_c$.
  \STATE Increment $j$ by $1$.
  \STATE Set $f$ to the sign of $\overline{z}_c[t+2]$.
  \RETURN $(j, f)$.
\end{algorithmic}
\end{algorithm}

\begin{algorithm}[H]
 \caption{Recursable Dictionary Learning}
\begin{algorithmic}[1]\label{alg:recovery}
 \STATE {\bf Input:} Positive integer $h$, observations $y^{(1)}, \dots, y^{(S_h)}$ and thresholds $\tau_1$ and $\tau_2$.
 \STATE{\bf Output:} Matrices $\hat{R}_1, \dots, \hat{R}_N \in \mathbb{R}^{\dsketch \times \dsketch}$ and vectors $\hat{x}^{(1)}, \dots, \hat{x}^{(S_h)} \in \mathbb{R}^{\dsketch \cdot N}$.\\ 
 \item[] 
  \FOR{each sample index $k \in [S_h]$}
     \FOR{each block index $\ell \in [\dsketch/b]$}
  \STATE Let $w_{k, \ell}$ be the product of $\sqrt{q/b}$ and the $\ell_1$-norm of the $\ell$th block of $y^{(k)}$.
  \IF{$w_{k, \ell}$ is not zero}
  \STATE Let $z_{k,\ell}$ be the coordinate-wise normalization of the $\ell$th block of $y^{(k)}$ by $w_{k, \ell}$.
  \STATE Let $z_{k, \ell, s}$, $z_{k, \ell, c}$ and $z_{k, \ell, m}$ be the first, middle and last $b/3$ coordinates of $z_{k,\ell}$ respectively.
  \ENDIF
  \ENDFOR
  \ENDFOR
 \STATE Initialize $n$ to zero, $\hat{R}_1, \dots, \hat{R}_N$ to the all-zeros matrices and $\hat{x}^{(1)}, \dots, \hat{x}^{(S_h)}$ to the all-zeros vectors.
 \FOR{each $i \in [N]$}
 \STATE Let $\hat{\sigma}_{i,m}$ be a vector in $\{\pm \frac{1}{\sqrt{q \dsketch}}\}^{b/3}$ that is initialized arbitrarily.
 \ENDFOR
 \FOR{each sample index $k \in [S_h]$}\label{line:first_for}
     \FOR{each block index $\ell \in [\dsketch/b]$}\label{line:second_for}
     \IF{$w_{k, \ell}$ is zero}\label{line:test_1}
     \STATE continue
     \ENDIF
      \IF{the $\ell$th block of $y^{(k)}$ has a coordinate whose absolute value is smaller than $\tau_1$ or larger than $\frac{2}{\sqrt{q \dsketch}}$}\label{line:test_2}
      \STATE continue
      \ENDIF
      \IF{$z_{k,\ell, s}$ is $\tau_2$-far in $\ell_{\infty}$-distance from the hypercube $\{\pm \frac{1}{\sqrt{q \dsketch}}\}^{b/3}$}\label{line:test_3}
      \STATE continue
      \ENDIF
       \STATE Let $S_{k,\ell}$ be the set of all block indices $\ell' \in [\dsketch/b]$ for which $w_{k, \ell'}$ is non-zero, all the coordinates of the $\ell'$th block of $y^{(k)}$ are between $\tau_1$ and $\frac{2}{\sqrt{q \dsketch}}$ and  $z_{k,\ell',s}$ is $2 \tau_2$-close in symmetric $\ell_{\infty}$-distance to $z_{k, \ell, s}$.
       \IF{the cardinality of $S_{k, \ell}$ is less than $(0.9)^3 q \dsketch$ }\label{line:test_4}
       \STATE continue
       \ENDIF
       \STATE Let $\overline{z}_{k, \ell}$ be the coordinate-wise rounding of $z_{k, \ell}$ to the hypercube $\{\pm \frac{1}{\sqrt{q \dsketch}}\}^{b}$.
         \STATE Let $\overline{z}_{k, \ell, s}$, $\overline{z}_{k, \ell, c}$, $\overline{z}_{k, \ell, m}$ be the most common first, middle and last $\frac{b}{3}$ coordinates of $\{\overline{z}_{k,\ell'}: ~ l' \in S_{k,\ell}\}$ respectively.\label{line:setting_most_common}
       \IF{there is $i \in [n]$ such that $\hat{\sigma}_{i,m}$ is $0.01b$-close in symmetric Hamming distance to $\overline{z}_{k, \ell, m}$}\label{line:begin_if_star}
       \STATE Set $i^*$ to $i$.
       \ELSE
       \STATE Increment $n$ by $1$ and set $i^*$ to the new value of $n$.\label{line:i_star_def}
       \STATE Set $\hat{\sigma}_{i^*,m}$ to the matrix signature $\overline{z}_{k, \ell, m}$.
       \ENDIF\label{line:end_if_star}
       \STATE\label{line:setting_j} Let $(j, f) = \Dec(\overline{z}_{k, \ell, c})$ be the decoded column index and sign respectively.
             \IF{the sign of the first coordinate of $\overline{z}_{k,\ell,m}$ is the opposite of $f$}
       \STATE Negate $w_{k, \ell}$ and each coordinate of $\overline{z}_{k, \ell}$.
       \ENDIF
 \IF{$\hat{R}_{i^*}[:, ~ j]$ is the all-zeros vector}
       \FOR{$\ell' \in S_{k, \ell}$}
       \STATE Set $\hat{R}_{i^*}[(\ell'-1)b+1: \ell' b + 1, ~ j]$ to $\overline{z}_{k, \ell}$.
       \ENDFOR
       \ENDIF
       \STATE Set $x^{(k)}_{(i^*-1)\dsketch + j}$ to $w_{k, \ell}$.\label{line:setting_x}
    \ENDFOR
\ENDFOR
\RETURN $\hat{R}_1, \dots, \hat{R}_N$ and $\hat{x}^{(1)}, \dots, \hat{x}^{(S_h)}$.
\end{algorithmic}
\end{algorithm}

Theorem~\ref{th:recovery_alg_guarantees} directly follows from Lemmas~\ref{le:recovery_soundness} and~\ref{le:recovery_completeness} below, applied in sequence. We point out that for simplicity of exposure, we chose large constants (which we did not attempt to optimize) in Lemmas~\ref{le:recovery_soundness} and~\ref{le:recovery_completeness}.

\begin{lemma}\label{le:recovery_soundness}
 For every absolute positive constant $t$ and every given positive real number $\epsilon_{h+1}$ satisfying $\epsilon_{h+1} \le \frac{1}{\dsketch^{1/\zeta}}$ where $\zeta$ is any positive constant, for any sufficiently large $b$ satisfying $b = \Theta \bigg(\frac{\log{N} + \log{\dsketch} + \log{S_h}}{\epsilon_{h+1}
^2} \bigg)$ and any sufficiently large $q$ satisfying $q = \Theta\bigg(\frac{\log{N} + \log{\dsketch} +\log{S_h}}{\epsilon_{h+1}^2 \cdot \sqrt{\dsketch}}\bigg)$, for $\epsilon_h = \Theta( \epsilon_{h+1}^c)$ where $c$ is a sufficiently large positive constant, with probability at least $1-N^{-t}$, the following holds during the operation of Algorithm~\ref{alg:recovery}:
 \begin{itemize}
     \item Whenever $i^*$ is set in lines \ref{line:begin_if_star}-\ref{line:end_if_star}, then $\hat{\sigma}_{i^*,m}$ is $0.001b$-close in Hamming distance to the matrix signature $\sigma_{\pi(i^*), m}$.
     \item If moreover $j$ is set in line \ref{line:setting_j}, then the $j$th column of $\hat{R}_{i^*}$ is $\frac{\epsilon_{h+1} \dsketch}{4}$-close in Hamming distance to the $j$th column of $R_{\pi(i^*)}$.
     \item If furthermore line~\ref{line:setting_x} is executed, then
     $$|x^{(k)}_{(\pi(i^*)-1)\dsketch + j} - x^{(k)}_{(i^*-1)\dsketch + j}| \le \epsilon_{h+1}.$$ 
 \end{itemize}
\end{lemma}

\begin{lemma}\label{le:recovery_completeness} For every absolute positive constant $t$ and every given positive real number $\epsilon_{h+1}$ satisfying $\epsilon_{h+1} \le \frac{1}{\dsketch^{1/\zeta}}$ where $\zeta$ is any positive constant, for any sufficiently large $b$ satisfying $b = \Theta \bigg(\frac{\log{N} + \log{\dsketch} + \log{S_h}}{\epsilon_{h+1}
^2} \bigg)$ and any sufficiently large $q$ satisfying $q = \Theta\bigg(\frac{\log{N} + \log{\dsketch} +\log{S_h}}{\epsilon_{h+1}^2 \cdot \sqrt{\dsketch}}\bigg)$ and for $\epsilon_h = \Theta(\epsilon_{h+1}^c)$ where $c$ is a sufficiently large positive constant, the following holds. With probability at least $1 - N^{-t}$, for all $k \in [S_h]$, $i \in [N]$ and $j \in [\dsketch]$:
\begin{itemize}
    \item If $|x^{(k)}_{(i-1)\dsketch+j}| \geq \epsilon_{h+1}$, the $j$th column of $\hat{R}_{\pi(i)}$ is $0.2 \dsketch$-close in Hamming distance to the $j$th column of $R_i$.
    \item It is the case that $|\hat{x}^{(k)}_{(\pi(i)-1)\dsketch + j} - x^{(k)}_{(i-1)\dsketch + j}| \le \epsilon_{h+1}$.
\end{itemize}
\end{lemma}

\paragraph{Proof Preliminaries.}
To prove Lemma~\ref{le:recovery_soundness}, we need the next two known theorems and the following two lemmas. 
Theorem~\ref{thm:Khintchine} below provides asymptotically tight bounds on the expected moments of a linear combination of independent uniform Bernouilli random variables. 
\begin{theorem}[Khintchine Inequality]\label{thm:Khintchine}
Let $\{\epsilon_n: n \in [N]\}$ be i.i.d. random variables with $\Pr[\epsilon_n = +1] = \Pr[\epsilon_n = -1] = 1/2$ for each $n \in [N]$, i.e., a sequence with Rademacher distribution. Let $0 < p < \infty$ and let $x_1, x_2, \dots, x_N \in \mathbb{R}$. Then, there exist positive absolute constants $A_p$ and $B_p$ (depending only on $p$) for which
\[
  A_p \cdot \left(\sum_{n=1}^N x_n^2\right)^{1/2}
    \le \left(\E{|\sum_{n=1}^N \epsilon_n x_n|^p}\right)^{1/p}
    \le B_p \cdot \left(\sum_{n=1}^N x_n^2\right)^{1/2}.
\]
\end{theorem}

The well-known Markov's inequality allows us to lower-bound the probability that a non-negative random variable is not much larger than its mean. The next theorem allows us to lower-bound the probability that a non-negative random variable is not much \emph{smaller} than its mean, provided that the random variable has finite variance.
\begin{theorem}[Paley-Zygmund Inequality]\label{thm:Paley_Zygmund}
If $Z \geq 0$ is a random variable with finite variance and if $0 \le \theta \le 1$, then
$$ \Pr[Z > \theta \cdot \E[Z] ] \geq (1-\theta)^2 \cdot \frac{\E[Z]^2}{\E[Z^2]}.$$
\end{theorem}

We will also need the following two lemmas.
\begin{lemma}[Separation Lemma]\label{lem:sum_indep_sym}
Let $v$ be a positive real number. Let $X$ and $Y$ be two random variables that are symmetric around $0$ and such that the absolute value of each of them is larger than $v$ with probability at least $\alpha$. Then, there exist real numbers $a_1 < a_2 < a_3 < a_4$ satisfying $\min(a_2-a_1, a_3-a_2, a_4-a_3) \geq v$ and such that the sum $X+Y$ lies in each of the intervals $(-\infty, a_1]$, $[a_2, a_3]$ and $[a_4, +\infty)$ with probability at least $\alpha/4$.
\end{lemma}
We will use the following multiplicative version of the Chernoff bound.
\begin{theorem}[Chernoff Bound -- Multiplicative Version]\label{thm:Chernoff_bd}
Suppose $X_1, \dots, X_n$ are independent random variables taking values in $\{0,1\}$. Let $X := \sum_{i \in [n]} X_i$ and let $\mu = \E[X]$. Then,
\begin{itemize}
\item For all $0 \le \delta \le 1$, $\Pr[X \le (1-\delta) \mu] \le e^{-\delta^2 \mu/2}$.
\item For all $\delta \geq 0$, $\Pr[X \geq (1+\delta) \mu] \le e^{-\delta^2 \mu /(2+\delta)}$.
\end{itemize}
\end{theorem}

\begin{proof}[Proof of Lemma~\ref{lem:sum_indep_sym}]
Define $m_{X,-}$ and $m_{X,+}$ to be the medians of the negative and positive parts (respectively) of the random variable $X$ conditioned on $|X| > v$ (note that $m_{X,-} = - m_{X,+}$ although we will not be using this fact in the proof).  Similarly, define $m_{Y,-}$ and $m_{Y,+}$ to be the negative and positive parts (respectively) of the random variable $Y$ conditioned on $|Y| \geq v$. Setting
$$a_1 = m_{X,-} + m_{Y,-},$$
$$a_2 = m_{X,-}+v,$$
$$a_3 = m_{Y,+}-v,$$
and
$$a_4 = m_{X,+} + m_{Y,+},$$
the statement of Lemma~\ref{lem:sum_indep_sym} now follows.
\end{proof}


We are now ready to prove the correctness of Algorithm~\ref{alg:recovery}.

\paragraph{Proof of Lemma~\ref{le:recovery_completeness}.}
Fix $k \in [S_h]$, $i \in [N]$ and $j \in [\dsketch]$, and assume that $|x^{(k)}_{(i-1)\dsketch+j}| \geq \epsilon_{h+1}$. For each $\ell \in [\dsketch/b]$, we consider the random vector
\begin{align*}
\mathrm{err}_{k, i, \ell, j} &:= \sum_{\genfrac{}{}{0pt}{}{i' \in [N], j' \in [\dsketch]:}{i' \neq i \text{ or }j' \neq j}} x^{(k)}_{(i'-1)\dsketch + j'} R_{i'}[(\ell-1)b+1: \ell b + 1, ~ j']\\ 
&+ z^{(k)}_\infty[(\ell-1)b +1: \ell b +1] + z^{(k)}_1[(\ell-1)b +1: \ell b +1].
\end{align*}
We think of the $b$-dimensional real vector $\mathrm{err}_{k, i, \ell, j}$ as containing the coordinate-wise errors incurred by estimating
$$x^{(k)}_{(i-1)\dsketch+j} R_i[(\ell-1)b+1: \ell b + 1, ~ j]$$
from the $\ell$th block of $y^{(k)}$. Let $T_{k,i,j}$ be the set of all block indices $\ell \in [\dsketch/b]$ for which $\eta_{i,\ell,j} = 1$ and
$$\|z^{(k)}_1[(\ell-1)b +1: \ell b +1]\|_{\infty} \le O(\frac{b}{\sqrt{\dsketch}}).$$
By an averaging argument, the Chernoff Bound (Theorem~\ref{thm:Chernoff_bd}) and the union bound, we have that for every fixed $k \in [S_h]$, $i \in [N]$ and $j \in [\dsketch]$, with probability at least $1-e^{-0.005 \cdot \frac{q \cdot \dsketch}{b}}$, it is the case that $|T_{k, i,j}| > 0.9 \cdot \frac{q \cdot \dsketch}{b}$.

Henceforth, we condition on a setting of $\{\eta_{i,\ell,j}: \ell \in [\dsketch/ b]\}$ for which $|T_{k, i,j}| > 0.9 \cdot \frac{q \cdot \dsketch}{b}$. We will upper-bound the probability, over the randomness of $\{\eta_{i', \ell, j'}: i' \in [N], j' \in [\dsketch], \text{ such that } i' \neq i \text{ or } j' \neq j\}$, that the $\ell_{\infty}$-norm of the random vector $\mathrm{err}_{k, i, \ell, j}$ is large, and then we will show, by again applying the Chernoff Bound (Theorem~\ref{thm:Chernoff_bd}), that for at least a $0.9$-fraction of the $\ell$'s in $T_{k,i,j}$, the $\ell_{\infty}$-norm is small. We have that
\begin{align*}
\| \mathrm{err}_{k, i, \ell, j} \|_{\infty} &\le |\sum_{\genfrac{}{}{0pt}{}{i' \in [N], j' \in [\dsketch]:}{i' \neq i \text{ or } j' \neq j}} x^{(k)}_{(i'-1)\dsketch+j'} \cdot R_{i'}[(\ell-1)b+1: \ell b + 1, ~ j']| +\epsilon_h+ O(\frac{b}{\sqrt{\dsketch}})\\ 
&:= \epsilon_{k,i,\ell,j},
\end{align*}
where the first inequality above uses the assumption that $\|z^{(k)}_\infty\|_{\infty} \le \epsilon_h$.
By the desynchronization property (Section~\ref{sec:desynch}), the fact that the $\ell_2$-norm of $x^{(k)}$ is guaranteed to be at most $O(1)$ for all $k \in [S_h]$, and the assumptions that $q \gg \frac{\sqrt{b \cdot \log{N}}}{\sqrt{\dsketch}}$ and $b \gg \log{N}$, the expected value of $\epsilon_{k,i,\ell,j}$ satisfies
\begin{align*}
\E[\epsilon_{k,i,\ell,j}] 
&\le \epsilon_h + O(\frac{b}{\sqrt{\dsketch}}).
\end{align*}
By Markov's inequality, we get that with probability at least $0.9$, it is the case that
\begin{equation}\label{eq:eps_bd}
\epsilon_{k,i,\ell,j} \le 10 \cdot \epsilon_1 +O(\frac{b}{\sqrt{\dsketch}}).
\end{equation}
Since the random variables in the set $\{\epsilon_{k,i,\ell,j}: \ell \in T_{k, i,j}\}$ are independent, the Chernoff Bound (Theorem~\ref{thm:Chernoff_bd}) implies that with probability at least $1-e^{-0.1^2 \cdot 0.9 \cdot |T_{k, i,j}|/2} \geq 1- e^{-0.004 \cdot \frac{q \dsketch}{b}}$, for at least a $(0.9)^2$-fraction of the $\ell$'s in $T_{k, i,j}$, it is the case that $\epsilon_{k,i,\ell,j}$ satisfies Equation~(\ref{eq:eps_bd}) and hence $\| \mathrm{err}_{k,i,\ell,j} \|_{\infty} \le 10 \cdot \epsilon_h + O(\frac{b}{\sqrt{\dsketch}})$. For each of these values of $\ell$, the block $(k,\ell)$ will fail the tests in lines \ref{line:test_1} and \ref{line:test_2} of Algorithm~\ref{alg:recovery} as long as
\begin{equation}\label{eq:comp_lem_first_test}
\frac{\epsilon_{h+1}}{\sqrt{q \cdot \dsketch}} - 10 \cdot \epsilon_h - O(\frac{b}{\sqrt{\dsketch}}) > \tau_1.
\end{equation}
Moreover for each of these values of $\ell$, the block $(k,\ell)$ will fail the test in line \ref{line:test_3} of Algorithm~\ref{alg:recovery} as long as
\begin{equation}\label{eq:comp_lem_second_test}
\frac{10 \cdot \epsilon_h + O(\frac{b}{\sqrt{\dsketch}})}{\frac{\epsilon_{h+1}}{\sqrt{q \cdot \dsketch}} - 10 \cdot \epsilon_h - O(\frac{b}{\sqrt{\dsketch}})} < \tau_2.
\end{equation}
Furthermore for each of these values of $\ell$, we will also have that $|S_{k,\ell}| > (0.9)^3 \cdot \frac{q \cdot \dsketch}{b}$ (as long as Equation~(\ref{eq:comp_lem_first_test}) above holds) and hence the block $(k, \ell)$ will fail the test in line \ref{line:test_4} of Algorithm~\ref{alg:recovery}.

Thus, the block $(k, \ell)$ fails the tests in lines \ref{line:test_1}, \ref{line:test_2}, \ref{line:test_3} and \ref{line:test_4} of Algorithm~\ref{alg:recovery} with probability at least $1 - 2 \cdot e^{-0.004 \cdot \frac{q \cdot \dsketch}{b}}$ as long as Equations~(\ref{eq:comp_lem_first_test}) and (\ref{eq:comp_lem_second_test}) above hold.
In this case, a vector that is at a (symmetric relative) Hamming distance of at most $0.1$ from $r_{m,j}$ will be added to the set of atoms (if no such vector was already present in the set). Moreover, the value of $x^{(k)}_{(i-1)\dsketch+j}$ will be recovered up to an absolute error of $\epsilon_{h+1}$ as long as
\begin{equation}\label{eq:completeness_lem_x_recovery}
\frac{10 \cdot \epsilon_h + O(\frac{b}{\sqrt{\dsketch}})}{\frac{\epsilon_{h+1}}{\sqrt{q \cdot \dsketch}} - 10 \cdot \epsilon_h - O(\frac{b}{\sqrt{\dsketch}})} < \epsilon_{h+1}.
\end{equation}

By a union bound, we get that with probability at least $1 - 2 \cdot S_h \cdot \dsketch \cdot N \cdot e^{-0.004 \cdot \frac{q \dsketch}{b}}$, the above recovery guarantees simultaneously hold for all $k \in [S_h]$ and all $i \in [N]$ and $j \in [\dsketch]$ for which $|x^{(k)}_{(i-1)\dsketch+j}| \geq \epsilon_{h+1}$. This probability is at least the desired $1-N^{-t}$ in the statement of Lemma~\ref{le:recovery_completeness} as long as
\begin{equation}\label{eq:completeness_lem_prob_bd}
2 \cdot S_h \cdot \dsketch \cdot N \cdot e^{-0.004 \cdot \frac{q \cdot \dsketch}{b}} < N^{-t}.
\end{equation}
Finally, we note that Equations~(\ref{eq:comp_lem_first_test}), (\ref{eq:comp_lem_second_test}), (\ref{eq:completeness_lem_x_recovery}) and (\ref{eq:completeness_lem_prob_bd}) above are all simultaneously satisfied (assuming that $t$ is a given positive absolute constant) if we let $c$ be a sufficiently large positive constant and set
$$ \tau_1 = \Theta\bigg(\frac{\epsilon_{h+1}}{\sqrt{q \cdot \dsketch}}\bigg),$$
$$\text{ any } \tau_2 = \Theta( \epsilon_{h+1}^2),$$
$$ \epsilon_h = \Theta(\epsilon_{h+1}^c),$$
$$ b = \Theta \bigg(\frac{\log{N} + \log{\dsketch} + \log{S_h}}{\epsilon_{h+1}
^2} \bigg),$$
and
$$q = \Theta\bigg(\frac{\log{N} + \log{\dsketch} +\log{S_h}}{\epsilon_{h+1}^2 \cdot \sqrt{\dsketch}}\bigg),$$
where $b$ and $q$ are sufficiently large.

\paragraph{Proof of Lemma~\ref{le:recovery_soundness}.}
We start by considering the special case where the ``$\ell_1$ error'' $z^{(k)}_1$ is zero (for all $k \in [S_h]$). We will later generalize the proof to the case where we only assume that $\norm{z^{(k)}_1}{1} \le O(\sqrt{\dsketch})$.

For each $k \in [S_h]$ and $\ell \in [\dsketch/b]$, we define $x^{(k,\ell)} \in \mathbb{R}^{N \cdot \dsketch}$ by setting $x^{(k, \ell)}_{(i-1)\dsketch+j} = x^{(k)}_{(i-1)\dsketch+j} \cdot \mathbbm{1}[\eta_{i,\ell,j} = 1]$ for each $i \in [N]$ and $j \in [\dsketch]$. For each coordinate $u \in [b]$, we consider the scalar random variable
$$ |y^{(k)}[(\ell-1)b+u]| = |\sum_{i \in [N], j \in [\dsketch]} x^{(k,\ell)}_{(i-1)\dsketch+j} R_i[(\ell-1)b+u, j] + z^{(k)}_\infty[(\ell-1)b +u]|.$$
Let $\lambda \geq 1$ be a real parameter to be specified later. If $\|x^{(k, \ell)}\|_2 \le \epsilon_{h+1}/\lambda$, then for each $u \in [b/3]$, Khintchine's Inequality (Theorem~\ref{thm:Khintchine}) with $p=1$, along with the fact that $\| z^{(k)}_\infty \|_{\infty} \le \epsilon_h$, imply that
\begin{align*}
\E[|y^{(k)}[(\ell-1)b+u]|] &\le \frac{B_2}{\sqrt{q \cdot \dsketch}} \cdot (\sum_{i \in [N], j \in [\dsketch]} (x^{(k,\ell)}_{(i-1)\dsketch+j})^2)^{1/2} + \epsilon_h\\ 
&\le \frac{B_2 \cdot \epsilon_{h+1}}{\lambda \cdot \sqrt{q \cdot \dsketch}} + \epsilon_h,
\end{align*}
where $B_2$ is a positive absolute constant. By Markov's inequality, we get that
$$ \Pr[ |y^{(k)}[(\ell-1)b+u]| > \frac{10 \cdot B_2 \cdot \epsilon_{h+1}}{\lambda \cdot \sqrt{q \cdot \dsketch}} + 10 \cdot \epsilon_h] \le 0.1.$$
Since the coordinates $u \in [b/3]$  are independent, the probability that all of them satisfy $|y^{(k)}[(\ell-1)b+u]| > 10 \cdot B_2 \cdot \epsilon_{h+1}/(\lambda \sqrt{q \dsketch}) + 10 \cdot \epsilon_h$ is at most $(0.1)^{b/3}$. Thus, the probability that block $(k, \ell)$ will pass the test in line~\ref{line:test_2} of Algorithm~\ref{alg:recovery} is at least $1 - (0.1)^{b/3}$ as long as
\begin{equation}\label{eq:soundness_pf_first_test}
\frac{10 \cdot B_2 \cdot \epsilon_{h+1}}{\lambda \cdot \sqrt{q \cdot \dsketch}} + 10 \cdot \epsilon_h < \tau_1. 
\end{equation}
By a union bound, we get that with probability at least
\begin{equation}\label{eq:soundness_pf_prob_union_bd_first_part}
1-S_h \cdot \frac{\dsketch}{b} \cdot (0.1)^{b/3},
\end{equation}
for all $k \in [S_h]$ and all $\ell \in [\dsketch/b]$ for which $\|x^{(k,\ell)}\|_2 \le \epsilon_{h+1}/\lambda$, block $(k,\ell)$ will pass the test in line~\ref{line:test_2} of Algorithm~\ref{alg:recovery}.

Henceforth, we assume that $\|x^{(k,\ell)}\|_2 > \epsilon_{h+1}/\lambda$. Assume moreover that for all $i \in [N]$ and $j \in [\dsketch]$ for which $\eta_{i, \ell, j} = 1$, it is the case that
\begin{equation}\label{eq:coordinate_bd_sq_ell_2}
(x_{(i-1)\dsketch+j}^{(k)})^2 \le g \cdot \sum_{\genfrac{}{}{0pt}{}{i' \in [N], j' \in [\dsketch]:}{i' \neq i \text{ or } j' \neq j}} (x^{(k)}_{(i'-1)\dsketch+j'})^2 \cdot \mathbbm{1}[\eta_{i', \ell, j'} = 1],
\end{equation}
where $g > 1$ is an absolute constant to be specified later. Equation~(\ref{eq:coordinate_bd_sq_ell_2}) is equivalent to saying that for all $i \in [N]$ and $j \in [\dsketch]$,
\begin{equation}\label{eq:coordinate_bd_sq_ell_2_equiv}
(x_{(i-1)\dsketch+j}^{(k, \ell)})^2 \le g \cdot \sum_{\genfrac{}{}{0pt}{}{i' \in [N], j' \in [\dsketch]:}{i' \neq i \text{ or } j' \neq j}} (x^{(k, \ell)}_{(i'-1)\dsketch+j'})^2.
\end{equation}

For each coordinate $u \in [b/3]$, we can then break the sum into two parts
\begin{align}
\sum_{\genfrac{}{}{0pt}{}{i \in [N],}{j \in [\dsketch]}} x^{(k,\ell)}_{(i-1)\dsketch+j} R_i[(\ell-1)b+u, j] &= \sum_{(i,j) \in \mathcal{W}} x^{(k,\ell)}_{(i-1)\dsketch+j} R_i[(\ell-1)b+u, j]\nonumber\\ 
&+ \sum_{(i,j) \in [N] \times [\dsketch] \setminus \mathcal{W}} x^{(k,\ell)}_{(i-1)\dsketch+j} R_i[(\ell-1)b+u, j],\label{eq:sum_two_parts}
\end{align}
where $\mathcal{W} \subseteq [N] \times [\dsketch]$ and such that each of these two parts has a weight vector with $\ell_2$-norm at least $\epsilon_{h+1} /(\lambda \cdot \sqrt{2 \cdot (g+1)})$, i.e.,
\begin{align*}
&\min\bigg(\sqrt{\sum_{(i,j) \in \mathcal{W}} (x^{(k,\ell)}_{(i-1)\dsketch+j})^2}, \sqrt{\sum_{(i,j) \in [N] \times [\dsketch] \setminus \mathcal{W}} (x^{(k,\ell)}_{(i-1)\dsketch+j})^2}\bigg)\\ 
&\geq \frac{\epsilon_{h+1}}{\lambda \cdot \sqrt{2 \cdot (g+1)}}.
\end{align*}
Such a partition can be obtained by sorting the coefficients $\{(x^{(k,\ell)}_{(i-1)\dsketch+j})^2: i \in [N], j \in [\dsketch]\}$ in non-increasing order and then including every other coefficient in the set $\mathcal{W}$ (which will thus contain the largest, $3$rd largest, $5$th largest etc. coefficients).
Applying Khintchine's Inequality (Theorem~\ref{thm:Khintchine}) with $p=1$ and $p=2$, we get that
\begin{align}
\frac{A_1}{\sqrt{q \cdot \dsketch}} \cdot \sqrt{\sum_{(i,j) \in \mathcal{W}} (x^{(k,\ell)}_{(i-1)\dsketch+j})^2} &\le \E[|\sum_{(i,j) \in \mathcal{W}} x^{(k,\ell)}_{(i-1)\dsketch+j} R_i[(\ell-1)b+u, j]|]\nonumber\\ 
&\le \frac{B_1}{\sqrt{q \cdot \dsketch}} \cdot \sqrt{\sum_{(i,j) \in \mathcal{W}} (x^{(k,\ell)}_{(i-1)\dsketch+j})^2}\label{eq:Khintchine_p_1}
\end{align}
and
\begin{align}\label{eq:Khintchine_p_2}
\frac{A_2}{\sqrt{q \cdot \dsketch}} \cdot \sqrt{\sum_{(i,j) \in \mathcal{W}} (x^{(k,\ell)}_{(i-1)\dsketch+j})^2} &\le \sqrt{\E[|\sum_{(i,j) \in \mathcal{W}} x^{(k,\ell)}_{(i-1)\dsketch+j} R_i[(\ell-1)b+u, j]|^2]}\nonumber\\ 
&\le \frac{B_2}{\sqrt{q \cdot \dsketch}} \cdot \sqrt{\sum_{(i,j) \in \mathcal{W}} (x^{(k,\ell)}_{(i-1)\dsketch+j})^2},
\end{align}
where $A_1$, $B_1$, $A_2$ and $B_2$ are some positive absolute constants. Applying the Paley–Zygmund Inequality (Theorem~\ref{thm:Paley_Zygmund}) to the random variable $Z = |\sum_{(i,j) \in \mathcal{W}} x^{(k,\ell)}_{(i-1)\dsketch+j} R_i[(\ell-1)b+u, j]|$ with $\theta = 0.5$, we get that $Z > 0.5 \cdot \E[Z]$ with probability at least $0.25 \cdot \frac{\E[Z]^2}{\E[Z^2]}$.
Note that for our setting of $Z$, Equation~(\ref{eq:Khintchine_p_1}) implies that $$\frac{\epsilon_{h+1} \cdot A_1}{\lambda \cdot \sqrt{2 \cdot (g+1) \cdot q \cdot \dsketch}} \le \frac{A_1}{\sqrt{q \cdot \dsketch}} \cdot (\sum_{(i,j) \in \mathcal{W}} (x^{(k,\ell)}_{(i-1)\dsketch+j})^2)^{1/2} \le \E[Z].$$
Moreover, Equation~(\ref{eq:Khintchine_p_2}) implies that $$\E[Z^2] \le \frac{B_2^2}{q \cdot \dsketch} \cdot \sum_{(i,j) \in \mathcal{W}} (x^{(k,\ell)}_{(i-1)\dsketch+j})^2.$$ Putting these inequalities together, we get that
$$|\sum_{(i,j) \in \mathcal{W}} x^{(k,\ell)}_{(i-1)\dsketch+j} R_i[(\ell-1)b+u, j]| > \frac{\epsilon_{h+1} \cdot A_1}{2 \cdot \lambda \cdot \sqrt{2 \cdot (g+1) \cdot q \cdot \dsketch}}$$
with probability at least
$$0.25 \cdot \frac{A_1^2 \cdot \sum_{(i,j) \in \mathcal{W}} (x^{(k, \ell)}_{(i-1)\dsketch+j})^2}{B_2^2 \cdot \sum_{(i,j) \in \mathcal{W}} (x^{(k, \ell)}_{(i-1)\dsketch+j})^2} = 0.25 \cdot \frac{A_1^2}{B_2^2}.$$
Thus, $v_{\mathcal{W}} := \sum_{(i,j) \in \mathcal{W}} x^{(k, \ell)}_{(i-1)\dsketch+j} R_i[(\ell-1)b+u, j]$ is a random variable that is symmetric around $0$ and whose absolute value is larger than $v := \epsilon_{h+1} \cdot A_1 /(2 \cdot \sqrt{2 \cdot (g+1) \cdot q \cdot \dsketch} \cdot \lambda)$ with probability at least $\gamma := 0.25 \cdot A_1^2 / B_2^2$. Applying the same reasoning to the second summand in Equation~(\ref{eq:sum_two_parts}), we deduce that the random variable $v_{[N] \times [\dsketch] \setminus \mathcal{W}} := \sum_{(i,j) \in [N] \times [\dsketch] \setminus \mathcal{W}} x^{(k, \ell)}_{(i-1)\dsketch+j} R_i[(\ell-1)b+u, j]$ is also a random variable that is symmetric around $0$ and whose absolute value is larger than $v$ with probability at least $\gamma$. Since the random variables $v_{\mathcal{W}}$ and $v_{[N] \times [\dsketch] \setminus \mathcal{W}}$ are independent, Lemma~\ref{lem:sum_indep_sym} then implies that their sum belongs to each of the $3$ intervals $(-\infty, a_1]$, $[a_2, a_3]$ and $[a_4, +\infty)$ with probability at least $\gamma/4$ where $\min(a_2-a_1, a_3-a_2, a_4-a_3) \geq v$. Thus, any normalization of this random variable by a positive scalar that has value smaller than $2 \sqrt{b/q}$ (in particular, normalizing by $w_{k,\ell}$ which has value smaller than $2 \sqrt{b/q}$ if block $(k,\ell)$ fails the test in line~\ref{line:test_2} of Algorithm~\ref{alg:recovery}) results in a value which is at an absolute distance of more than
$$\frac{v}{2} \cdot \frac{\sqrt{q}}{2\sqrt{b}} = \frac{\epsilon_{h+1} \cdot A_1}{8 \sqrt{2 \cdot (g+1) \cdot b \cdot \dsketch}}$$
from the set $\{\pm \frac{1}{\sqrt{q \dsketch}}\}$ with probability at least $\gamma/4 = A_1^2/(16 \cdot B_2^2)$. Since the $b/3$ coordinates are independent, we get that block $(k,\ell)$ fails the test in line~\ref{line:test_3} of Algorithm~\ref{alg:recovery} with probability at most $(A_1^2/(16 \cdot B_2^2))^{b/3}$ as long as
\begin{equation}\label{eq:soundness_pf_second_test}
\frac{\epsilon_{h+1} \cdot A_1}{8 \sqrt{2 \cdot (g+1) \cdot b \cdot \dsketch}} - \epsilon_h > \tau_2.
\end{equation}
By a union bound, we get that with probability at least
\begin{equation}\label{eq:soundness_pf_prob_union_bd_second_part}
1-N \cdot \frac{\dsketch}{b} \cdot (A_1^2/(16 \cdot B_2^2))^{b/3},
\end{equation}
for all $k \in [S_h]$ and all $\ell \in [\dsketch/b]$ for which $\|x^{(k,\ell)}\|_2 > \epsilon_2/\lambda$ and for which Equation~(\ref{eq:coordinate_bd_sq_ell_2_equiv}) is satisfied (for all $i \in [N]$ and all $j \in [\dsketch]$), block $(k,
\ell)$ will pass the test in line~\ref{line:test_3} of Algorithm~\ref{alg:recovery}.

Next, we still assume that $\|x^{(k,\ell)}\|_2 > \epsilon_{h+1}/\lambda$ but we assume that Equation~(\ref{eq:coordinate_bd_sq_ell_2_equiv}) is violated for some $i \in [N]$ and $j \in [\dsketch]$, i.e.,
\begin{equation}\label{eq:coordinate_bd_sq_ell_2_equiv_violated}
(x_{(i-1)\dsketch+j}^{(k, \ell)})^2 > g \cdot \sum_{\genfrac{}{}{0pt}{}{i' \in [N], j' \in [\dsketch]:}{i' \neq i \text{ or } j' \neq j}} (x^{(k, \ell)}_{(i'-1)\dsketch+j'})^2.
\end{equation}
This implies that
\begin{equation}\label{eq:low_bd_dominant}
(x_{(i-1)\dsketch+j}^{(k, \ell)})^2 > \frac{g}{g+1} \cdot \| x^{(k,\ell)} \|_2^2
\end{equation}
and
\begin{equation}\label{eq:up_bd_non_dominant}
\sum_{\genfrac{}{}{0pt}{}{i' \in [N], j' \in [\dsketch]:}{i' \neq i \text{ or } j' \neq j}} (x^{(k, \ell)}_{(i'-1)\dsketch+j'})^2 < \frac{1}{g+1} \cdot \| x^{(k,\ell)} \|_2^2.
\end{equation}
We now consider the rounding $\overline{y}^{(k)}$ of $y^{(k)}$ to the hypercube $\{\pm \frac{1}{\sqrt{q \dsketch}}\}^{\dsketch}$ obtained by setting each coordinate of $y^{(k)}$ to $+\frac{1}{\sqrt{q \dsketch}}$ if it is non-negative and to $-\frac{1}{\sqrt{q \dsketch}}$ otherwise. We argue that $\overline{y}^{(k)}[(\ell-1)b+1:\ell b +1]$ is at a Hamming distance of at most $0.05b$ from the vector $R_i[(\ell-1)b+1:\ell b +1, j]$ (note that Equation~(\ref{eq:coordinate_bd_sq_ell_2_equiv_violated}) and the assumption that $\|x^{(k,\ell)}\|_2 > \alpha/\lambda$ imply that $\eta_{i, \ell, j} = 1$). To prove this, we consider, for every coordinate $u \in [b]$, the random variable
\begin{equation*}
\Delta_{k,i,j,\ell,u} = |\sum_{\genfrac{}{}{0pt}{}{i' \in [N], j' \in [\dsketch]:}{i' \neq i \text{ or } j' \neq j}} x^{(k, \ell)}_{(i'-1)\dsketch+j'} R_i[(\ell-1)b+u, j] + z^{(k)}_\infty[(\ell-1)b +u]|.
\end{equation*}
Note that the $u$-th coordinate $\overline{y}^{(k)}[(\ell-1)b+u]$ of $\overline{y}^{(k)}$ is equal to $R_i[(\ell-1)b+u, j]$ if
\begin{equation}\label{eq:gap_ineq_x_delta}
\frac{|x_{(i-1)\dsketch+j}^{(k, \ell)}|}{\sqrt{q \cdot \dsketch}} > \Delta_{k,i,j,\ell,u}.
\end{equation}
Thus, to show that the Hamming distance between $\overline{y}^{(k)}[(\ell-1)b+1:\ell b +1]$ and $R_i[(\ell-1)b+1:\ell b +1, j]|]$ is at most $\epsilon_{h+1} b /4$, it suffices to show that for at least a $(1-\epsilon_2/4)$-fraction of the coordinates $u \in [b]$ it is the case that $|x_{(i-1)\dsketch+j}^{(k, \ell)}| > \Delta_{k,i,j,\ell,u} \cdot \sqrt{q \cdot \dsketch}$. Since $\|z^{(k)}_\infty\|_{\infty} < \epsilon_h$, we have that
\begin{align}
\E[\Delta_{k,i,j,\ell,u}] &\le |\sum_{\genfrac{}{}{0pt}{}{i' \in [N], j' \in [\dsketch]:}{i' \neq i \text{ or } j' \neq j}} x^{(k, \ell)}_{(i'-1)\dsketch+j'} R_i[(\ell-1)b+u, j]| + \epsilon_h\nonumber\\ 
&\le \frac{B_1}{\sqrt{q \cdot \dsketch}} \cdot \sqrt{\sum_{\genfrac{}{}{0pt}{}{i' \in [N], j' \in [\dsketch]:}{i' \neq i \text{ or } j' \neq j}} (x^{(k, \ell)}_{(i'-1)\dsketch+j'})^2} + \epsilon_h\nonumber\\ 
&\le \frac{B_1}{\sqrt{q \cdot \dsketch}} \cdot \frac{\| x^{(k,\ell)} \|_2}{\sqrt{g+1}} + \epsilon_h,\label{eq:expected_coordinate_dev}
\end{align}
where the penultimate inequality follows from Khintchine's Inequality (Theorem~\ref{thm:Khintchine}) with $p=1$ and the last inequality follows from Equation~(\ref{eq:up_bd_non_dominant}). By Markov's inequality, we get that for each fixed $u \in [b]$, with probability at least $1-\epsilon_{h+1}/8$,
\begin{equation*}
\Delta_{k,i,j,\ell,u} \le \frac{80 \cdot B_1 \cdot \| x^{(k,\ell)} \|_2}{\sqrt{q \cdot \dsketch} \cdot \epsilon_{h+1} \cdot \sqrt{g+1}} + \frac{80 \cdot \epsilon_h}{\epsilon_{h+1}},
\end{equation*}
and Equation~(\ref{eq:low_bd_dominant}) then implies that in this case Equation~(\ref{eq:gap_ineq_x_delta}) is satisfied as long as
\begin{equation}\label{eq:soundness_pf_last_case_r}
\sqrt{\frac{g}{g+1}} \cdot \| x^{(k,\ell)} \|_2 > \frac{80 \cdot B_1 \cdot \| x^{(k,\ell)} \|_2}{\sqrt{q \cdot \dsketch} \cdot \epsilon_{h+1} \cdot \sqrt{g+1}} + \frac{80 \cdot \epsilon_h}{\epsilon_{h+1}}.
\end{equation}
Since the first $b/3$ coordinates are independent, the Chernoff Bound (Theorem~\ref{thm:Chernoff_bd}) hence implies that the Hamming distance between the first $b/3$ coordinates of $\overline{y}^{(k)}[(\ell-1)b+1:\ell b +1]$ and the first $b/3$ coordinates of $R_i[(\ell-1)b+1:\ell b +1, j]|]$ is at most $\epsilon_{h+1}b/4$ with probability at least
\begin{equation}\label{eq:prob_lb_Chernoff}
1- e^{-\epsilon_{h+1}^2 \cdot (1-\frac{\epsilon_{h+1}}{8}) \cdot \frac{b}{128}}.    
\end{equation}
We now briefly explain why in this case, $\overline{z}_{k, \ell, c}$ and $\overline{z}_{k, \ell, m}$ (which are set in line~\ref{line:setting_most_common} of Algorithm~\ref{alg:recovery}) will be exactly equal to the middle and last $b/3$ coordinates of $R_i[(\ell-1)b+1:\ell b +1, j]|]$ respectively. We describe the argument for the middle coordinates (the same holds for the last coordinates). By Equation~(\ref{eq:expected_coordinate_dev}), we have that the expected sum of the $\Delta_{k,i,j,\ell,u}$ values for all middle coordinates $u$ is at most
$$\frac{b}{3} \cdot (\frac{B_1}{\sqrt{q \cdot \dsketch}} \cdot \frac{\| x^{(k,\ell)} \|_2}{\sqrt{g+1}} + \epsilon_h).$$
By Markov's inequality and the Chernoff bound, we thus get that with probability at least $1-1/\poly(N)$, it is the case that $\overline{z}_{k, \ell, c}$ is exactly equal to the middle $b/3$ coordinates of $R_i[(\ell-1)b+1:\ell b +1, j]|]$, as long as:
\begin{equation}\label{eq:middle_and_last}
100 \cdot \frac{b}{3} \cdot (\frac{B_1}{\sqrt{q \cdot \dsketch}} \cdot \frac{\| x^{(k,\ell)} \|_2}{\sqrt{g+1}} + \epsilon_h) < \sqrt{\frac{g}{g+1}} \cdot \| x^{(k,\ell)} \|_2,
\end{equation}
We also note that as long as
\begin{equation}\label{eq:soundness_pf_last_case_x}
(1-\frac{\epsilon_{h+1}}{4}) \cdot (80 \cdot \frac{B_1}{\sqrt{q \cdot \dsketch}} \cdot \frac{\| x^{(k,\ell)} \|_2}{\sqrt{g+1}} + 80 \cdot \epsilon_h) + \frac{\epsilon_{h+1}}{4} \cdot \frac{2}{\sqrt{q \cdot \dsketch}} < \epsilon_{h+1},
\end{equation}
an identical application of the Chernoff bound implies that with probability at least the quantity in Equation~(\ref{eq:prob_lb_Chernoff}), if line~\ref{line:setting_x} is executed when Algorithm~\ref{alg:recovery} examines block $(k,\ell)$, then 
$|x^{(k)}_{(i^*-1)\dsketch + j} - x^{(k)}_{(i-1)\dsketch + j}| \le \epsilon_{h+1}$ (we note that we are here using the fact that the block $(k, \ell)$ has failed the test in line~\ref{line:test_2} of Algorithm~\ref{alg:recovery}, hence the multiplicative factor of $2/\sqrt{q \cdot \dsketch}$ on the left-hand side of Equation~(\ref{eq:soundness_pf_last_case_x})). By a union bound, we get that these properties simultaneously hold for all blocks $(k,\ell) \in [S_h] \times [\dsketch/b]$ for which $\|x^{(k,\ell)}\|_2 > \epsilon_{h+1}/\lambda$ and Equation~(\ref{eq:coordinate_bd_sq_ell_2_equiv_violated}) holds (for some $i \in [N]$ and $j \in [\dsketch]$), with probability at least
\begin{equation}\label{eq:soundness_pf_prob_union_bd_part_3}
1- 2 \cdot S_h \cdot \frac{\dsketch}{b} \cdot e^{-\epsilon_{h+1}^2 \cdot (1-\frac{\epsilon_{h+1}}{8}) \cdot \frac{b}{128}}.
\end{equation}
By another union bound and using Equations~(\ref{eq:soundness_pf_prob_union_bd_first_part}), (\ref{eq:soundness_pf_prob_union_bd_second_part}) and (\ref{eq:soundness_pf_prob_union_bd_part_3}), we get with probability at least
\begin{equation}\label{eq:soudness_pf_union_bd_all_parts}
1 - \frac{S_h \cdot \dsketch}{b/} \cdot (0.1)^{b/3} - \frac{N \cdot \dsketch}{b} \cdot (\frac{A_1^2}{16 \cdot B_2^2})^{b/3} - \frac{2 \cdot S_h \cdot \dsketch}{b} \cdot e^{-\epsilon_{h+1}^2 \cdot (1-\frac{\epsilon_{h+1}}{8}) \cdot \frac{b}{128}},
\end{equation}
that for all $k \in [S_h]$ and $\ell \times [\dsketch/b]$, if, when examining block $(k, \ell)$, the tests in lines~\ref{line:test_1}, \ref{line:test_2} and \ref{line:test_3} of Algorithm~\ref{alg:recovery} all fail, then we are sure that $\overline{z}_{k, \ell}$ is $\epsilon_{h+1}$-close in Hamming distance to $R_i[(\ell-1)b+1:\ell b +1, j]$. Moreover, we have that if $i^*$ is set in lines \ref{line:begin_if_star}-\ref{line:end_if_star}, then $\hat{\sigma}_{i^*,m}$ is identical to to the matrix signature $\sigma_{i, m}$. Similarly, we can recover the correct column index $j \in [\dsketch]$. Recall that our permutation $\pi$ of $[N]$ is defined as follows. For each $i \in [N]$, we let $\pi(i)$ be the first index $i^*$ set in line~\ref{line:i_star_def} of Algorithm~\ref{alg:recovery} for which $\hat{\sigma}_{i^*,m}$ is $10\epsilon_{h+1}$-close in Hamming distance to $\hat{\sigma}_{i,m}$. If no such $i^*$ exists, we define $\pi(i)$ in an arbitrary way that doesn't result in a collision.

We now describe how to handle the general case where $z^{(k)}_1$ is not necessarily zero but where we are guaranteed that $\norm{z^{(k)}_1}{1} \le O(\sqrt{\dsketch})$. This can be done by observing that the number of blocks $\ell \in [\dsketch/b]$ for which 
$$\|z^{(k)}_1[(\ell-1)b +1: \ell b +1]\|_{\infty} \le O(\frac{b}{\sqrt{\dsketch}})$$
is at most $0.001 \cdot q\dsketch/b$. Thus, the same reasoning above combined with the test in line~\ref{line:test_4} of Algorithm~\ref{alg:recovery} implies that even in the presence of $\ell_1$ noise, the probability bound derived above on spurious blocks $(k,\ell)$ failing all tests still holds.

To sum up, the probability in Equation~(\ref{eq:soudness_pf_union_bd_all_parts}) is at least the desired $1-N^{-t}$ probability lower bound in the statement of Lemma~\ref{le:recovery_soundness} as long as
\begin{equation}\label{eq:soundness_pf_globa_prob_upper_bd}
\frac{S_h \cdot \dsketch}{b} \cdot (0.1)^{b} - \frac{N \cdot \dsketch}{b} \cdot (\frac{A_1^2}{16 \cdot B_2^2})^{b/3} - \frac{2 \cdot S_h \cdot \dsketch}{b} \cdot e^{-\epsilon_{h+1}^2 \cdot (1-\frac{\epsilon_{h+1}}{8}) \cdot \frac{b}{128}} < N^{-t}.
\end{equation}

Lemma~\ref{le:recovery_soundness} now follows by noting that Equations~(\ref{eq:soundness_pf_first_test}), (\ref{eq:soundness_pf_second_test}), (\ref{eq:soundness_pf_last_case_r}), (\ref{eq:middle_and_last}), (\ref{eq:soundness_pf_last_case_x}) and (\ref{eq:soundness_pf_globa_prob_upper_bd})  can be simultaneously satisfied (assuming that $S_h \le \poly(N)$ and $t$ is a given positive absolute constant) if we let $c$ be a sufficiently large positive constant and set
$$ \text{ any }\tau_1 = \Theta\bigg(\frac{\epsilon_{h+1}}{\sqrt{q \cdot \dsketch}}\bigg),$$
$$\tau_2 = \Theta( \epsilon_{h+1}^2),$$
$$ g = \Theta\bigg(\frac{1}{ \epsilon_{h+1}^2 \cdot q \cdot \dsketch}\bigg), $$
$$ \lambda = \Theta(1), $$
$$ \epsilon_h = \Theta( \epsilon_{h+1}^c),$$
$$ b = \Theta \bigg(\frac{\log{N} + \log{\dsketch} + \log{S_h}}{\epsilon_{h+1}
^2} \bigg),$$
and
$$q = \Theta\bigg(\frac{\log{N} + \log{\dsketch} +\log{S_h}}{\epsilon_{h+1}^2 \cdot \sqrt{\dsketch}}\bigg),$$
where $b$ and $q$ are sufficiently large.

}{}

\end{document}